%% file: vp-landscape.tex
\documentclass[journal]{IEEEtran}
\usepackage{epsfig}
\usepackage{graphicx}
\usepackage{amsmath}
\usepackage{amssymb}

\usepackage{fancyhdr,amsfonts,color}
\usepackage{algorithmic}
\usepackage{algorithm}

\usepackage{xcolor}

\newcommand{\E}{\mathcal{E}}

\newcommand{\C}{\mathcal C}

\newcommand{\x}{\mathbf x}

\newcommand{\vv}{\mathbf v}
\newcommand{\cc}{\mathbf c}

\newcommand{\q}{\mathbf q}
\newcommand{\p}{\mathbf p}
\renewcommand{\ll}{\mathbf l}

\renewcommand{\aa}{\mathbf a}

\renewcommand{\Re}{\mathbb R}

\renewcommand{\mathbf}{\boldsymbol}

\newcommand{\nop}[1]{}
\newcommand{\eg}{{\sl e.g.}}
\newcommand{\ie}{{\sl i.e.}}

\newcommand{\etal}{{\sl et al.}}

\newcommand{\cfbox}[2]{%
    \colorlet{currentcolor}{.}%
    {\color{#1}%
    \fbox{\color{currentcolor}#2}}%
}
\begin{document}
%
\title{Detecting Dominant Vanishing Points in Natural
 Scenes with Application to Composition-Sensitive Image Retrieval}
%
%
%

\author{Zihan Zhou,
        Farshid Farhat,
        and~James Z. Wang
\thanks{Z. Zhou and J. Z. Wang are with College of Information Sciences and Technology, The Pennsylvania State University, USA (e-mail: zzhou@ist.psu.edu; jwang@ist.psu.edu.). F. Farhat is with the School of Electrical Engineering and Computer Science, The Pennsylvania State University, USA (e-mail: fuf111@cse.psu.edu).}
}

%
%

\markboth{IEEE Transactions on Multimedia,~Vol.~XXX, No.~XXX, April~2017}%
{Shell \MakeLowercase{\textit{et al.}}: Bare Demo of IEEEtran.cls for IEEE Journals}
%



\maketitle

\begin{abstract}
Linear perspective is widely used in landscape photography to create the impression of depth on a 2D photo. Automated understanding of linear perspective in landscape photography has several real-world applications, including aesthetics assessment, image retrieval, and on-site feedback for photo composition, yet adequate automated understanding has been elusive. We address this problem by detecting the dominant vanishing point and the associated line structures in a photo. However, natural landscape scenes pose great technical challenges because often the inadequate number of strong edges converging to the dominant vanishing point is inadequate. To overcome this difficulty, we propose a novel vanishing point detection method that exploits global structures in the scene via contour detection. We show that our method significantly outperforms state-of-the-art methods on a public ground truth landscape image dataset that we have created. Based on the detection results, we further demonstrate how our approach to linear perspective understanding provides on-site guidance to amateur photographers on their work through a novel viewpoint-specific image retrieval system.
\end{abstract}


%
\IEEEpeerreviewmaketitle

\input{1intro}
\input{2relatedwork}

\input{3dataset}

\input{4vpdetection}

\input{5vpselection}
\input{6application}
\input{7conclusion}

\section*{Acknowledgment}
The authors would like to thank Jeffery Cao and Edward J. Chen 
for assistance with ground truth labeling and developing 
online systems for user studies. 
We would also like to acknowledge the comments and
constructive suggestions from anonymous reviewers and the associate
editor.

\ifCLASSOPTIONcaptionsoff
  \newpage
\fi


\bibliographystyle{IEEEtran}
\bibliography{vp-landscape}  

\begin{IEEEbiography}[{\includegraphics[width=1.2in,height=1.3in, clip,keepaspectratio,trim={25 0 25 0}]{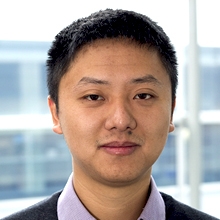}}]{Zihan Zhou} received the bachelor's degree in Automation from Tsinghua University, Beijing, China, in 2007, and the MS and PhD degrees in Electrical and Computer Engineering from University of Illinois at Urbana-Champaign in 2010 and 2013, respectively. He is currently a faculty member in the College of Information Sciences and Technology at Pennsylvania State University. His research interests lie in computer vision and image processing, specially in developing new computational tools for efficient and robust discovery of low-dimensional data structure for 3D scene analysis and modeling. He is a member of the IEEE.
\end{IEEEbiography}

\begin{IEEEbiography}[{\includegraphics[width=1.2in,height=1.3in, clip,keepaspectratio,trim={15 0 15 0}]{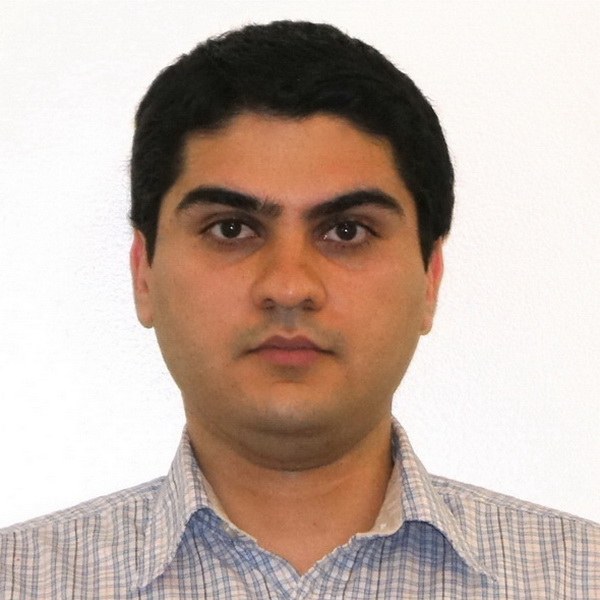}}]{Farshid Farhat} is a PhD candidate at the School of Electrical Engineering and Computer Science, The Pennsylvania State University. He obtained his B.Sc., M.Sc., and Ph.D. degrees in Electrical Engineering from Sharif University of Technology, Tehran, Iran. His current research interests include computer vision, image processing, distributed systems, and networking. He is working on the modeling and analysis of image composition and aesthetics using deep learning and image retrieval on different platforms ranging from smartphones to high-performance computing clusters.
\end{IEEEbiography}

\begin{IEEEbiography}[{\includegraphics[width=1.2in,height=1.3in, clip,keepaspectratio,trim={0 0 0 0}]{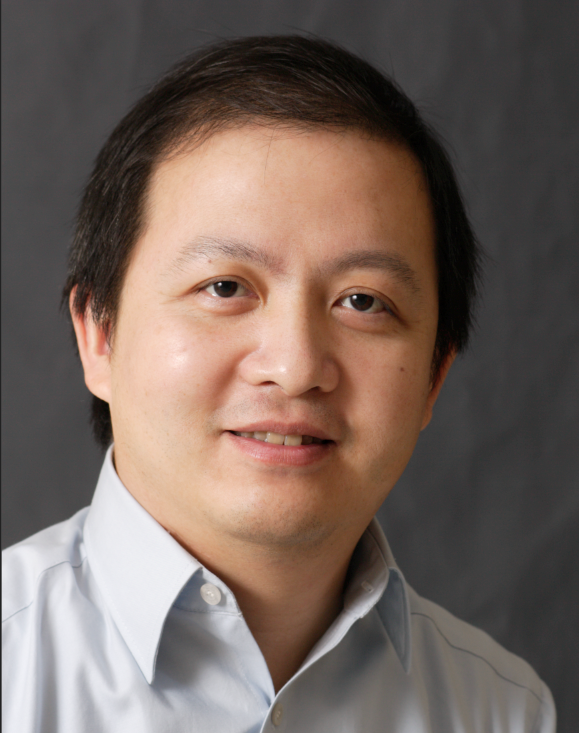}}]{James Z. Wang} is a Professor of Information Sciences and Technology at The Pennsylvania State University. He received the bachelor's degree in mathematics and computer science {\it summa cum laude} from the University of Minnesota, and the MS degree in mathematics, the MS degree in computer science and the PhD degree in medical information sciences, all from Stanford University. His research interests include computational aesthetics and emotions, automatic image tagging, image retrieval, and computerized analysis of paintings. He was a visiting professor at the Robotics Institute at Carnegie Mellon University (2007-2008), a lead special section guest editor of the IEEE Transactions on Pattern Analysis and Machine Intelligence (2008), and a program manager at the Office of the Director of the National Science Foundation (2011-2012). He was a recipient of a National Science Foundation Career award (2004).
\end{IEEEbiography}


\end{document}

%% file: 1intro.tex
\section{Introduction}
\label{sec:intro}

%
%
%
%

\IEEEPARstart{R}{ecently}, with the widespread use of digital cameras and other mobile
imaging devices, the multimedia
community has become increasingly interested in building intelligent programs to automatically analyze
the visual aesthetics and composition of photos. Information about
photo aesthetics and composition~\cite{Krages:2005} is shown
to benefit many real-world applications. 
For example, it can be
used to suggest improvements to the aesthetics and composition of
photographers' work through image re-targeting~\cite{batta10,Liu2010},
as well as provide on-site feedback to the photographer at the point
of photographic creation~\cite{yao2012oscar, NiXCWYT13}. 

In this paper, we focus on an important principle in photo composition, namely, the use of \emph{linear perspective}. It corresponds to a relatively complex spatial system that primarily concerns the \emph{parallel lines} in a scene.
Indeed, parallel lines are one of the most prevalent geometric structures in both man-made and natural environments. Under the pinhole camera model, parallel lines in 3D space project to converging lines in the image plane. The common point of intersection, perhaps at infinity, is called the \emph{vanishing point} (VP)~\cite{HartleyR2000}. Because the VPs provide crucial information about the
geometric structure of the scene, automatic detection of VPs have long
been an active research problem in image understanding.

Existing VP detection methods mainly focus on \emph{man-made environments}, which
typically consist of a large number of edges or line segments aligned
to one or more dominant directions. Numerous methods have been proposed to
cluster line segments into groups, each representing a VP in the
scene~\cite{Tardif09, TretiakBKL12,
  WildenauerH12, XuOH13, LezamaGRM14}. These methods have successfully
found real-world applications such as
self-calibration, 3D reconstruction of urban
scenes, and stereo matching. 

\begin{figure}[t!]
\centering
\includegraphics[height =0.78in]{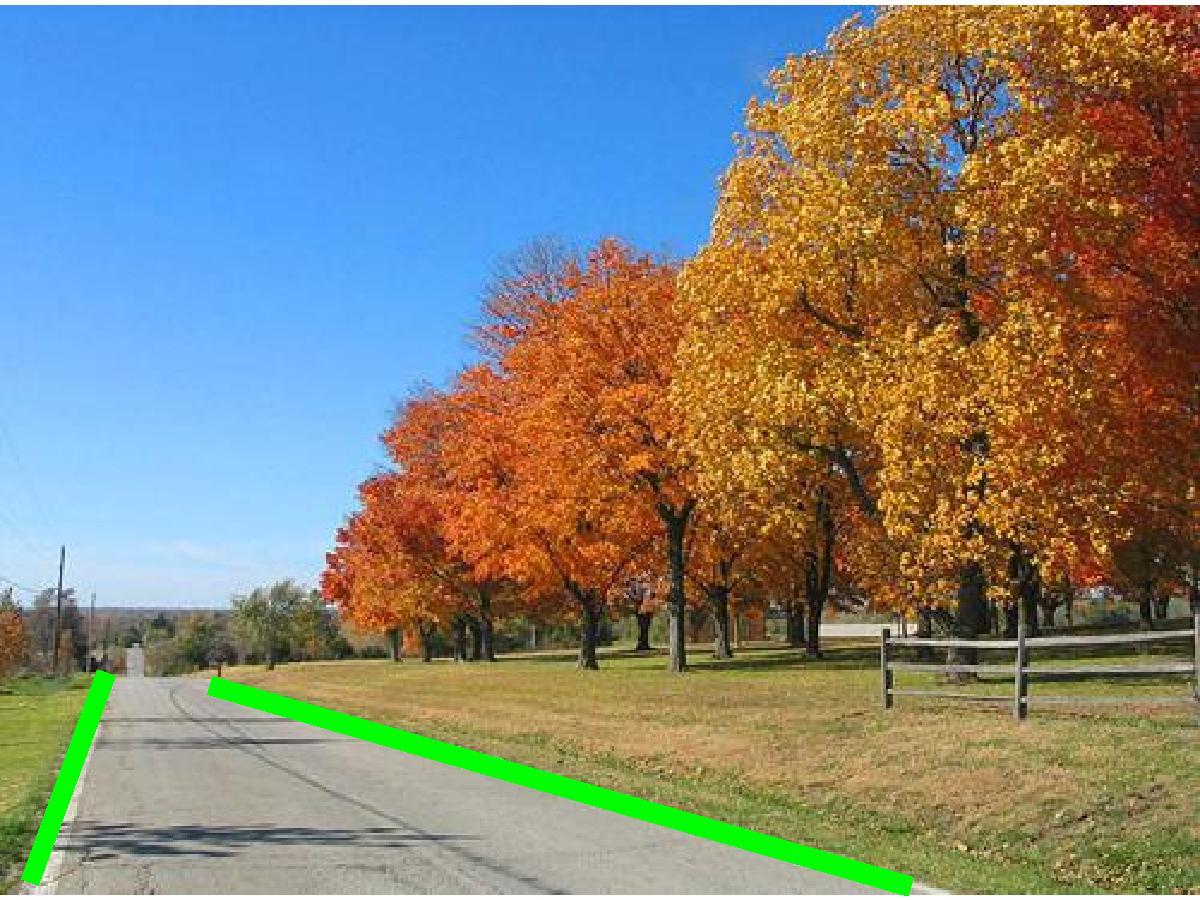}
\includegraphics[height =0.78in]{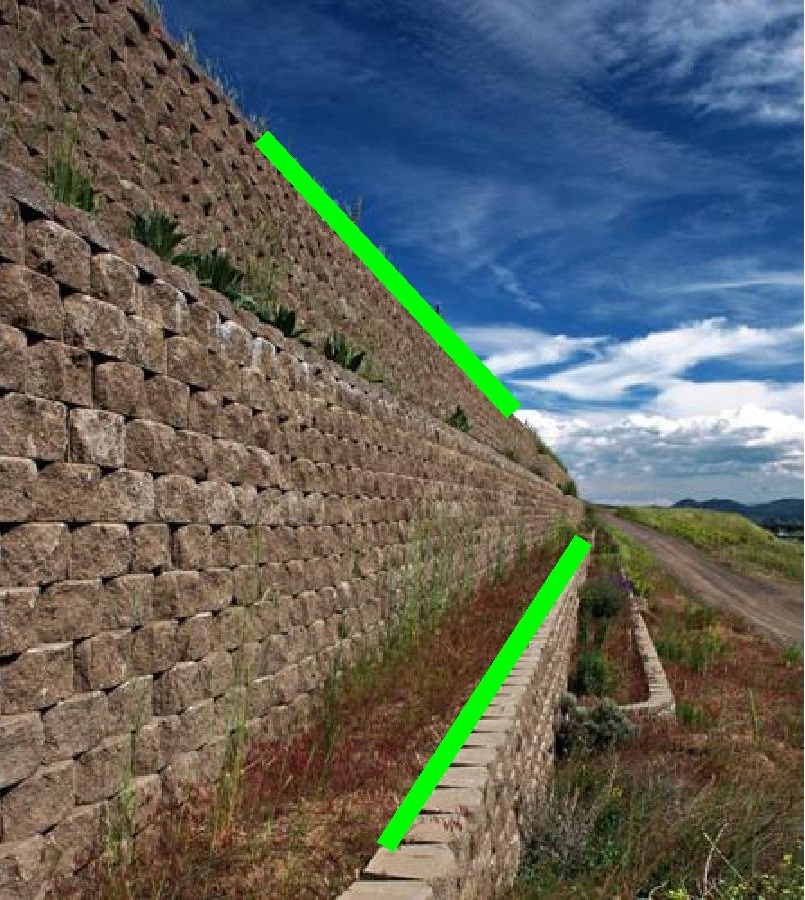}
\includegraphics[height =0.78in]{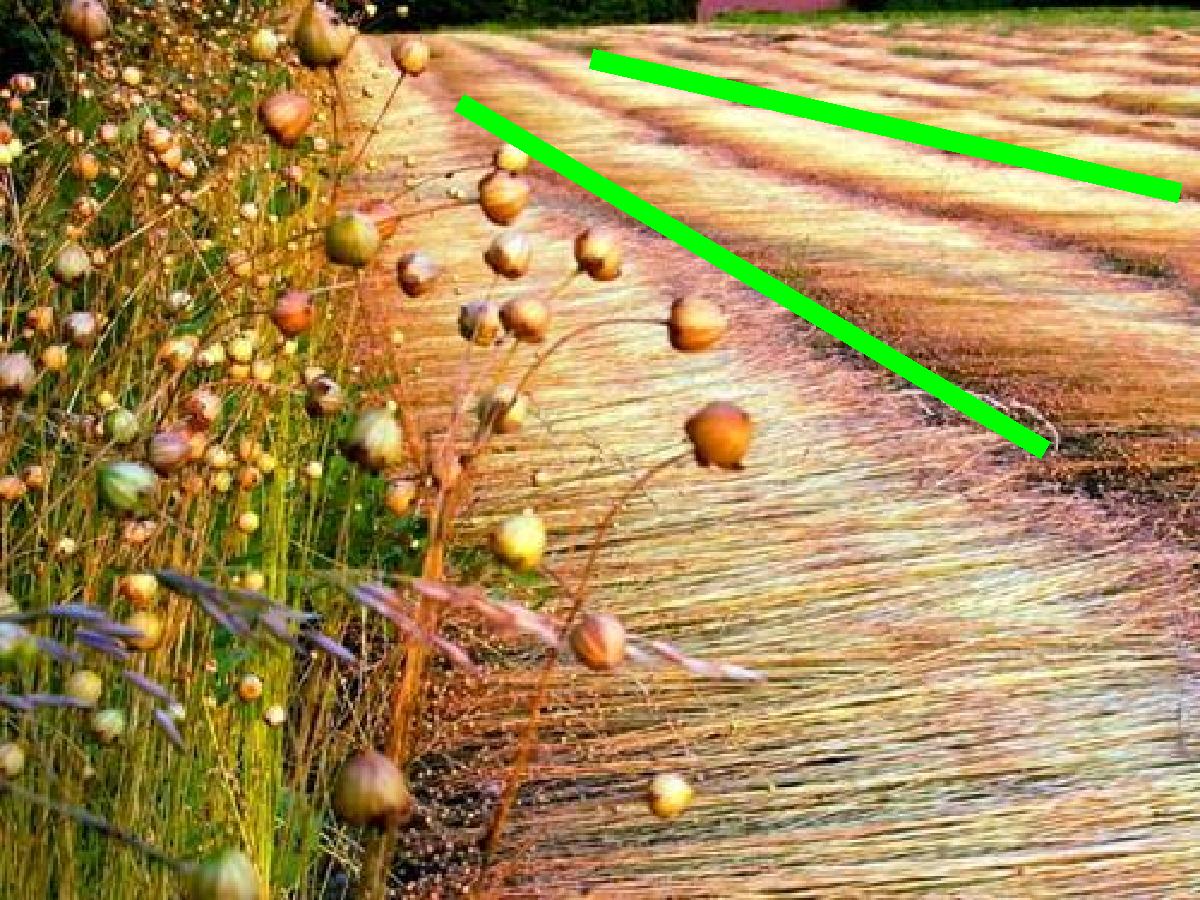}
\includegraphics[height =0.78in]{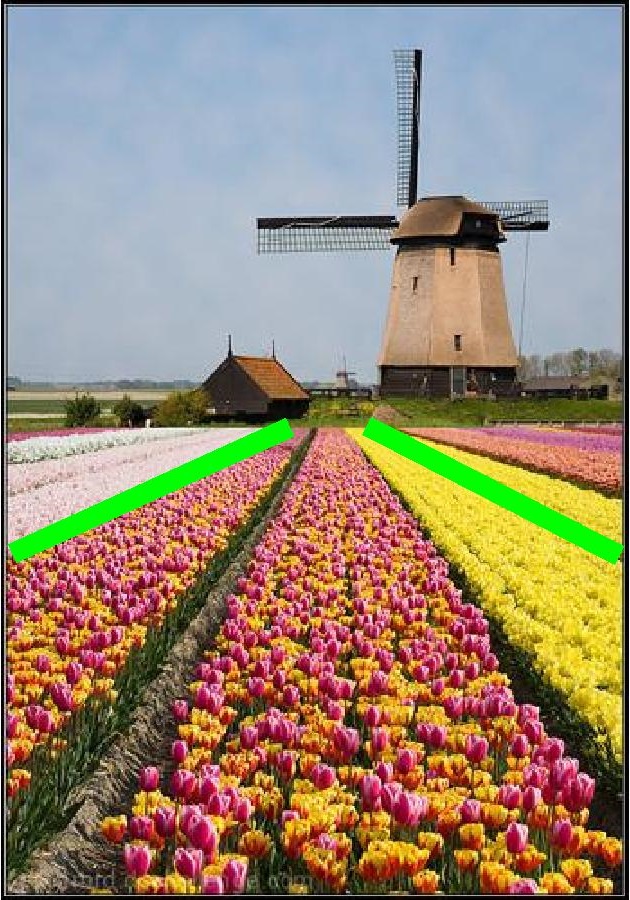}\\
\vspace{0.3mm}
\includegraphics[height =0.685in]{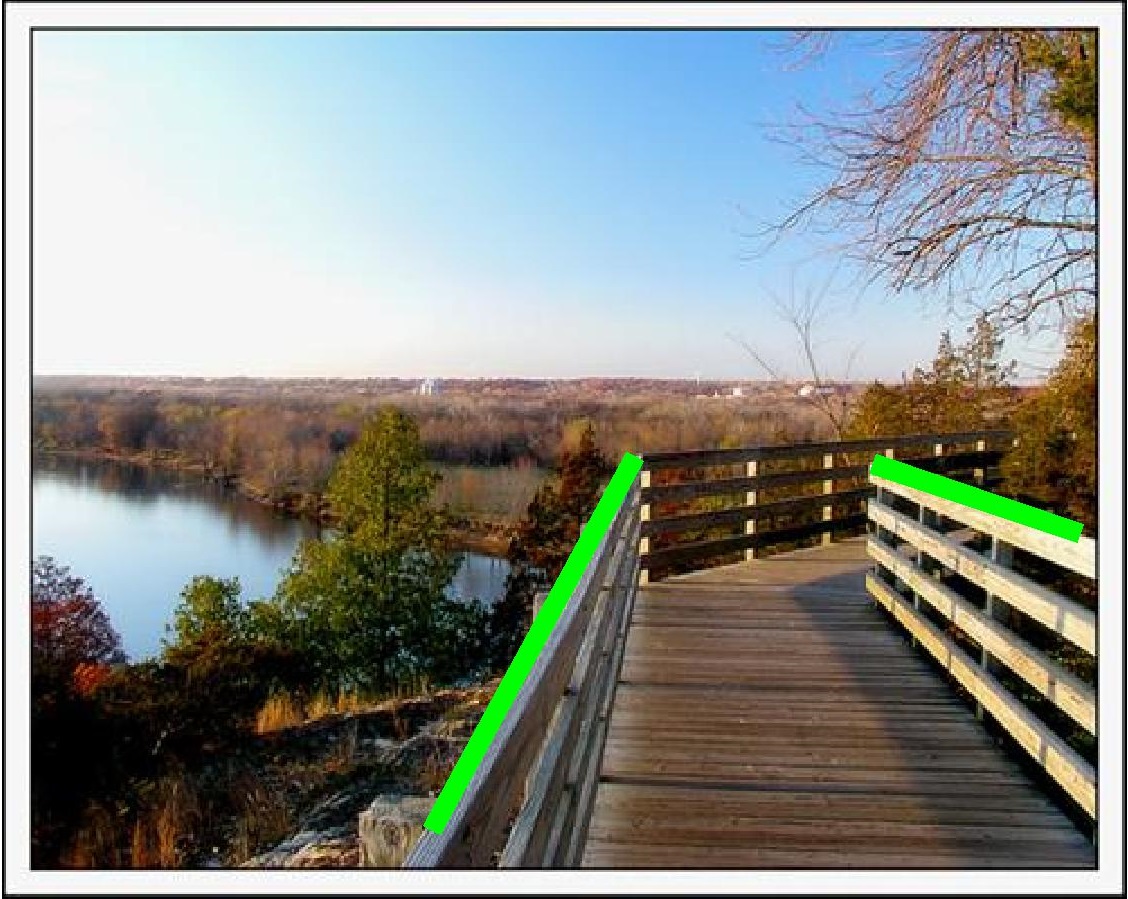}
\includegraphics[height =0.685in]{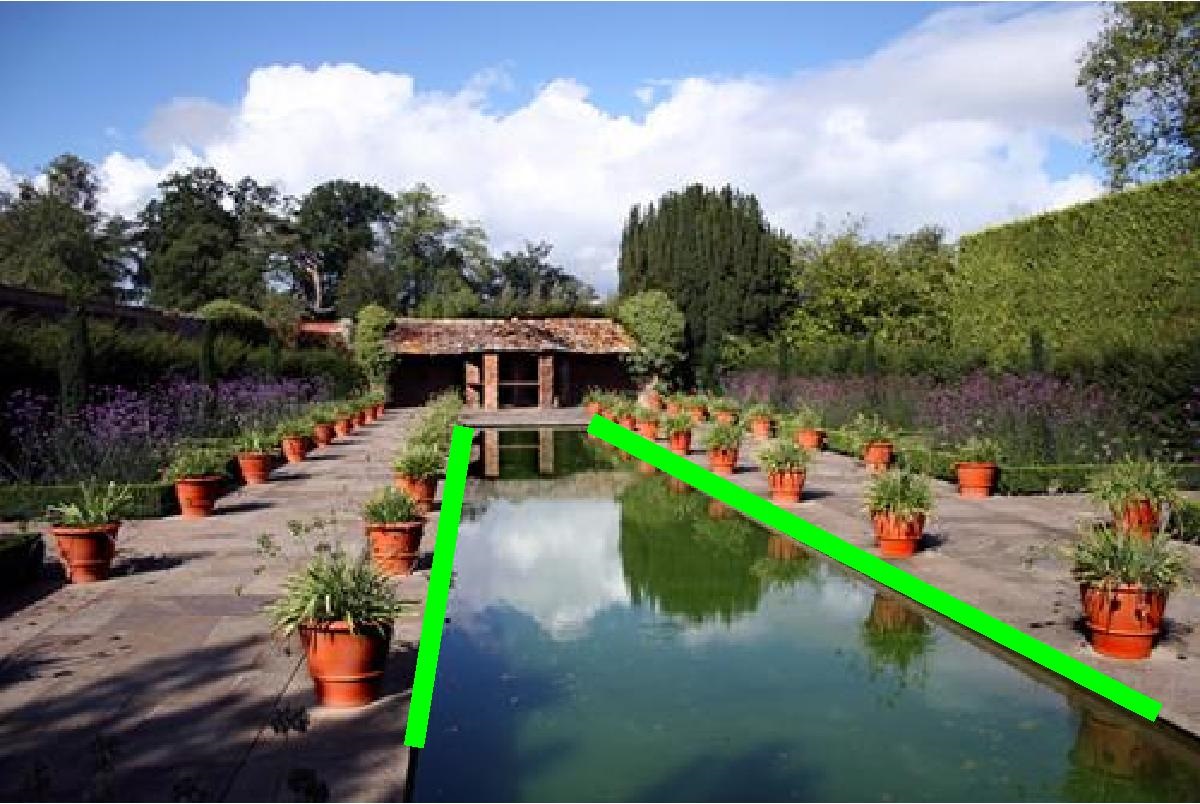}
\includegraphics[height =0.685in]{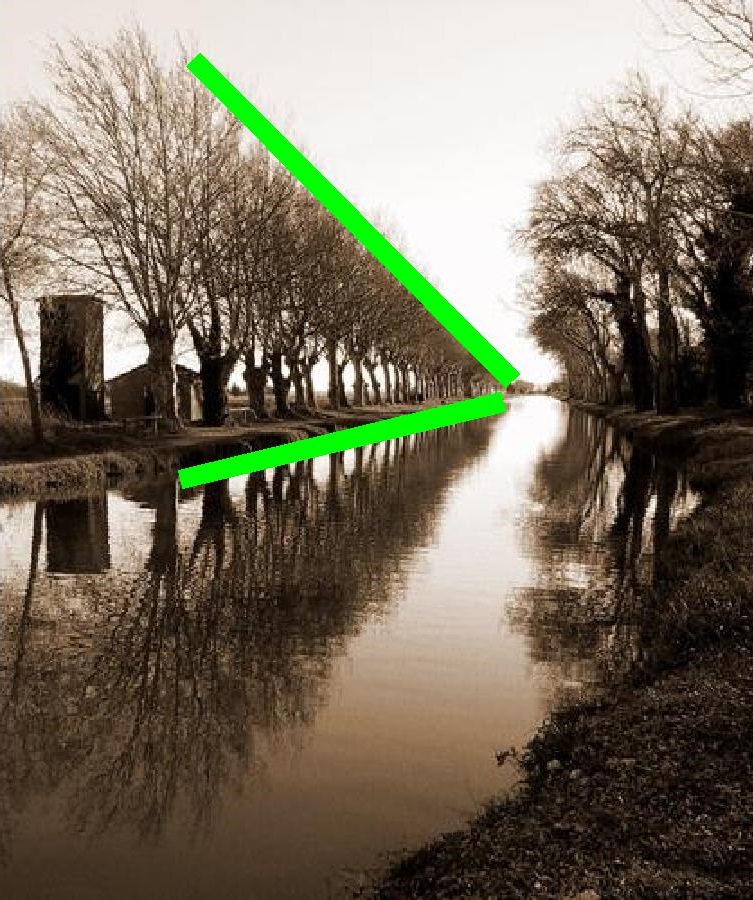}
\includegraphics[height =0.685in]{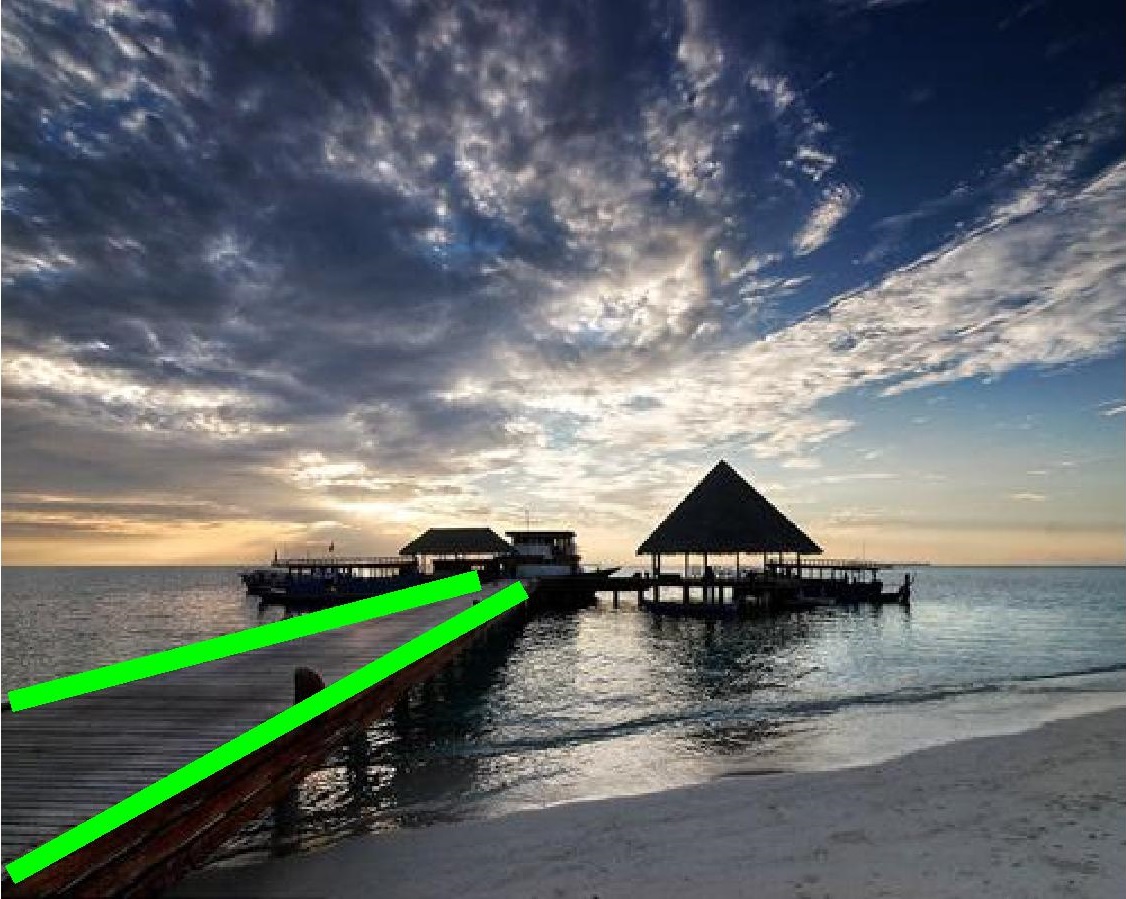}\\
\vspace{0.3mm}
\includegraphics[height =0.64in]{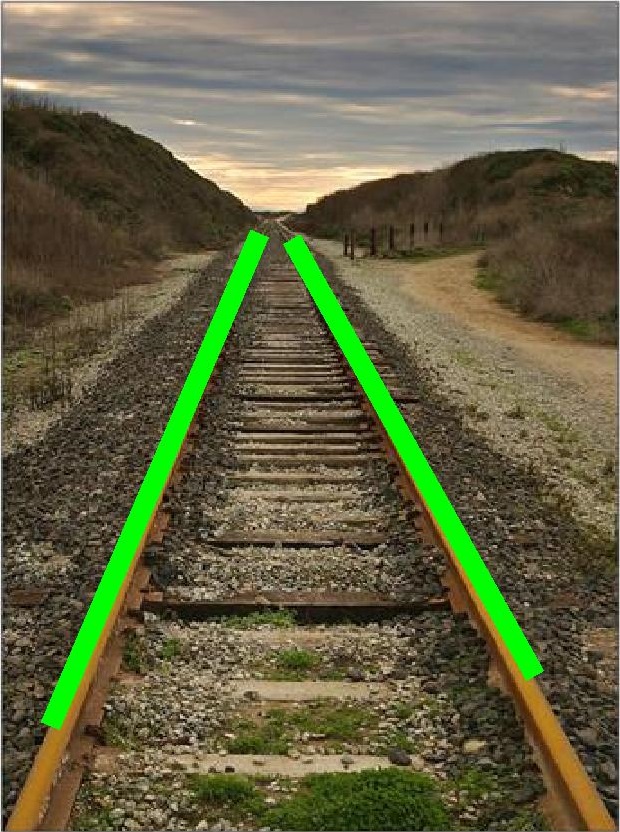}
\includegraphics[height =0.64in]{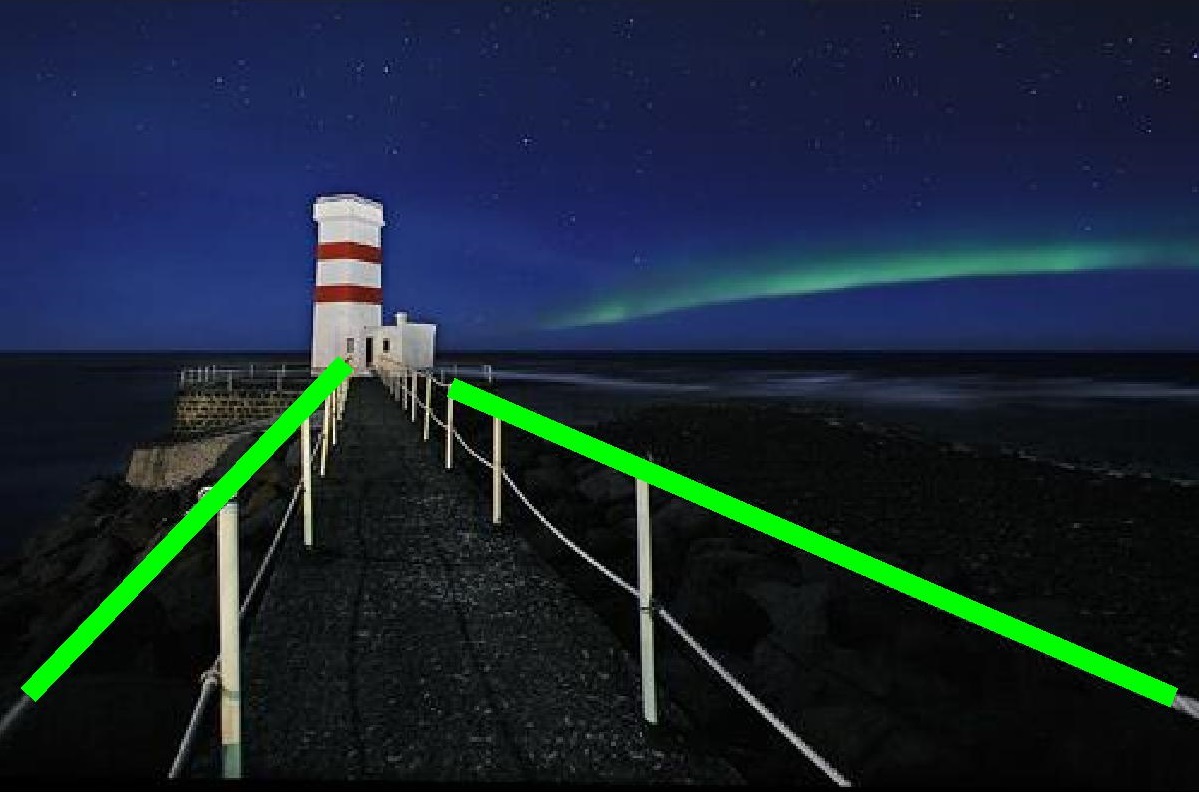}
\includegraphics[height =0.64in]{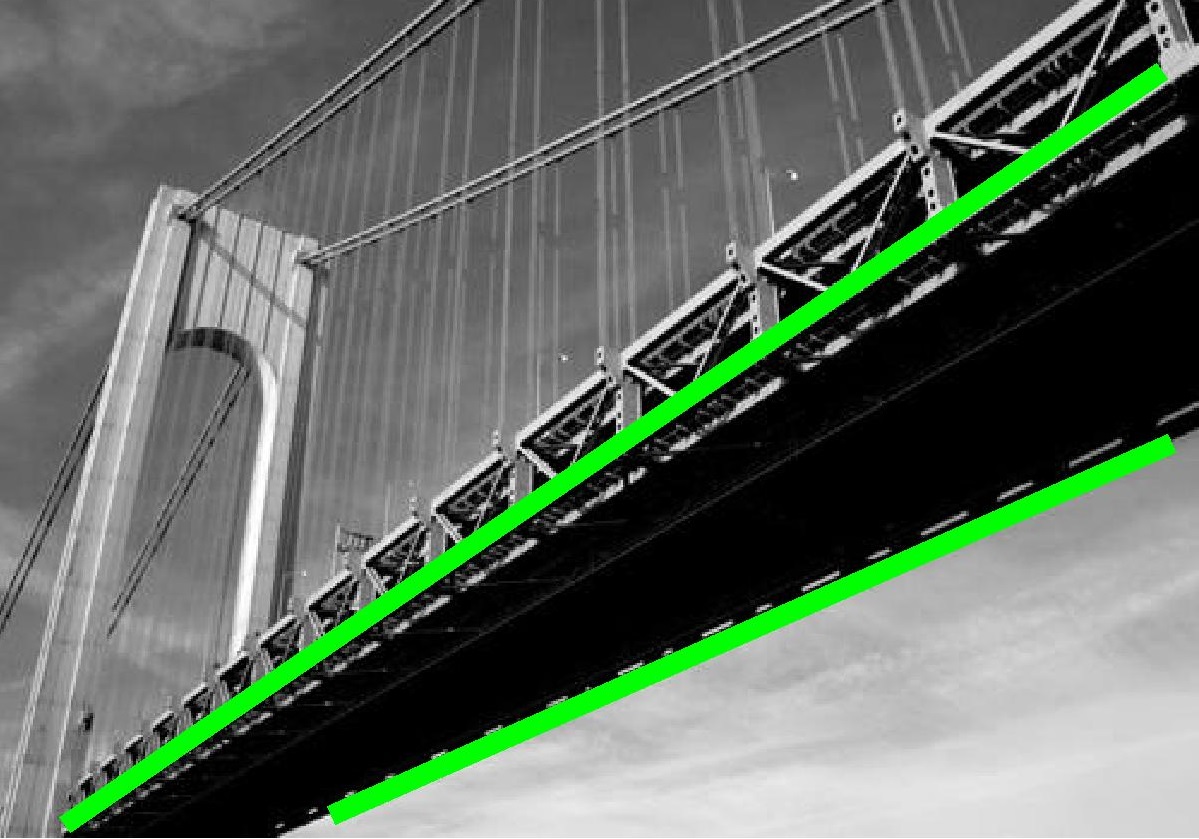}
\includegraphics[height =0.64in]{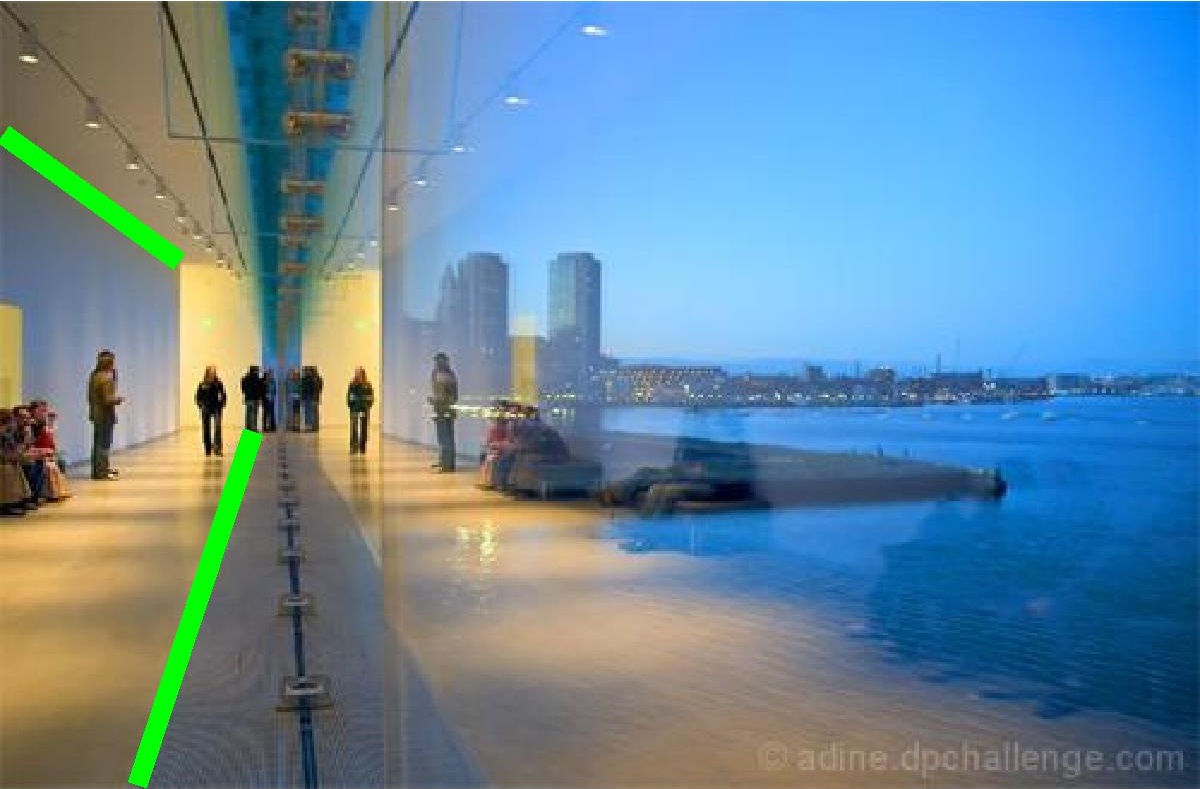}
\caption{Natural scene images with dominant vanishing points. Images are from the ``landscape'' category of the AVA dataset. Manually labeled ground truth lines are marked in green. }
\label{fig:example}
\end{figure}

However, little attention has been paid to \emph{natural landscape scenes} -- a significant genre in both professional and consumer photography. Known as an effective tool in recognizing the viewer as a specific unique individual in a distinct place with a point of view~\cite{LauerP11}, linear perspective has been widely used in natural landscape photography. To bridge this gap between man-made environments and natural landscapes, we conduct a systematical study on the use of linear perspective in such scenes through the detection of the \emph{dominant vanishing point} in the image. Specifically, we regard a VP in the image the dominant VP if it (i) associates with the major geometric structures (\eg, roads, bridges) of the scene, and (ii) conveys a strong impression of 3D space or depth to the viewers. As illustrated in Figure~\ref{fig:example}, the dominant VP provides us with important cues about the scene geometry as well as the overall composition of the image.

In natural scene images, a VP is detectable when there are as few as two parallel lines in space. 
While human eyes have little difficulty identifying the VPs in these images, automatic detection of VPs poses a great challenge to computer systems for two main reasons. First, the visible edges can be weak and not detectable via local photometric cues. Existing line segment detection methods typically assume the gradient magnitude of an edge pixel to be above a certain threshold (\eg, Canny edge detector~\cite{Canny86a}) or the number of pixels with aligned gradient orientations to be above a certain threshold (\eg, LSD~\cite{von2012lsd}). However, determining the threshold can be difficult due to the weak edges and image noise. Second, the number of edges converging to the VP may be small compared to irrelevant edges in the same scene. As a result, even if a computer system can detect the converging edges, clustering them into one group can be demanding due to the large number of outliers.

To overcome these problems, we propose the use of \emph{image contours} to detect edges in natural images. Compared to local edge detectors, image contours encode valuable global information about the scene, and thus they are more effective in recognizing weak edges while reducing the number of false detections due to textures. By combining the contour-based edge detection with J-Linkage~\cite{ToldoF08}, a popular multi-model detection algorithm, our method significantly outperforms state-of-the-art methods on detecting the dominant VP in natural scene images.

As an application of our VP detection method, we demonstrate how the detected VPs can be used to improve the usefulness of existing content-based image retrieval systems in providing on-site feedback to amateur photographers.
Specifically, given a photo taken by the user, we study the problem of finding photos about similar scenes and with \emph{similar viewpoints} in a large collection of photos. As these photos demonstrate, the use of various photographic techniques, in a situation similar to the one that the user is currently engaged in, could potentially provide effective guidance to the user on his or her own work. Further, in this task, we are also the first to answer an important yet largely unexplored question in the literature: \emph{How to determine whether there exists a dominant VP in a photo?} To this end, we design a new measure of strength for a given candidate VP, and systematically examine its effectiveness on our dataset. 

In summary, the {\bf main contributions} are as follows:

\begin{itemize}

\item We propose the research problem of detecting VPs in natural landscape scenes and a new method for dominant VP detection. By combining a contour-based edge detector with J-Linkage, our method significantly outperforms state-of-the-art methods for natural scenes.

\item We develop a new strength measure for VPs and demonstrate its effectiveness in identifying images with a dominant VP.

\item We demonstrate the application of our method for assisting amateur photographers at the point of photographic creation via viewpoint-specific image retrieval.

\item To facilitate future research, we have created and made available two manually labeled datasets for dominant VPs in 2,275 real-world natural scene images.

\end{itemize}

%% file: 2relatedwork.tex
\section{Related Work}\label{sec:rel}

\subsection{Vanishing Point Detection}

Vanishing point detection has long been an active research topic in computer vision with many real-world applications such as camera calibration~\cite{CaprileT90, CipollaDR99}, pose estimation~\cite{AntoneT00, KoseckaZ02, DenisEE08}, height measurement~\cite{AndaloTG15}, object detection~\cite{HoiemEH08}, and 3D reconstruction~\cite{ParodiP96, CriminisiRZ00}. Since the VPs can be represented by normalized 2D homogeneous coordinates on a Gaussian sphere, early works on VP detection use a Hough transform of the line segments on the Gaussian sphere~\cite{Barnard83, QuanM89, CollinsW90, Brillault-OMahony91} or a cube map~\cite{TuytelaarsPG97}. However, researchers have criticized these methods as being unreliable in the presence of noise and outliers and as having the potential to lead to spurious detections~\cite{Shufelt99}. To reduce spurious responses, Rother~\cite{Rother00}  enforces the orthogonality of multiple VPs via an exhaustive search, but this process is computationally expensive and requires a criterion to distinguish finite from infinite vanishing points. Alternatively, the Helmholtz principle has been employed to reduce the number of false alarms and reach a high precision level~\cite{AlmansaDV03}.

In~\cite{AntoneT00}, Antone and Teller propose an Expectation-Maximization (EM) scheme for VP detection, which is later extended to handle uncalibrated cameras~\cite{KoseckaZ02, SchindlerD04}. The EM method iteratively estimates the probability that a line segment belongs to each parallel line groups or an outlier group (E-step) and refines the VPs according to the line segment groups (M-step). The approach, however, requires a good initialization of the vanishing points, which is often obtained by Hough transform or clustering of edges based on their orientations~\cite{McLeanK95}.

More recent VP detection algorithms use RANSAC-like algorithms to generate VP hypotheses by computing the intersections of line segments and then select the most probable ones using image-based consistency metrics~\cite{Tardif09, TretiakBKL12}. In~\cite{XuOH13}, Xu~\etal~studied various consistency measures between the VPs and line segments and developed a new method that minimizes the uncertainty of the estimated VPs. Lezama~\etal~\cite{LezamaGRM14} proposed to find the VPs via point alignments based on the \emph{a contrario} methodology. Instead of using edges, Vedaldi and Zisserman proposed to detect VPs by aligning self-similar structures~\cite{VedaldiZ12}. 
Zhai~\etal~\cite{ZhaiWJ16} used global image context extracted from a deep convolutional network to guide the search for VPs. 
As we discussed before, these methods are designed for man-made environments. Identifying and clustering edges for VP detection in natural scene images remain a challenging problem.

In addition, there is an increasing interest in exploiting special scene
structures for VP detection lately. For example, many methods assume
the ``Manhattan World'' model~\cite{CoughlanY03, DenisEE08}, indicating that
three orthogonal parallel-line clusters are present~\cite{MirzaeiR11,
  BazinSDVIKP12, WildenauerH12, AntunesB13}. When the
assumption holds, it is shown to improve the VP detection
results. Unfortunately, such an assumption is invalid for typical natural
scenes. Other related efforts detect VPs in specific scenes such as
unstructured roads~\cite{Rasmussen04, KongAP09, MoghadamSW12}, but how these methods can be extended to general natural scenes remains unclear.

\subsection{Photo Composition Modeling}

Photo composition, which describes the placement or arrangement of visual elements or objects within the frame, has long been a subject of study in computational photography. A line of work has addressed known photographic rules and design principles, such as simplicity, depth of field, golden ratio, rule of thirds, and visual balance. Based on these rules, various image retargeting and recomposition tools have been proposed to improve the image quality~\cite{batta10, Liu2010, ZhangWH13, FangLMS14}. We refer readers to~\cite{Islam2016} for a comprehensive survey on this topic. However, the use of linear perspective has largely been ignored in the existing work. Compared to the aforementioned rules, which mainly focus on the 2D rendering of visual elements, linear perspective enables photographers to convey the sense of 3D space to the viewers.

Recently, data-driven approaches to composition modeling have gained increasing attention in the multimedia community. These methods use community-contributed photos to automatically learn composition models from the data. For example, Yan~\etal~\cite{YanLKT15} propose a set of composition features and learn a model for automatic removal of distracting content and enhancement of the overall composition. In~\cite{DHongZT16}, a unified photo enhancement framework is proposed based on the discovery and modeling of aesthetic communities on Flickr. Besides offline image enhancement, composition models learned from exemplar photos can also been used to provide online aesthetics guidance to the photographers, such as selecting the best view~\cite{SuCKHC12, NiXCWYT13}, recommending the locations and poses of human subjects in a photograph~\cite{MaFC14, XuYJLS14, RawatK15}, and suggesting the appropriate camera parameters (\eg, aperture, ISO, and exposure)~\cite{YinMCL14}. Our work also takes advantage of vast data available through photo sharing websites. Unlike existing work that each focuses on certain specific aspects of the photo, however, we take a completely different approach to on-site feedback and aim to provide comprehensive photographic guidance through a novel composition-sensitive image retrieval system.

\subsection{Image Retrieval} 

The classic approaches to content-based image retrieval
\cite{smeulders2000content} typically measure the visual similarity
based on low-level features ({\it e.g.}, color,
texture, and shape). Recently, thanks to the availability of
large-scale image datasets and computing resources, complicated models
have been trained to capture the high-level semantics about the scene
\cite{datta2008image, JegouDSP10, ShrivastavaMGE11, RazavianASC14}. However, because many visual descriptors are
generated by local feature extraction processes, the overall spatial
composition of the image (\ie, from which viewpoint the image is
taken) is usually neglected. To remedy this issue, \cite{yao2012oscar} first classifies images into pre-defined composition categories such as ``horizontal'', ``vertical'', and ``diagonal''. Similar to our work, \cite{ZhouHLW15} also explores the VPs in the image for retrieval, but it assumes known VP locations in all images, and thus cannot be applied to a general image database where the majority of the images do not contain a VP. Further, \cite{ZhouHLW15} does not consider image semantics for retrieval, hence its usage can be limited in practice.

%% file: 3dataset.tex
\section{Ground Truth Dataset}
\label{sec:dataset}

To create our ground truth datasets, we leverage both the open AVA dataset~\cite{MurrayMP12} and the online photo-sharing website Flickr. We have made our datasets publicly available.\footnote{https://faculty.ist.psu.edu/zzhou/projects/vpdetection} Below we describe them in detail. 

\medskip
\noindent{\bf AVA landscape dataset.} The original AVA dataset contains over 250,000 images along with a variety of annotations. The dataset provides semantic tags describing the semantics of the images for over 60 categories, such as ``natural'', ``landscape'', ``macro'', and ``urban''. For this work, we used the 21,982 images under the category ``landscape''.

For each image, we need to determine whether it contains a dominant VP and, if so, label its location. Note that our ground truth data is quite different from those in existing datasets such as York Urban Dataset (YUD)~\cite{DenisEE08} and Eurasian Cities Dataset (ECD)~\cite{TretiakBKL12}. While these datasets are focused on urban scenes and attempt to identify \emph{all} VPs in each image, our goal is to identify a \emph{single dominant} VP associated with the main structures in a wide variety of scenes. The ability to identify the dominant VP in a scene is critical in our targeted applications related to aesthetics and photo composition.

\begin{figure}[t!]
\centering
\includegraphics[height =0.62in]{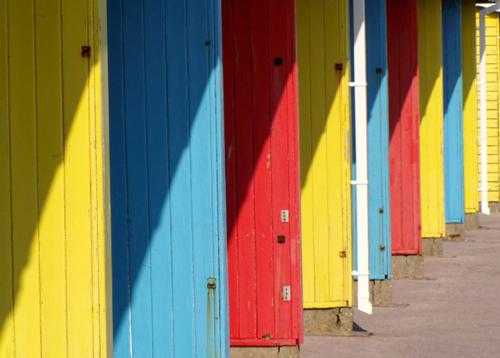}
\includegraphics[height =0.62in]{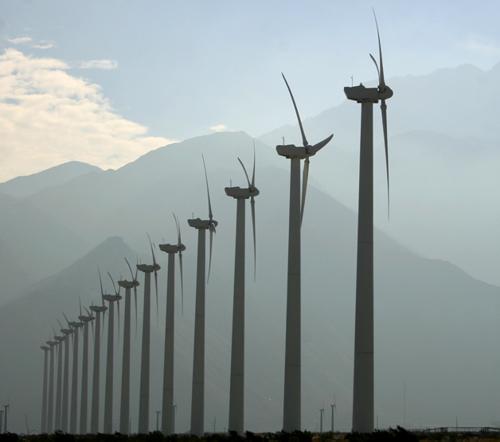}
\includegraphics[height =0.62in]{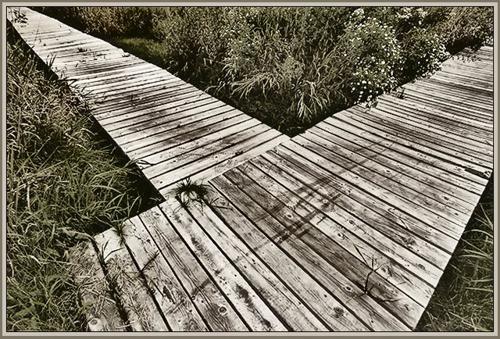}
\includegraphics[height =0.62in]{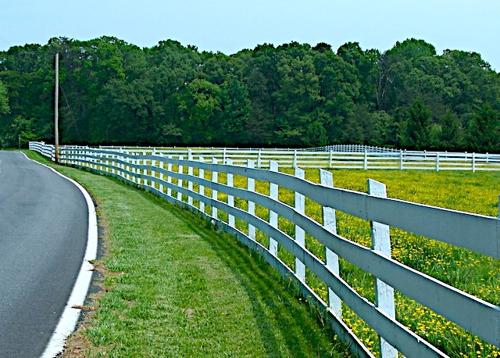}
\caption{Example natural scene images that are {\bf not} suitable for this work. The first two images show diminishing perspective. The third image has two VPs. The last image contains parallel curves, not parallel lines.}
\label{fig:example2}
\end{figure}

Like existing datasets, we label the dominant VP by manually
specifying at least two parallel lines in the image, denoted as
$\ll_1$ and $\ll_2$ (see Figure~\ref{fig:example}). The dominant VP
location is then computed as $\vv = \ll_1 \times \ll_2$.  Because our
goal is to identify the dominant VPs only, we make a few assumptions
during the process. First, each VP must correspond to at least two
visible parallel lines in the image. This assumption eliminates other types of
perspective in photography such as diminishing perspective, which is
formed by placing identical or similar objects at different
distances. Second, for a VP to be the dominant VP in an image, it must
correspond to some major structures of the scene and clearly carries
more visual weight than other candidates, if any. We do not consider
images with two or more VPs carrying similar visual importance, which
are typically seen in urban scenes. Similarly, we also exclude images
where it is impossible to determine a single dominant direction due to
parallel curves (Figure~\ref{fig:example2}). Finally, observing that
only those VPs which lie within or near the image frame convey a
strong sense of perspective to the viewers, we resize each image so
that the length of its longer side is 500 pixels, keeping only the
dominant VPs that lie within a $1,000\times 1,000$ frame, with the image
placed at the center. We used the size $500$ pixels as a trade-off
between keeping details and providing fast runtime for
large-scale applications.

We collected a total of 1,316 images with annotations of ground truth
parallel lines. 

\medskip
\noindent{\bf Flickr dataset.} To test the generality of our VP detection method, we have also downloaded images from Flickr by querying the keyword ``vanishing point''. These images cover a wide variety of scenes from nature, rural, and travel to street and cityscape. Following the same procedure above, for each image, we determine whether it contains a dominant VP and, if so, label its location. We only keep the images with a dominant VP. This step results in a dataset consisting of 959 annotated images.

%% file: 4vpdetection.tex
\section{Contour-based Vanishing Point Detection for Natural Scenes}

\begin{figure*}[t!]
\centering
\begin{tabular}{ccc}
\includegraphics[height =1.2in]{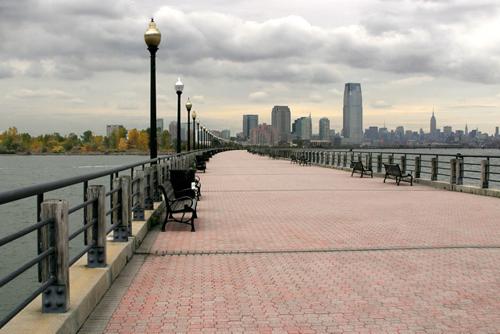} &
\fbox{\includegraphics[height =1.2in]{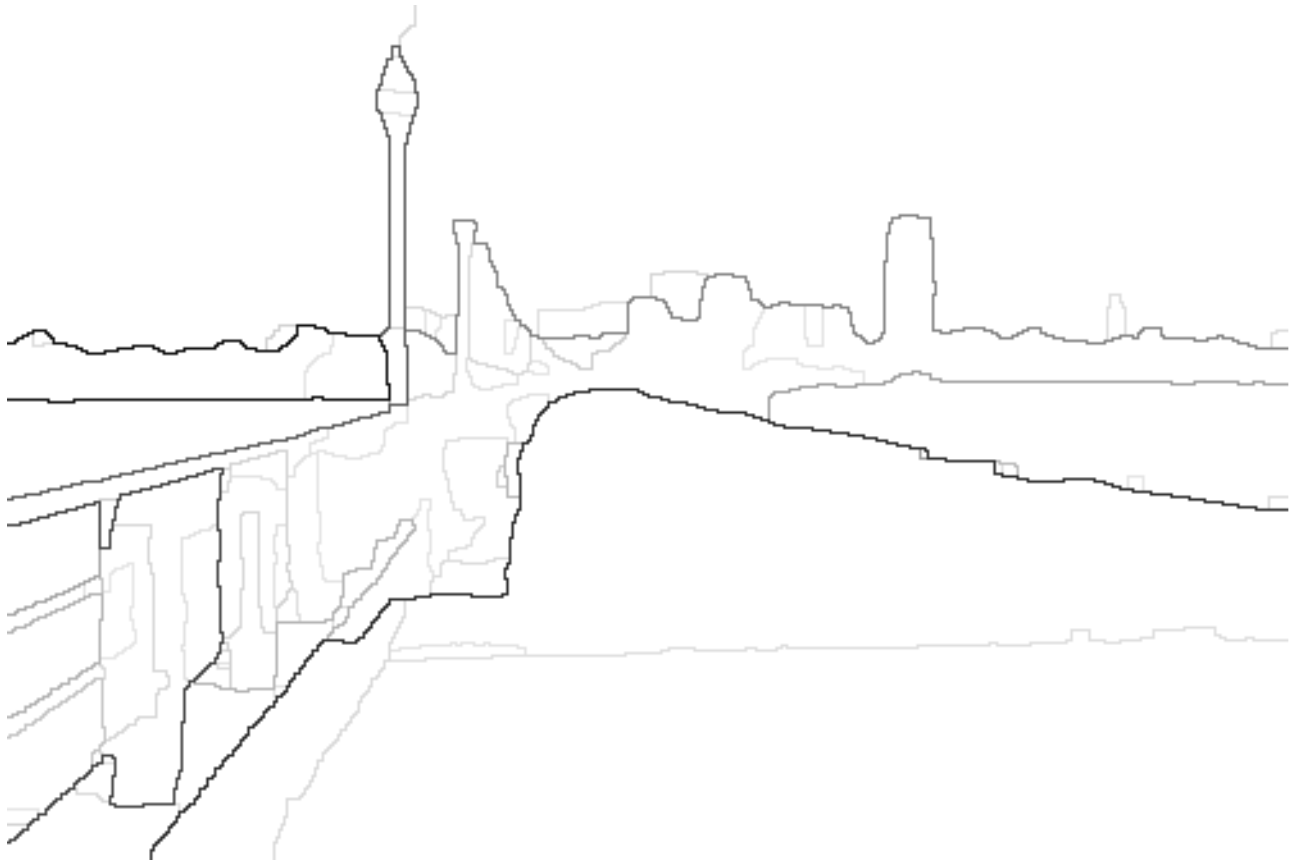}} &
\fbox{\includegraphics[height =1.2in]{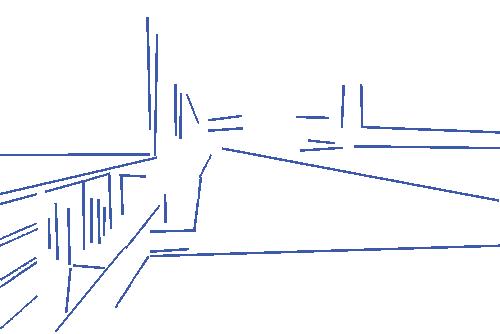}}\\
(a) & (b) & (c)
\end{tabular}
\caption{Contour-based edge detection. (a) Original image. (b) The ultrametric contour map (UCM). (c) Filtered edges.}
\label{fig:ct}
\end{figure*}

Given a set of edges $\E = \{E_1, \ldots, E_N\}$, a VP detection method aims to classify the edges into several classes, one for each VP in the scene, plus an ``outlier'' class. Similar to~\cite{Tardif09}, we employ the J-Linkage algorithm~\cite{ToldoF08} for multiple model estimation and classification. The most important new idea of our method lies in the use of contours to generate the input edges. As we will see in this section, our contour-based method can effectively identify weak edges in natural scene images and reduce the number of outliers at the same time, leading to significantly higher VP detection accuracy.

\subsection{Edge Detection via Contours}
\label{sec:edge}

Because we rely on edges to identify the dominant VP in an image, an ideal edge detection method should have the following properties: (i) it should detect all edges that converge to the true VPs, (ii) the detected edges should be as complete as possible, and (iii) it should minimize the number of irrelevant or cluttered edges. Unfortunately, local edge-detection methods do not meet these criteria. A successful method must go beyond local measurements and utilize \emph{global} visual information. 

Our key insight is that in order to determine if an edge is present at a certain location, it is necessary to examine the \emph{relevant regions} associated with it. This assertion is motivated by observing that humans label the edges by first identifying the physical objects in an image. In addition, based on the level of details they choose, different people may make different decisions on whether to label a particular edge. 

Accordingly, for edge detection, we employ the widely-used contour detection method~\cite{ArbelaezMFM11}, which proposed a unified framework for contour detection and image segmentation using an agglomerative region clustering scheme. In the following, we first discuss the main difference between the contours and edges detected by local methods before showing how to obtain straight edges from the contours.

\medskip
\noindent{\bf Globalization in contour detection.}
Comparing to the local methods, the contours detected by~\cite{ArbelaezMFM11} enjoy two levels of globalization. 

First, as a global formulation, \emph{spectral clustering} has been widely used in image segmentation to suppress noise and boost weak edges. Generally, let $W$ be an affinity matrix whose entries encode the (local) similarity between pixels, this method solves for the generalized eigenvectors of the linear system:
$(D-W)\vv = \lambda D\vv$, where the diagonal matrix $D$ is defined as $D_{ii} = \sum_j W_{ij}$. Let $\{\vv_0, \vv_1, \ldots, \vv_K\}$ be the eigenvectors corresponding the $K+1$ smallest eigenvalues $0=\lambda_0 \leq \lambda_1 \leq \cdots \leq \lambda_K$. Using all the eigenvectors except $\vv_0$, one can then represent each image pixel with a vector in $\Re^K$. As shown in~\cite{ArbelaezMFM11}, the distances between these new vectors provide a denoised version of the original affinities, making them much easier to cluster.

Second, a graph-based \emph{hierarchical clustering} algorithm is used in~\cite{ArbelaezMFM11} to construct an \emph{ultrametric contour map} (UCM) of the image (see Figure~\ref{fig:ct}(b)). The UCM defines a duality between closed, non-self-intersecting weighted contours and a hierarchy of regions, where different levels of the hierarchy correspond to different levels of detail in the image. Thus, each weighted contour in UCM represents the dissimilarity of two, possibly large, regions in the image, rather than the local contrast of small patches.

\medskip
\noindent{\bf From contours to edges.} Let $\C = \{C_1, C_2, \ldots\}$ denote the set of all weighted contours. To recover straight edges from the contour map, we apply a scale-invariant contour subdivision procedure. Specifically, for any contour $C_j$, let $\cc_j^1$ and $\cc_j^2$ be the two endpoints of $C_j$, we first find the point on $C_j$ which has the maximum distance to the straight line segment connecting its endpoints: 
\begin{equation}
\p^* = \arg\max_{\p\in C_j}dist(\p, \overline{\cc_j^1 \cc_j^2})\;.
\end{equation}
We then subdivide $C_j$ at $\p^*$ if the maximum distance is greater than a fixed fraction $\alpha$ of the contour length: 
\begin{equation}
dist(\p^*, \overline{\cc_j^1 \cc_j^2}) > \alpha \cdot |C_j|\;.
\end{equation}

By recursively applying the above procedure to all the contours, we
obtain a set of approximately straight edges $\E = \{E_1, \ldots,
E_N\}$. We only keep edges that are longer than certain threshold $l_{\min}$, because short edges are very sensitive to image noises (Figure~\ref{fig:ct}(c)).

\medskip
\noindent{\bf Quantitative evaluation.} We compare the performance of our contour-based edge detection to two popular edge detectors: the Canny detector~\cite{Canny86a} and the Line Segment Detector (LSD)~\cite{von2012lsd}. For this experiment, we have randomly chosen 100 images from our AVA landscape dataset and manually labeled all the edges that converge to the dominant VP in each image. With the ground truth edges, all methods are then evaluated quantitatively by means of the \emph{recall} and \emph{precision} as described in~\cite{MartinFM04}. Here, recall is the fraction of true edge pixels that are detected, whereas precision is the fraction of edge pixel detections that are indeed true positives ({\it i.e.}, those consistent with the dominant VP). Note that in the context of edge detection, the particular measures of recall and precision allow for some small tolerance in the localization of the edges (see~\cite{MartinFM04} for details).

\begin{figure}[t]
\centering
\begin{tabular}{cc}
\hspace{-3mm}\includegraphics[height =1.35in]{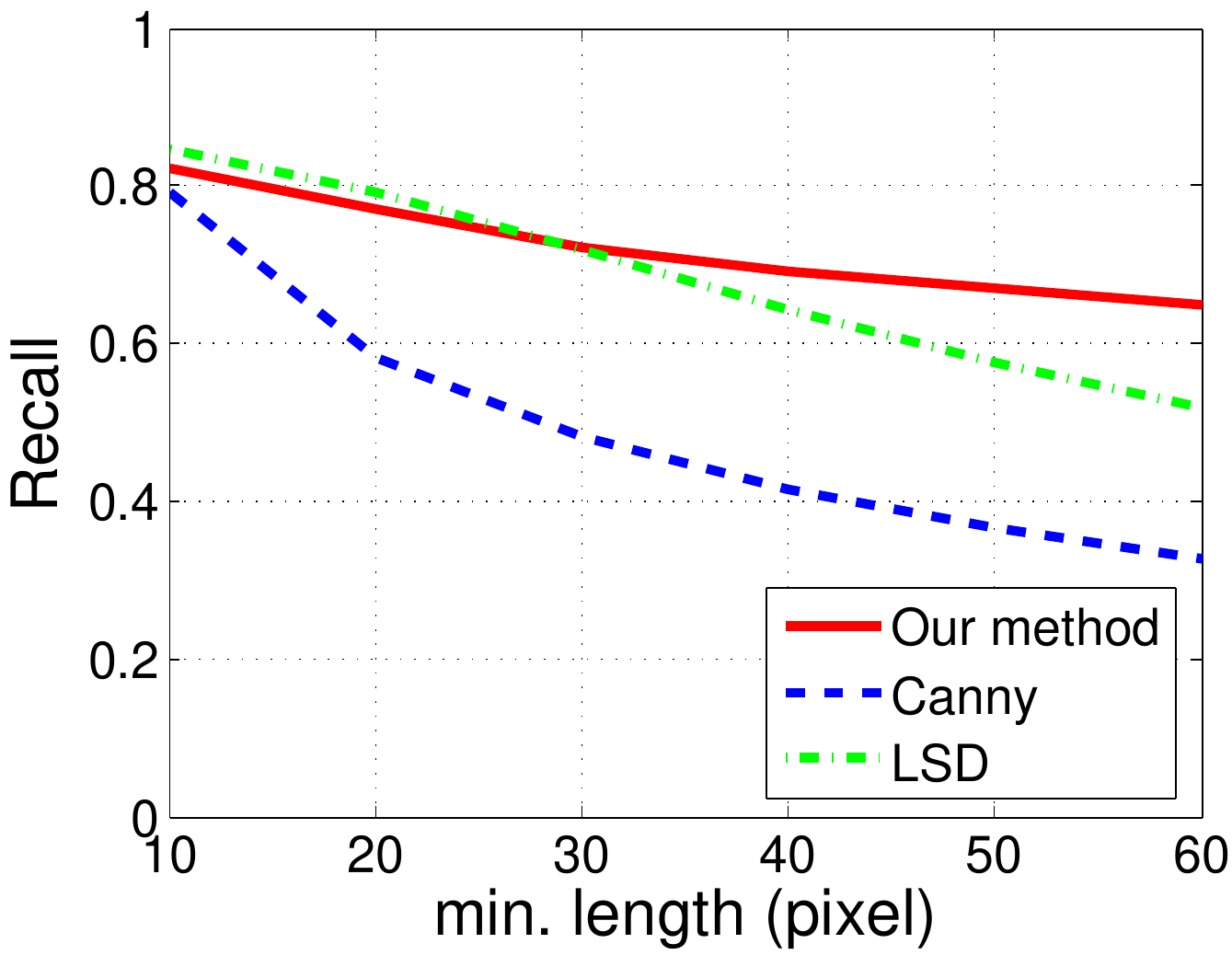} &
\hspace{-3mm}\includegraphics[height =1.35in]{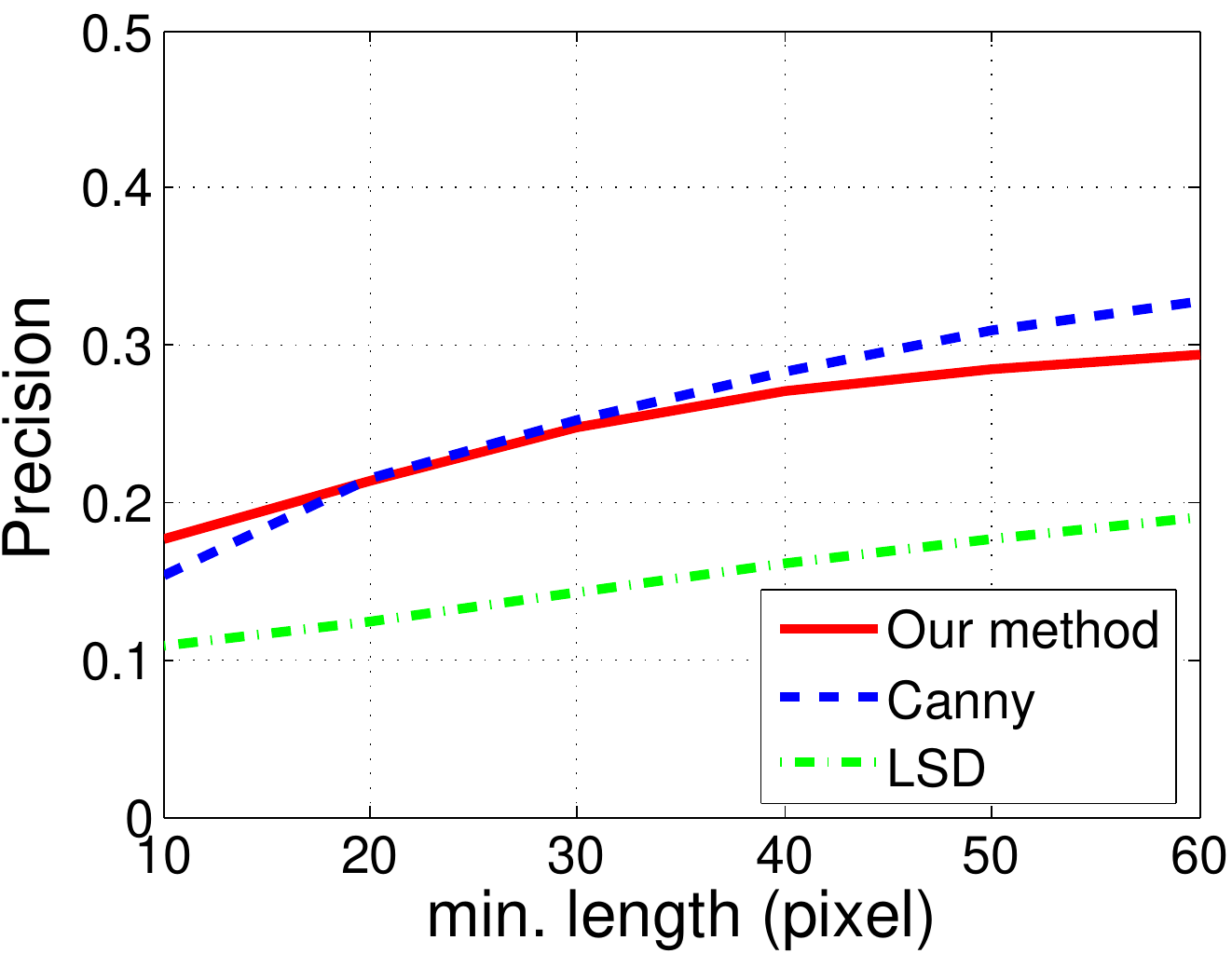} \\
(a) Recall & (b) Precision 
\end{tabular}
\caption{Quantitative evaluation of edge detection methods.}
\label{fig:edge-accuracy}
\end{figure}

In Figure~\ref{fig:edge-accuracy}, we report the recall and precision of all methods as a function of the minimum edge length $l_{\min}$. Since most VP detection methods rely on clustering the detected edges, an ideal edge detector should maximize the number of edges consistent with the ground truth dominant VP and \emph{and} minimize the number of irrelevant edges. Thus, higher recall and precision indicate better performance. As shown in Figure~\ref{fig:edge-accuracy}, compared to LSD, our method achieves comparable recall but with much higher precision. Meanwhile, while Canny has similar precision as our method, its recall is substantially lower. This result is expected as Canny operates on pixels with high gradient values, but such pixels are often absent in natural scenes ({\it i.e.}, low recall). While LSD is able to overcome this difficulty and achieve higher recall by computing the statistics in surrounding rectangular regions instead, the ambiguity in the region size often leads to repeated detections ({\it i.e.}, low precision).
Thus, our contour-based method is an overall better choice for the vanishing point detection task.

\fboxrule=1pt

\begin{figure*}[t!]
\centering
\includegraphics[height =0.92in]{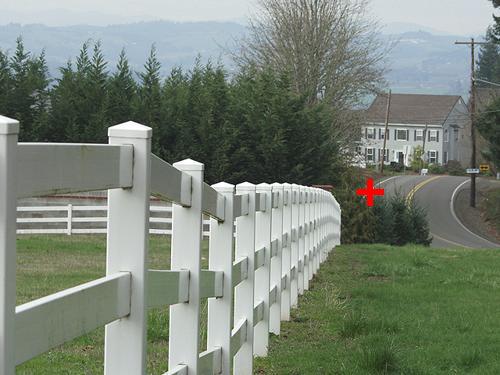}
\includegraphics[height =0.92in]{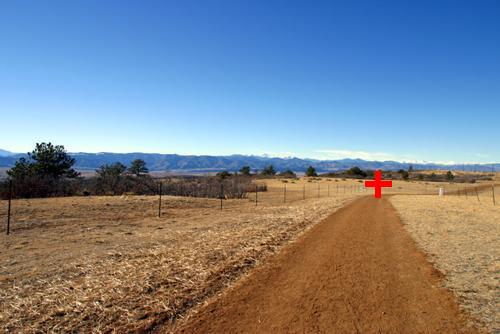}
\includegraphics[height =0.92in]{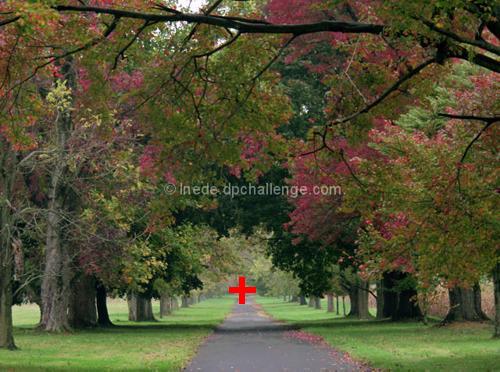}
\includegraphics[height =0.92in]{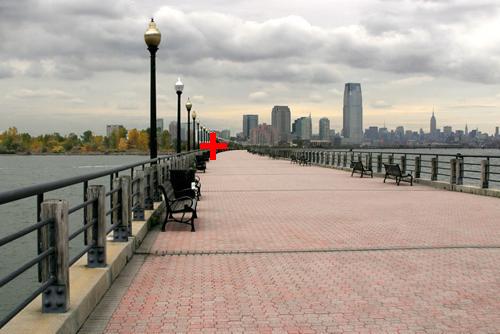}
\includegraphics[height =0.92in]{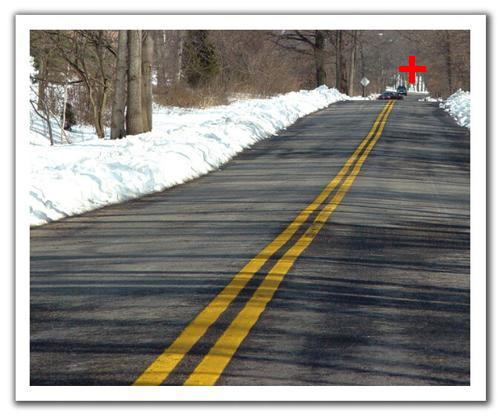}
\\
\vspace{2mm}
\fbox{\includegraphics[height =0.9in]{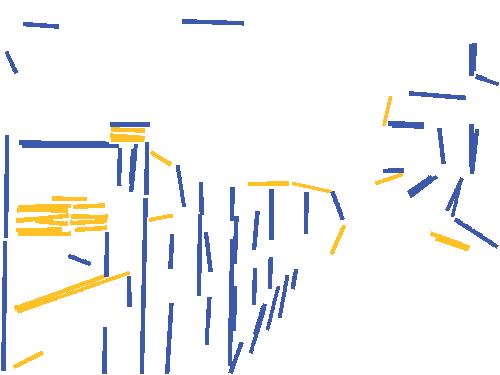}}
\fbox{\includegraphics[height =0.9in]{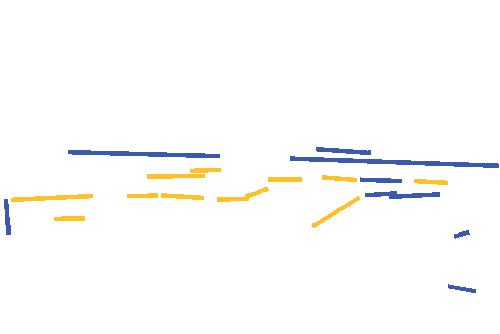}}
\fbox{\includegraphics[height =0.9in]{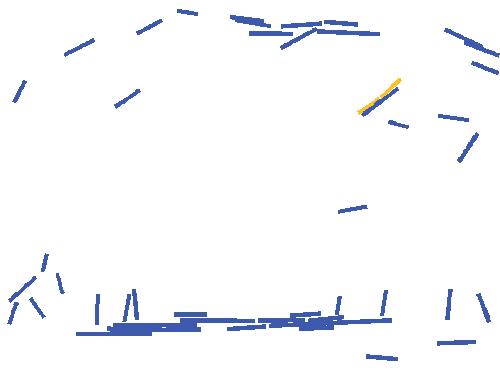}}
\fbox{\includegraphics[height =0.9in]{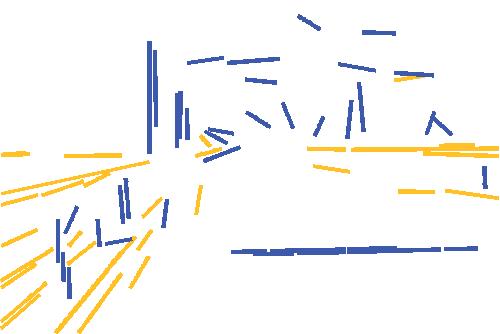}}
\fbox{\includegraphics[height =0.9in]{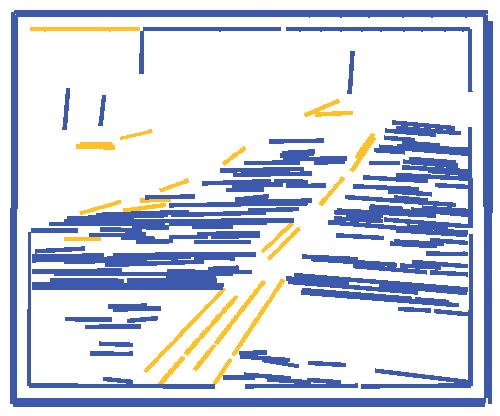}}
\\
\vspace{2mm}
\fbox{\includegraphics[height =0.9in]{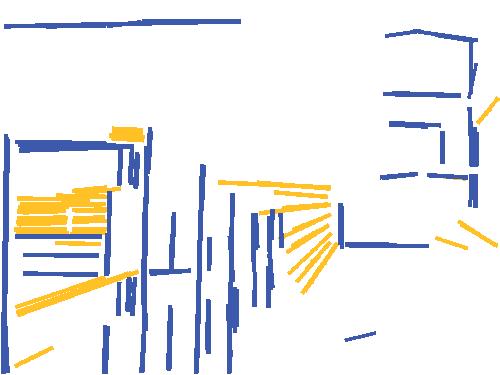}}
\fbox{\includegraphics[height =0.9in]{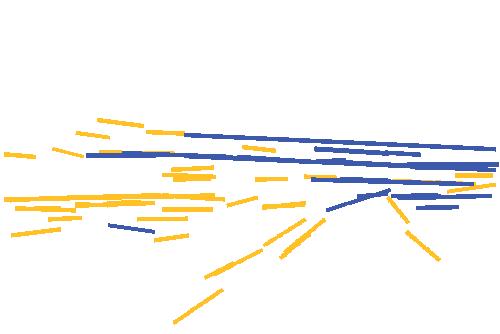}}
\fbox{\includegraphics[height =0.9in]{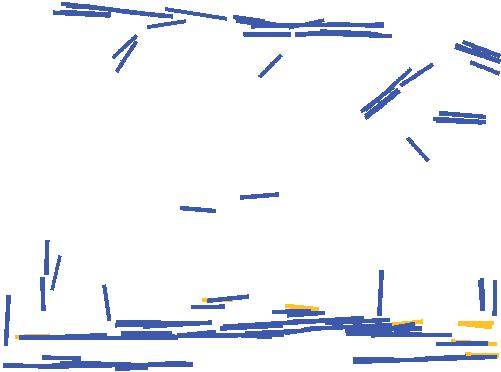}}
\fbox{\includegraphics[height =0.9in]{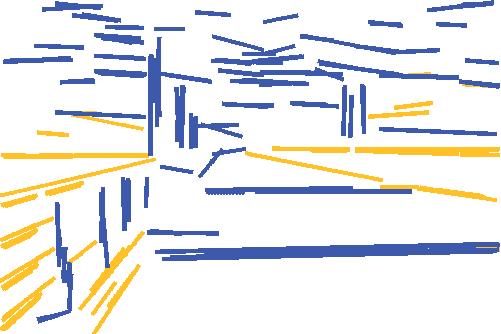}}
\fbox{\includegraphics[height =0.9in]{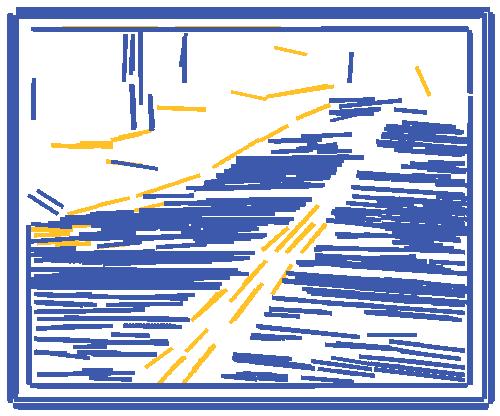}}
\\
\vspace{2mm}
\fbox{\includegraphics[height =0.9in]{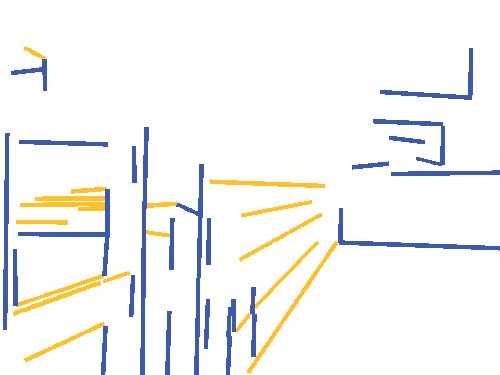}}
\fbox{\includegraphics[height =0.9in]{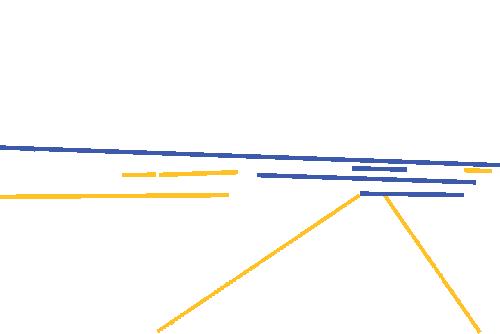}}
\fbox{\includegraphics[height =0.9in]{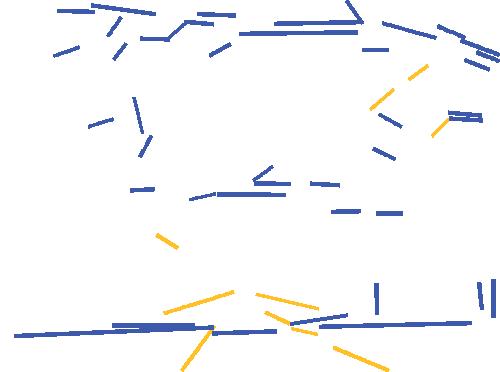}}
\fbox{\includegraphics[height =0.9in]{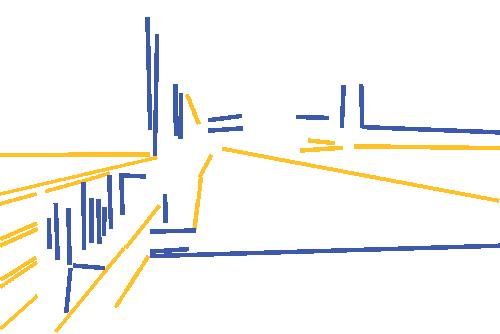}}
\fbox{\includegraphics[height =0.9in]{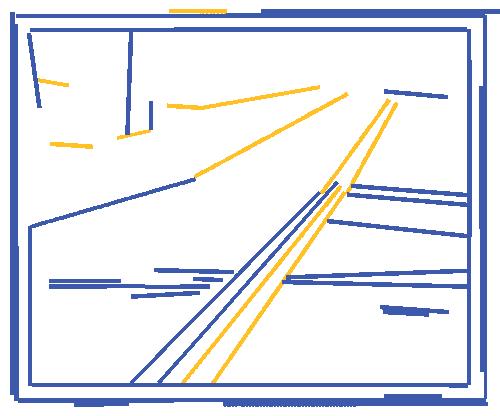}}
\\
(a) \hspace{28mm} (b) \hspace{28mm} (c) \hspace{28mm} (d) \hspace{26mm} (e)
\caption{Example detections of different edge detection methods. The four rows show the original images, and the edges detected by Canny detector, LSD, and our contour-based method, respectively. Yellow edges indicate the edges consistent with the ground truth dominant VP.}
\label{fig:edge}
\end{figure*}

In Figure~\ref{fig:edge}, we further show some example edge detection results. As shown, our contour-based method can better detect weak yet important edges in terms of both the quantity and the completeness. For example, our method is able to detect the complete edges of the road in Figure~\ref{fig:edge}(b), while the local methods only detected parts of them. Also, only our method successfully detected the edges of the road in Figure~\ref{fig:edge}(c). 

Another important distinction between our contour-based method and the local methods concerns the textured areas in the image. Local methods tend to confuse image texture with true edges, resulting in a large number of detections in these areas (\eg, the sky region and the road in \ref{fig:edge}(d) and (e), respectively). Such false positives often lead to incorrect clustering results in the subsequent VP detection stage. Meanwhile, our method treats the textured area as a whole, thereby greatly reducing the number of false positives. 

\subsection{J-Linkage}

In this section, we give an overview of the J-Linkage algorithm~\cite{ToldoF08}. Similar to RANSAC, J-Linkage first randomly chooses $M$ minimal sample sets and computes a putative model for each of them. For VP detection, the $j$-th minimal set consists of two randomly chosen edges: $(E_{j_1}, E_{j_2})$. To this end, we first fit a line $\ll_i$ to each edge $E_j\in \E$ using least squares. Then, we can generate the hypothesis $\vv_j$ using the corresponding fitted lines: $\vv_j = \ll_{j_1} \times \ll_{j_2}$\;.

Next, J-Linkage constructs a $N\times M$ preference matrix $P$, where the $(i,j)$-th entry is defined as:
\begin{equation}
p_{ij} = \left\{ \begin{array}{ll} 1 & \textup{if } D(E_i, \vv_j) \leq \phi \\ 0 & \textup{otherwise} \end{array} \right.\;.
\label{eq:consistency}
\end{equation}
Here, $D(E_i, \vv_j)$ is a measure of consistency between edge $E_i$ and VP hypothesis $\vv_j$, and $\phi$ is a threshold. Note that $i$-th row indicates the set of hypotheses edge $E_i$ has given consensus to, and is called the \emph{preference set} (PS) of $E_i$. J-Linkage then uses a bottom-up scheme to iteratively group edges that have similar PS. Here, the PS of a cluster is defined as the intersection of the preference sets of its members. In each iteration, the two clusters with the smallest distance are merged, where the Jaccard distance is used to measure the similarity between any two clusters $A$ and $B$:
\begin{equation}
d_J(A,B) = \frac{|A\bigcup B| - |A \bigcap B|}{|A\bigcup B|}\;.
\end{equation}
The operation is repeated until the distance between any two clusters is 1.

\smallskip
\noindent{\bf Consistency measure.} We intuitively define the consistency measure $D(E_i, \vv_j)$ as the root mean square (RMS) distance from all points on $E_i$ to a line $\hat{\ll}$, such that $\hat{\ll}$ passes through $\vv_j$ and minimizes the distance:
\begin{equation}
D_{RMS}(E_i, \vv_j) = \min_{l: l\times \vv_j = 0} \left(\frac{1}{N} \sum_{\p \in E_i} dist(\p, \ll)^2\right )^\frac{1}{2}\;,
\end{equation}
where $N$ is the number of points on $E_i$, and $dist(\p, \l)$ is the perpendicular distance from a point $\p$ to a line $\l$, \ie, the length of the line segment which joins $\p$ to $\l$ and is perpendicular to $\l$.

\subsection{Experiments}
\label{sec:detection-exp}

In this section, we present a comprehensive performance study of our contour-based VP detection method and compare it to the state-of-the-art methods. We use both the AVA landscape dataset and the Flickr dataset described in Section~\ref{sec:dataset}. Similar to previous work (\eg, \cite{Tardif09, WildenauerH12}), we evaluate the performance of a VP detection method based on the consistency of the ground truth edges with the estimated VPs. Specifically, let $\{E_k^G\}_{k=1}^K$ be the set of ground truth edges, the consistency error of a detection $\hat{\vv}$ is:
\begin{equation}
err(\hat{\vv}) = \frac{1}{K} \sum_k D_{RMS}(E_k^G, \hat{\vv})\;.
\label{eq:evaluation}
\end{equation}
For all experiments, we compute the average consistency error over five independent trials. 

\subsubsection{Dominant Vanishing Point Detection Results} 
\begin{figure}[t]
\centering
\begin{tabular}{cc}
\hspace{-2mm}\includegraphics[height =1.4in]{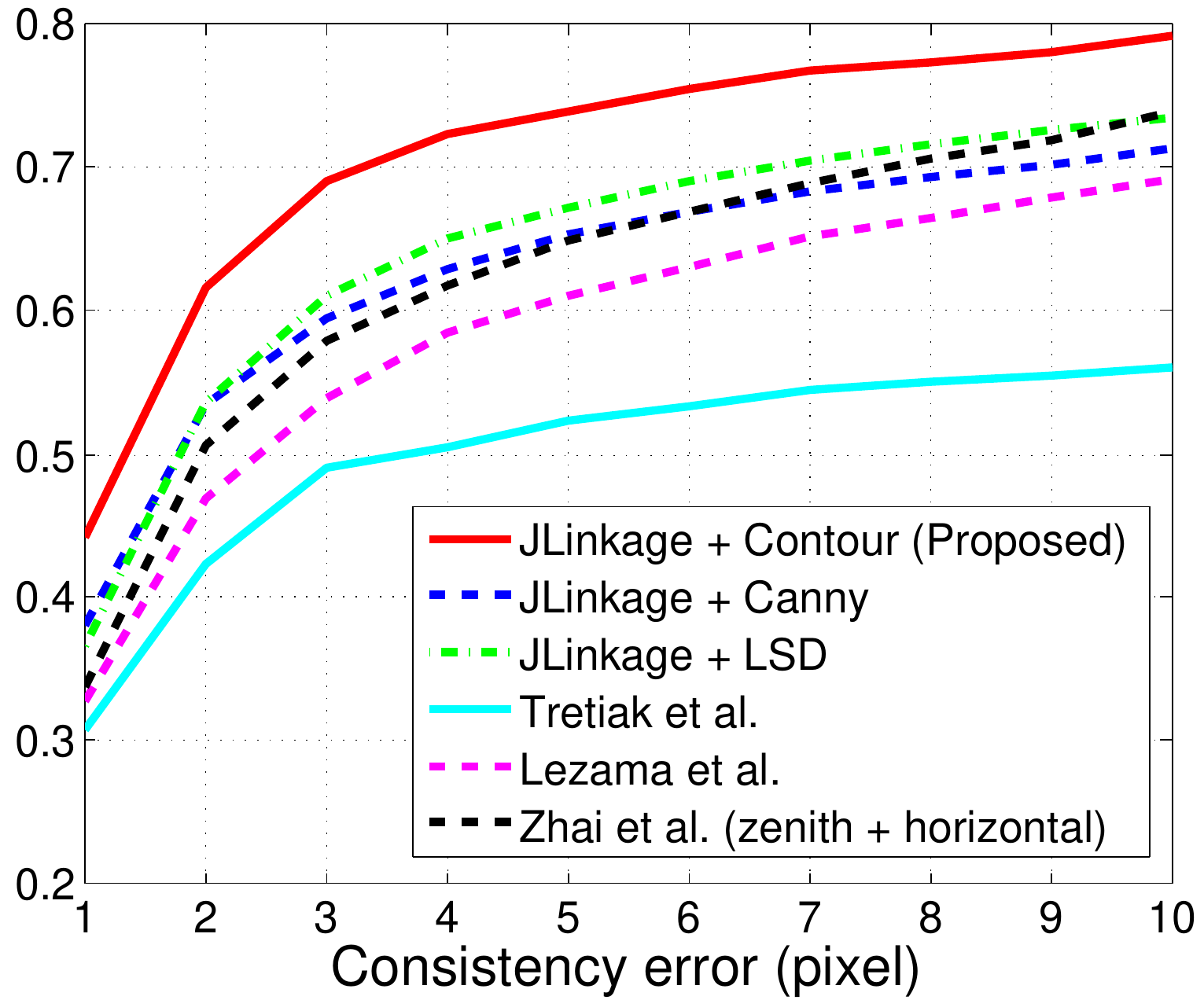} &
\hspace{-2mm}\includegraphics[height =1.4in]{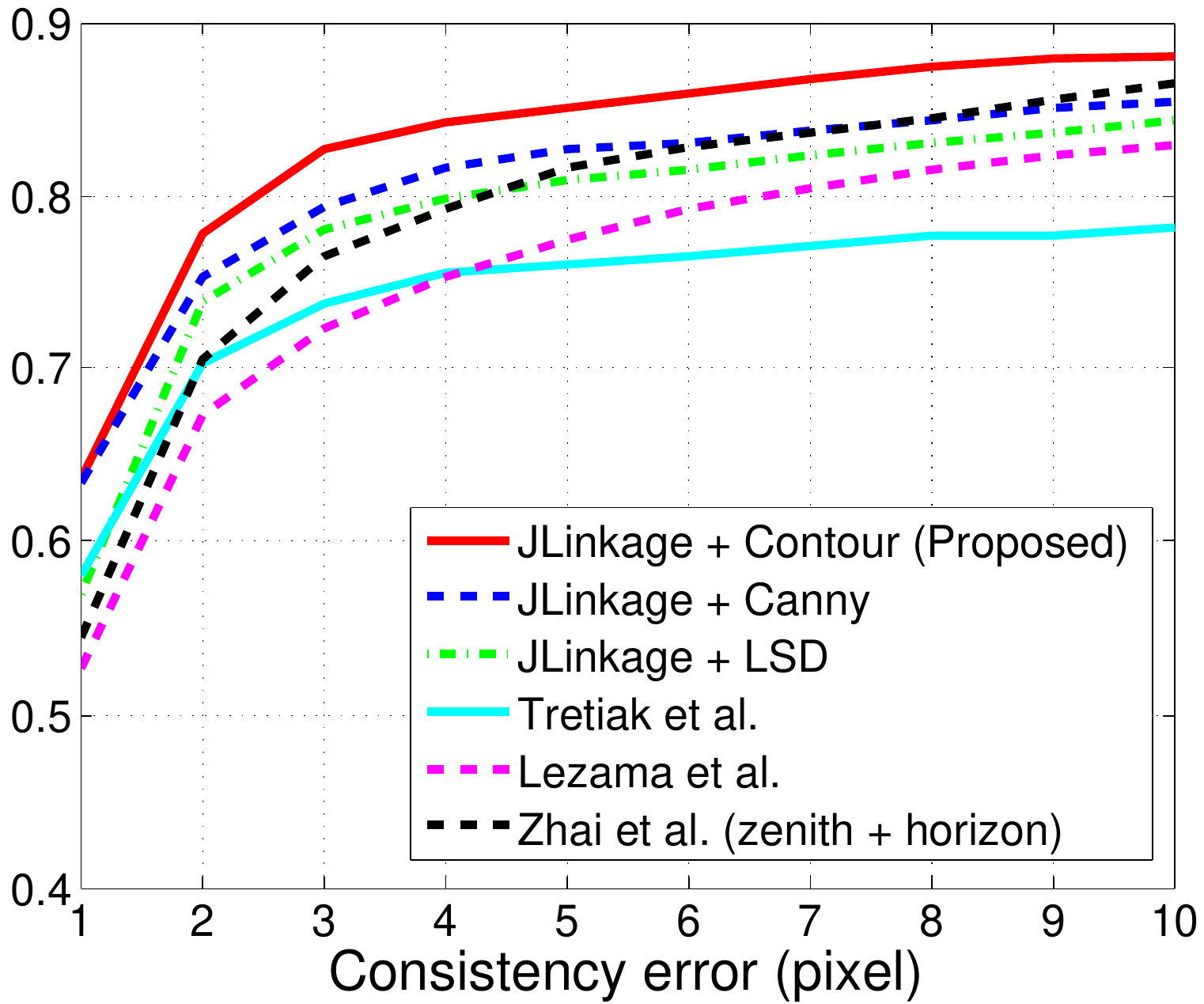} \\
(a) AVA Landscape & (b) Flickr 
\end{tabular}
\caption{Dominant vanishing point detection results. We show the cumulative histograms of vanishing point consistency error w.r.t. the ground truth edges (Eq.~\eqref{eq:evaluation}) for all candidate methods.}
\label{fig:result-detection}
\end{figure}

Figure~\ref{fig:result-detection} reports the cumulative histograms of vanishing point consistency error w.r.t. the ground truth edges for various methods. Below we analyze the results in detail.

\noindent{\bf Comparison of edge detection methods.} We first compare our method to Canny detector~\cite{Canny86a} and LSD~\cite{von2012lsd} in terms of the accuracy of the detected VPs. For our contour-based method, the parameters are: $\alpha = 0.05$, $l_{\min} = 40$, and $\phi = 3$. For Canny detector and LSD, we tune the parameters $l_{\min}$ and $\phi$ so that the highest accuracy is obtained. In this experiment, we simply keep the VP with the largest support set as the detection result. Our contour-based method significantly outperforms the other edge detection methods.

\noindent{\bf Comparison with the state-of-the-art.} We also compare our method to state-of-the-art VP detection methods. As we discussed before, most existing methods focus on urban scenes and make strong assumptions about the scene structures, such as a Manhattan world model~\cite{MirzaeiR11, BazinSDVIKP12, WildenauerH12, AntunesB13}. Such strong assumptions render these methods inapplicable to natural landscape scenes. 

\begin{figure*}[t!]
\centering
\begin{tabular}{ccc}
\hspace{-1mm}\includegraphics[height =1.6in]{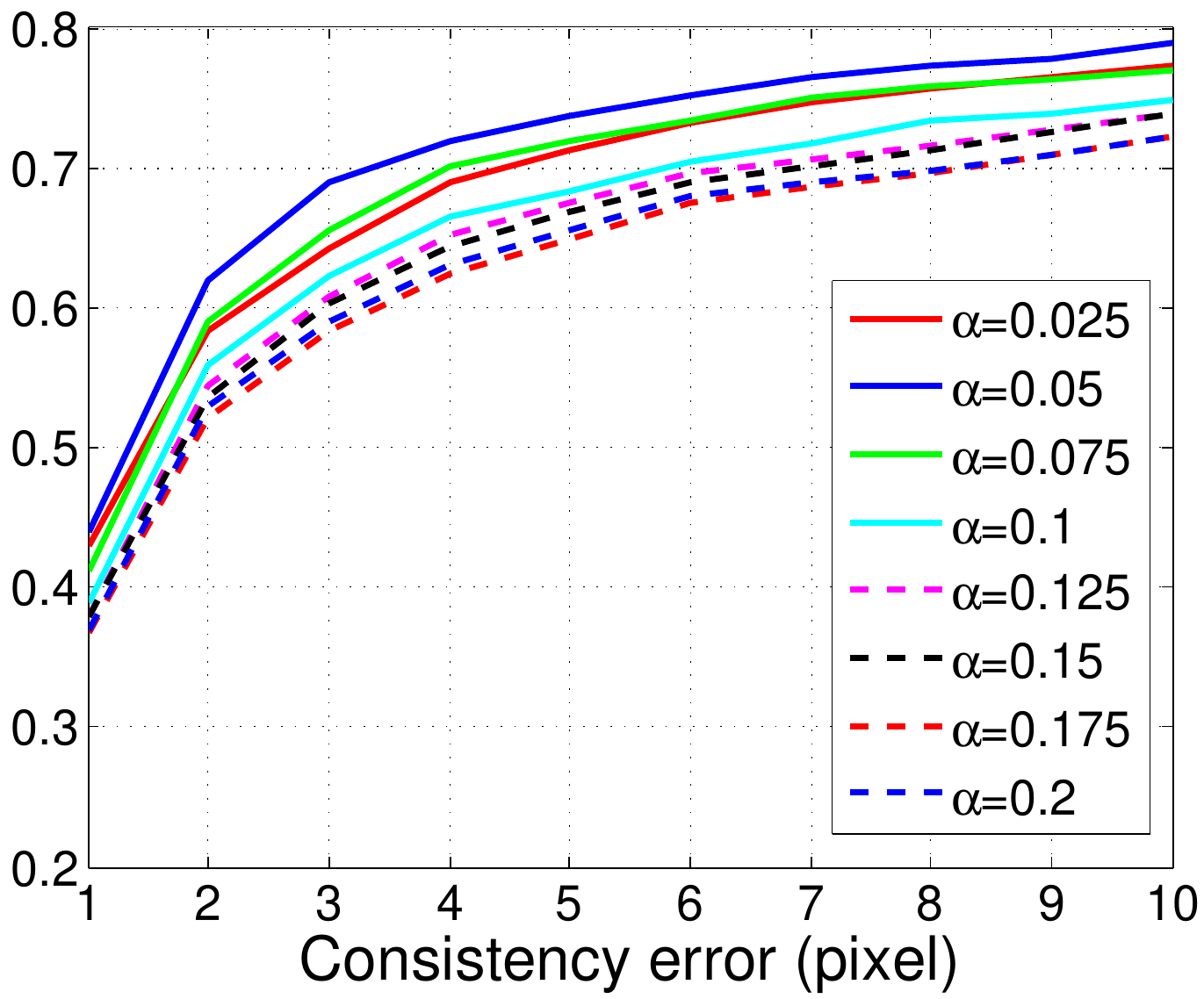} &
\hspace{-1mm}\includegraphics[height =1.6in]{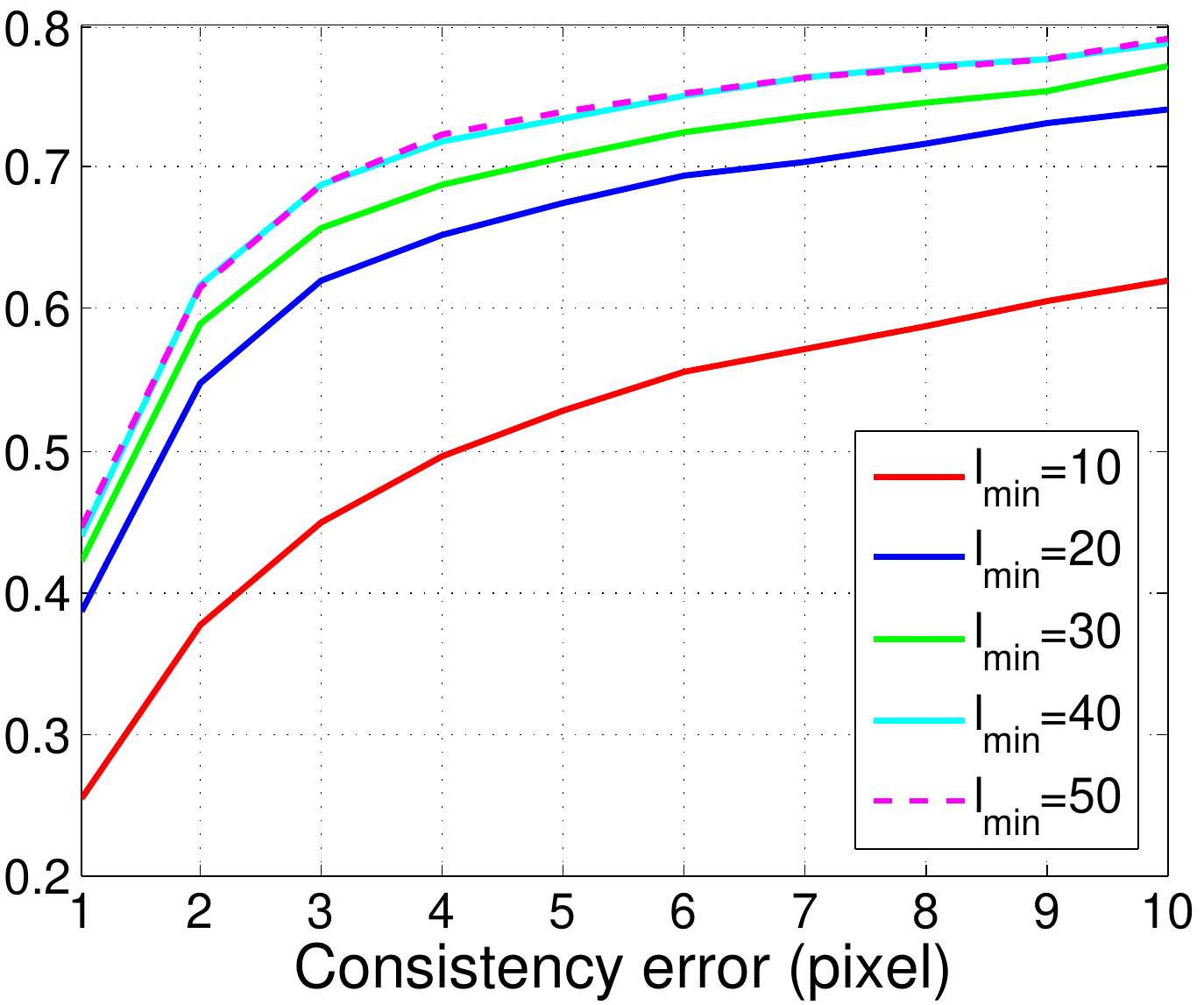} &
\hspace{-1mm}\includegraphics[height =1.6in]{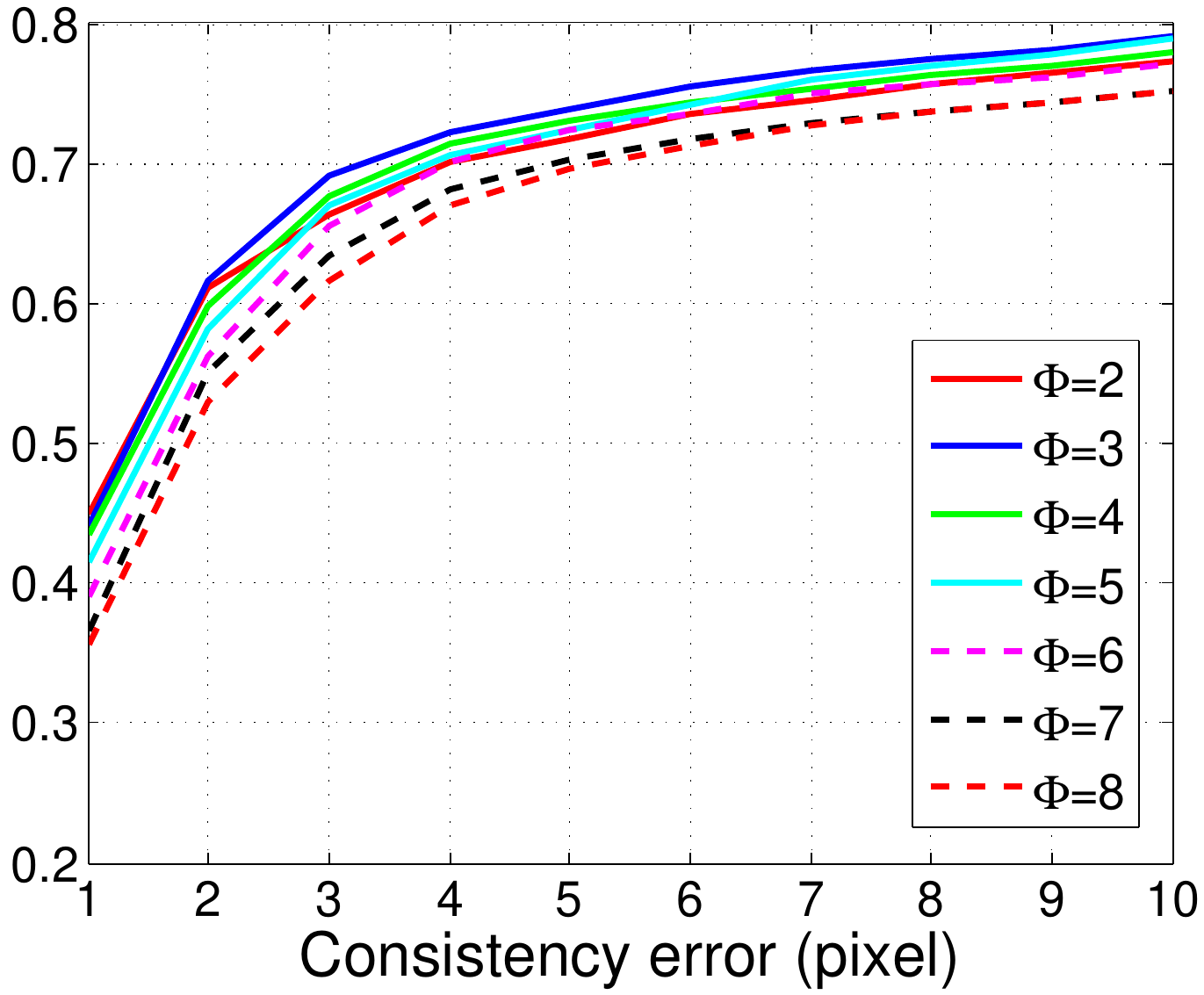}\\
\hspace{-1mm}\includegraphics[height =1.65in]{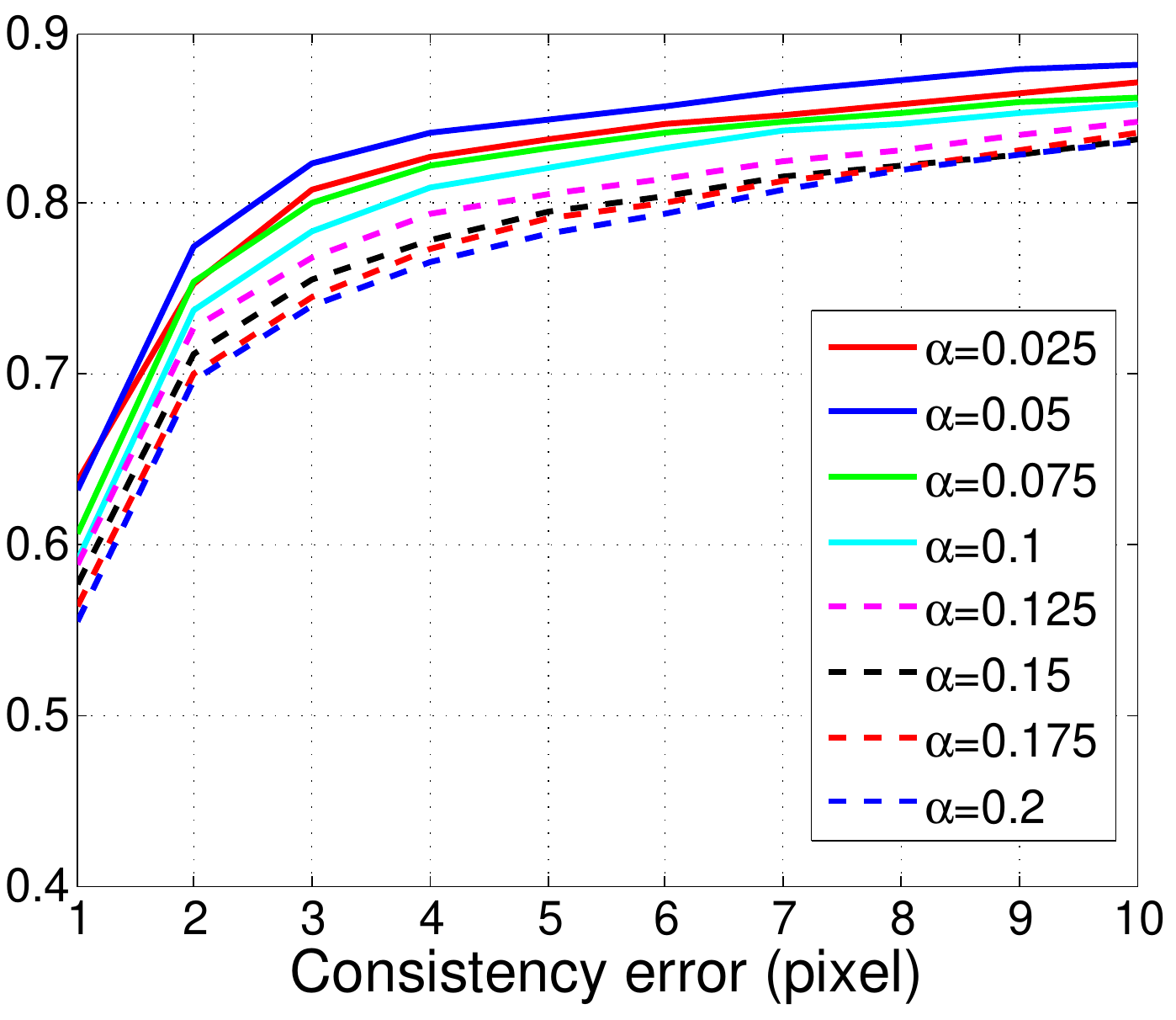} &
\hspace{-1mm}\includegraphics[height =1.65in]{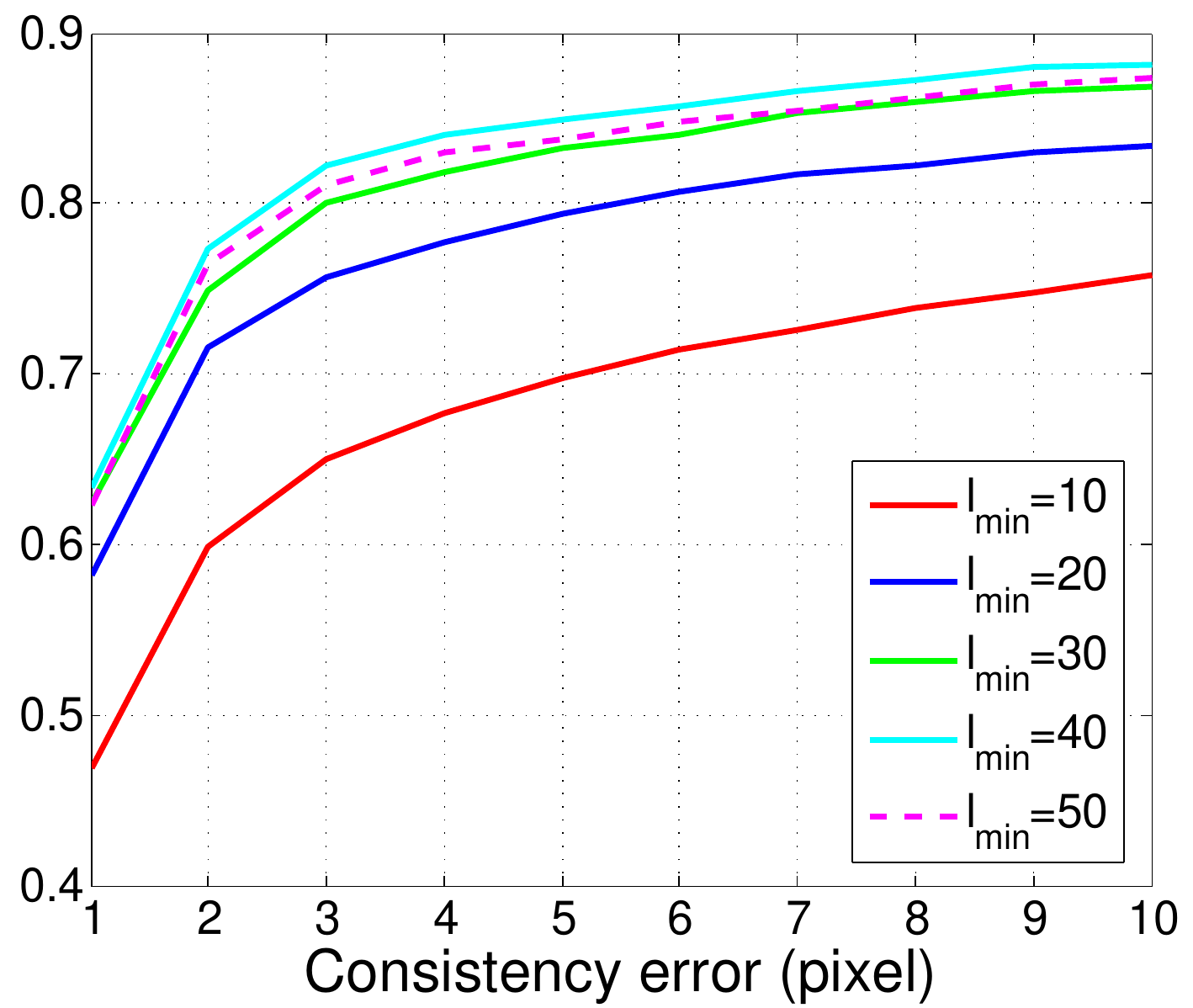} &
\hspace{-1mm}\includegraphics[height =1.65in]{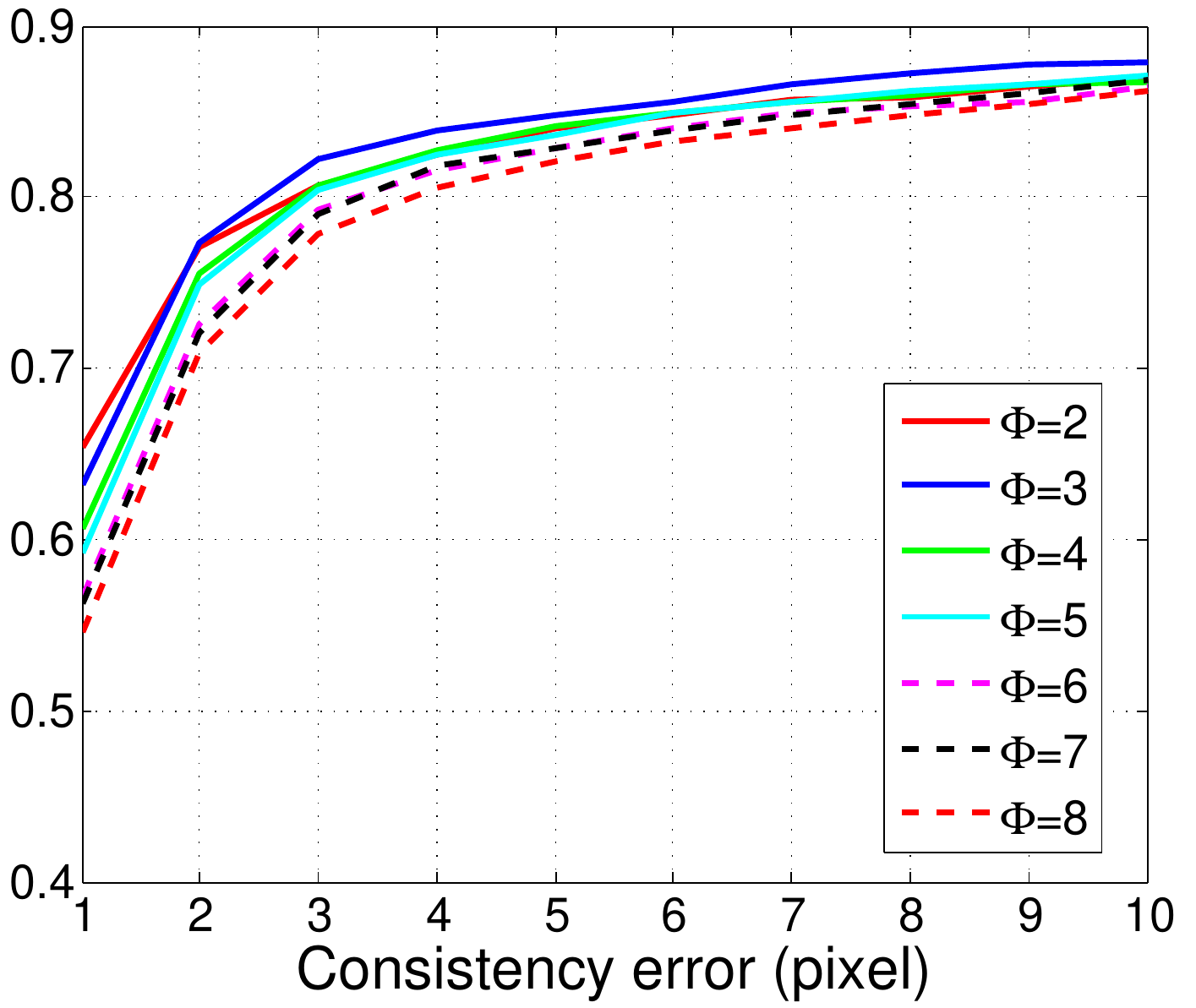}\\
(a) $\alpha$ & (b) $l_{\min}$ & (c) $\phi$
\end{tabular}
\caption{Accuracy of our method w.r.t. parameters. {\bf First row:} AVA landscape dataset. {\bf Second row:} Flickr dataset.}
\label{fig:parameter}
\end{figure*}

While other methods do not explicitly assume a specific model, they still benefit from the scene structures to various extents. In Figure~\ref{fig:result-detection}, we compare our method to three recent methods, namely Tretiak~\etal~\cite{TretiakBKL12}, Lezama~\etal~\cite{LezamaGRM14}, and Zhai~\etal~\cite{ZhaiWJ16}. For each method, we use the source code downloaded from its author's website. Note that~\cite{LezamaGRM14} uses the Number of False Alarms (NFA) to measure the importance of the VPs. For fair comparison, we keep the VP with the highest NFA. Further, \cite{ZhaiWJ16} takes a two-step approach which first detects one zenith VP and then detects the horizontal VPs. As it is unclear from~\cite{ZhaiWJ16} how to compare the strength of zenith VP with the horizontal VPs, we choose to favor~\cite{ZhaiWJ16} in our experiment setting by considering both the zenith VP and the top-ranked horizontal VP in the evaluation and keeping the VP which achieves the smallest consistency error w.r.t. the ground truth edges.

Figure~\ref{fig:result-detection} shows that the three methods do not perform well on the natural landscape images. The problem is that~\cite{TretiakBKL12} assumes multiple horizontal VP detections for horizon and zenith estimation, whereas~\cite{ZhaiWJ16} attempts to detect both zenith and horizontal VPs. However, there may not be more than one VP in natural scenes. Similarly, \cite{LezamaGRM14} relies on the multiple horizontal VP detections to filter redundant and spurious VPs. 

\subsubsection{Parameter Sensitivity}

Next, we study the performance of our contour-based VP detection method w.r.t. the parameters $\alpha$, the minimum edge length $l_{\min}$, and the distance threshold $\phi$ in Eq.~\eqref{eq:consistency}. We conduct experiments with one of these parameters varying while the others are fixed. The default parameter setting is $\alpha = 0.05$, $l_{\min} = 40$, and $\phi = 3$. 

\smallskip
\noindent{\bf Performance w.r.t. $\alpha$.} Recall from Section~\ref{sec:edge} that $\alpha$ controls the degree to which a contour segment may deviate from a straight line before it is divided into two sub-segments. Figure~\ref{fig:parameter}(a) shows that the best performance is achieved with $\alpha=0.05$.

\smallskip
\noindent{\bf Performance w.r.t. minimum edge length $l_{\min}$.} Figure~\ref{fig:parameter}(b) shows the performance of our method as a function of $l_{\min}$. Rather surprisingly, we find that the accuracy is quite sensitive to $l_{\min}$. This finding probably reflects the relatively small number of edges consistent with the dominant VP available in natural scenes. Therefore, if $l_{\min}$ is too small, these edges may be dominated by irrelevant edges in the scene; if $l_{\min}$ is too large, there may not be enough inliers to robustly estimate the VP location.

\smallskip
\noindent{\bf Performance w.r.t. threshold $\phi$.}  Figure~\ref{fig:parameter}(c) shows the accuracy of our method w.r.t. the threshold $\phi$ in Eq.~\eqref{eq:consistency}. As shown, our method is relatively insensitive to the threshold, and achieves the best performance when $\phi = 3$.

\smallskip
Finally, we note that the experiment results are very consistent across the two different datasets. This finding suggests that the choice of the parameters is insensitive to the datasets and can be well extrapolated to general images.

%% file: 5vpselection.tex
\section{Selection of the Dominant Vanishing Point}
\label{sec:selection}

In real-world applications concerning natural scene photos, it is often necessary to select the images in which a dominant VP is present since many images do not have a VP. Further, if multiple VPs are detected, we need to determine which one carries the most importance in terms of the photo composition. Therefore, given a set of candidates $\{\vv_j\}_{j=1}^n$ generated by a VP detection method, our goal is to find a function $f$ which well estimates the strength of a VP candidate. Then, we can define the dominant VP of an image as the one whose strength is (i) the highest among all candidates, and (ii) higher than certain threshold $T$:
\begin{equation}
\vv^* = \arg\max_{f(\vv_j) \geq T} f(\vv_j)\;.
\end{equation}

In practice, given a detected VP $\vv_j$ and the edges $\E_j \subseteq \E$ associated with the cluster obtained by a clustering method (\eg, J-Linkage), a simple implementation of $f$ would be the number of edges:
$f(\vv_j) = |\E_j|$. Note that it treats all edges in $\E_j$ equally. However, we have found this approach problematic for natural images because it does not consider the implied depth of each edge in the 3D space.

\subsection{The Strength Measure}

Intuitively, an edge conveys a strong sense of depth to the viewers if (i) it is long, and (ii) it is close to the VP (Figure~\ref{fig:example}). This observation motivates us to examine the implied depth of each individual point on an edge, instead of treating the edge as a whole. 

\begin{figure}[ht!]
\centering
\includegraphics[height =1.4in]{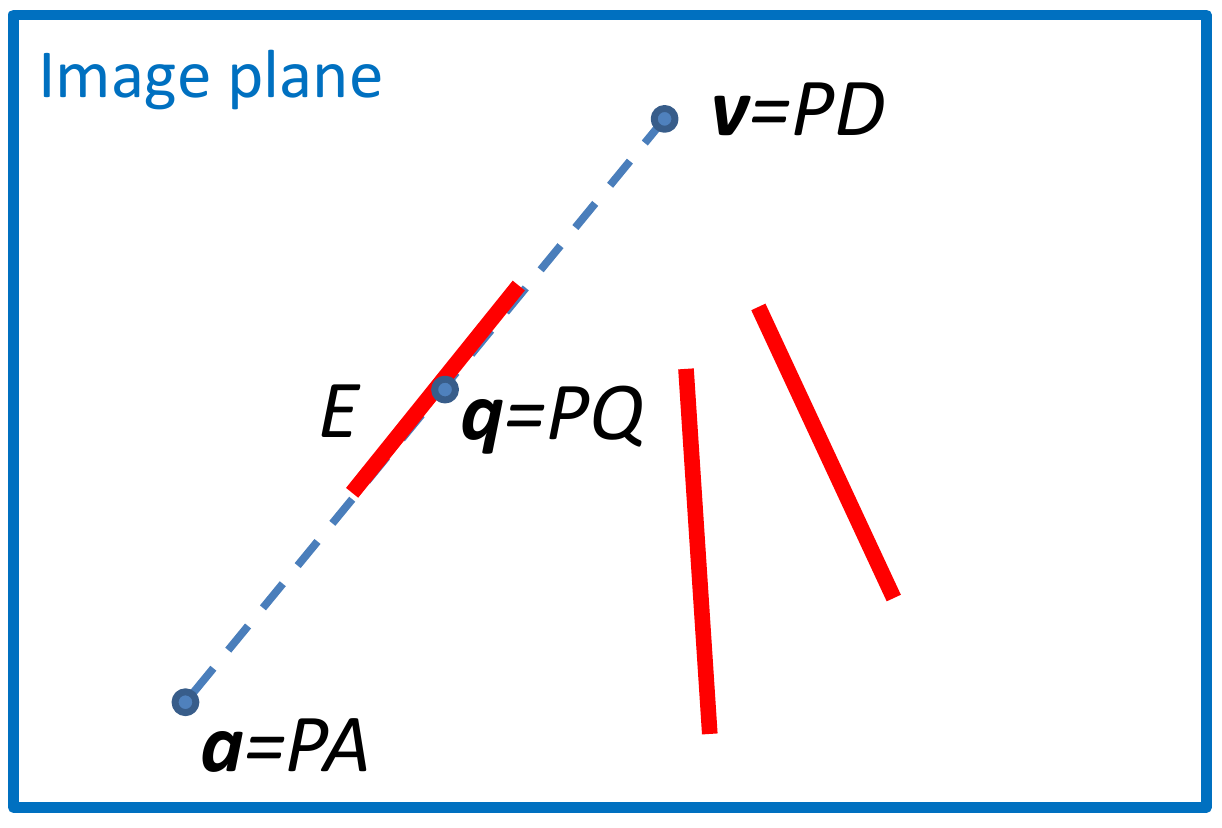}
\caption{Illustration of our edge strength measure.}
\label{fig:measure}
\end{figure}

Geometrically, as shown in Figure~\ref{fig:measure}, let $E$ be a line segment consistent with vanishing point $\vv=(v_x,v_y,1)^T$ in the image.\footnote{In this section, all 2D and 3D points are represented in homogeneous coordinates.} We further let $D$ be the direction in 3D space (\ie, a point at infinity) that corresponds to $\vv$: $\vv = P D$, where $P\in \Re^{3\times 4}$ is the camera projection matrix.

For any pixel on the line segment $\q=(q_x,q_y,1)^T \in E$, we denote $Q$ as the corresponding point in the 3D space. Then, we can represent $Q$ as a point on a 3D line with direction $D$:
$Q = A + \lambda D$, 
where $A$ is some reference point chosen on this line, and $\lambda$ can be regarded as the (relative) distance between $A$ and $Q$. Consequently, we have
\begin{equation}
\q = P Q = P (A + \lambda D) = \aa + \lambda \vv\;,
\end{equation}
where $\aa=(a_x,a_y,1)^T$ is the image of $A$.
Thus, let $l_{\q}$ and $l_{\aa}$ denote the distance on the image from $\q$ and $\aa$ to $\vv$, respectively, we have
\begin{equation}
\lambda = l_{\aa}/l_{\q} - 1\;.
\label{eq:lambda}
\end{equation}

Note that if we choose $A$ as the intersecting point of the 3D line corresponding to $E$ and the image plane, $\lambda$ represents the (relative) distance from any point $Q$ on this line to the image plane along direction $D$. In practice, although $l_{\aa}$ is typically unknown and varies for each edge $E$, we can still infer from Eq.~\eqref{eq:lambda} that $\lambda$ is a linear function of $1/l_{\q}$. This motivates us to define the weight of a pixel $\q \in E$ as $(l_{\q}+\tau)^{-1}$, where $\tau$ is a constant chosen to make it robust to noises and outliers.

Thus, our new measure of strength for $\vv_j$ is defined as
\begin{equation}
f(\vv_j) = \sum_{E\in \E_j} \sum_{\q\in E} \frac{1}{l_{\q} + \tau}\;.
\label{eq:strength}
\end{equation}
Clearly, edges that are longer and closer to the VP have more weights according to our new measure.

\subsection{Experiments}\label{sec:5experiments}

\begin{figure}[ht!]
\centering
\includegraphics[height =2in]{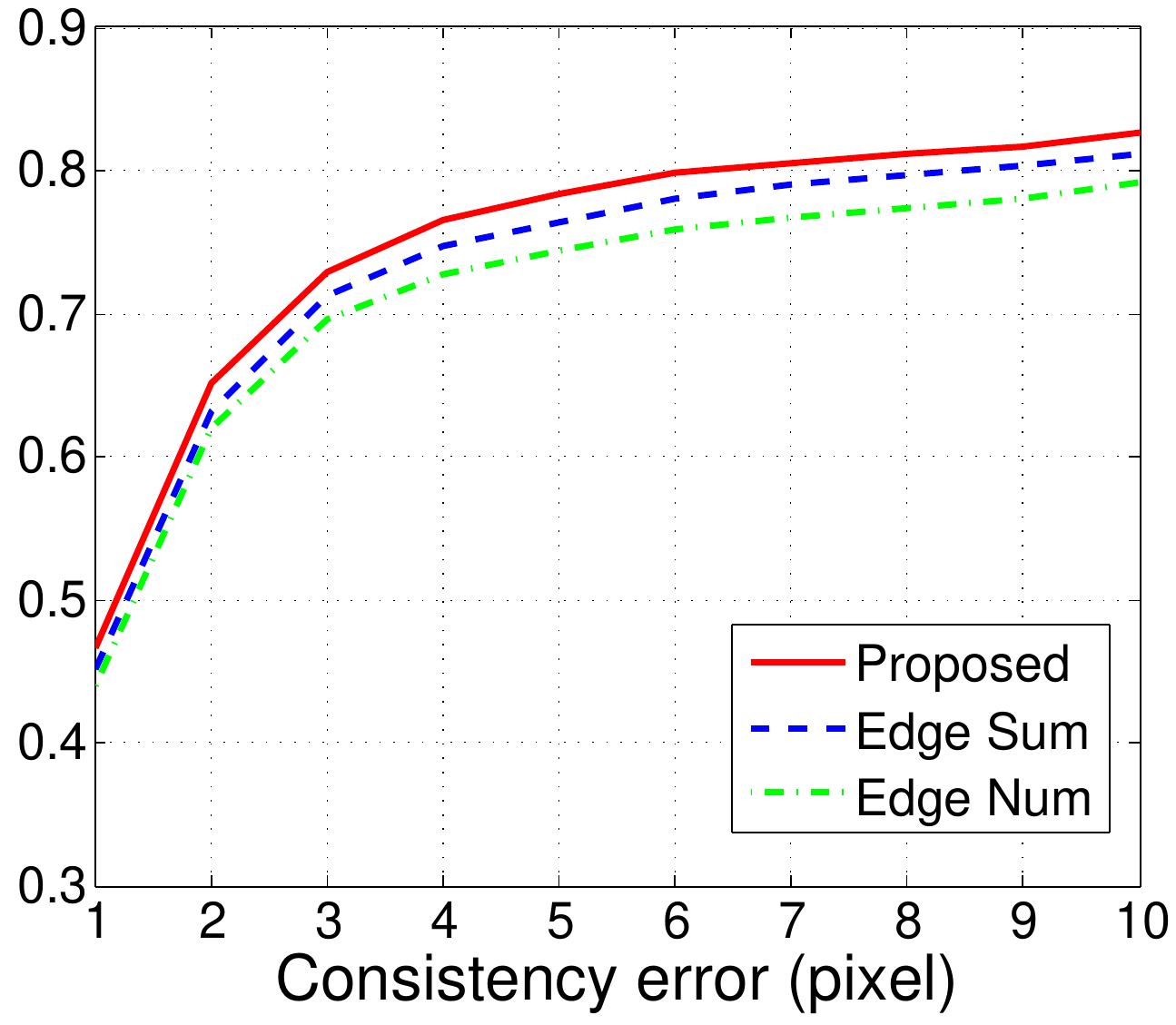}
\caption{Experiment results on
dominant vanishing point selection.}
\label{fig:selection}
\end{figure}

\noindent{\bf Dominant vanishing point selection.}
We first demonstrate the effectiveness of the proposed strength measure in selecting the dominant VP from the candidates obtained by our VP detection algorithm. In Figure~\ref{fig:selection}, we compare the following three measures in terms of the consistency error of the selected dominant VP:

\smallskip
\noindent{\tt Edge Num:} The number of edges associated with each VP.

\smallskip
\noindent{\tt Edge Sum:} The sum of the edge lengths associated with each VP.

\smallskip
\noindent{\tt Proposed:} Our strength measure Eq.~\eqref{eq:strength}.

\smallskip
As shown, by considering the length of an edge and its proximity to the VP, our proposed measure achieves the best performance in selecting the dominant VP in the image.

\begin{figure}[ht!]
\centering
\includegraphics[height =2in]{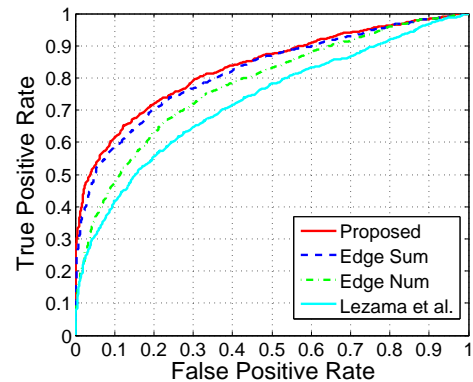}
\caption{Experiment results on dominant vanishing point verification.}
\label{fig:verification}
\end{figure}

\noindent{\bf Dominant vanishing point verification.}
Next, we evaluate the effectiveness of the proposed measure in determining the existence of a dominant VP in the image. For this experiment, we use all the 1,316 images with labeled dominant VPs as positive samples and randomly select 1,500 images without a VP from the ``landscape'' category of AVA dataset as negative samples. In Figure~\ref{fig:verification}, we plot the ROC curves of the three different measures. As a baseline, we also include the result of the Number of False Alarms (NFA) score proposed in~\cite{LezamaGRM14}, which measures the likelihood that a specific configuration (\ie, a VP) arises from a random image. One can clearly see that our proposed measure achieves the best performance.

\begin{figure}[ht]
\centering
\begin{tabular}{cc}
\hspace{-2mm}\includegraphics[height =1.35in]{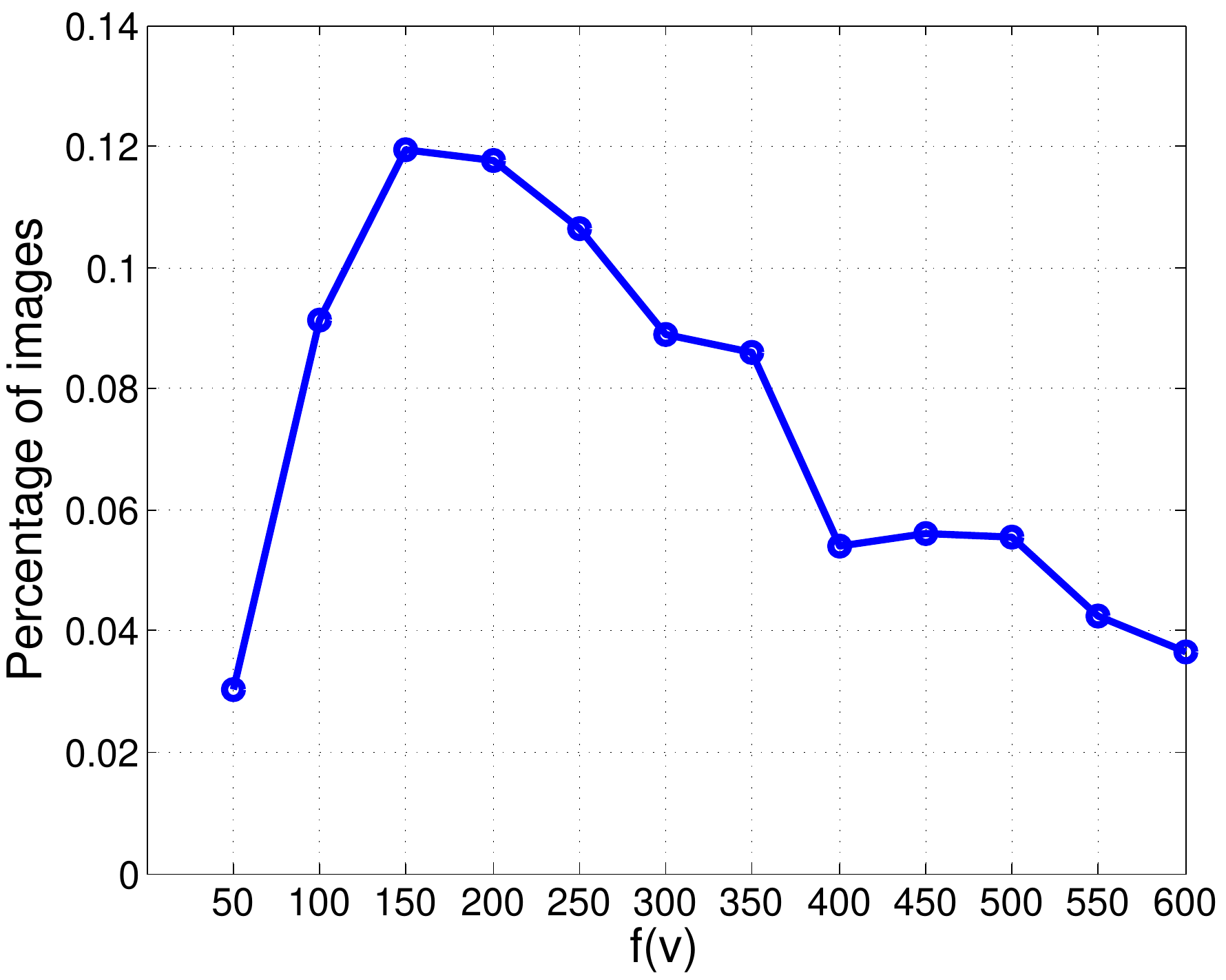} &
\hspace{-2mm}\includegraphics[height =1.35in]{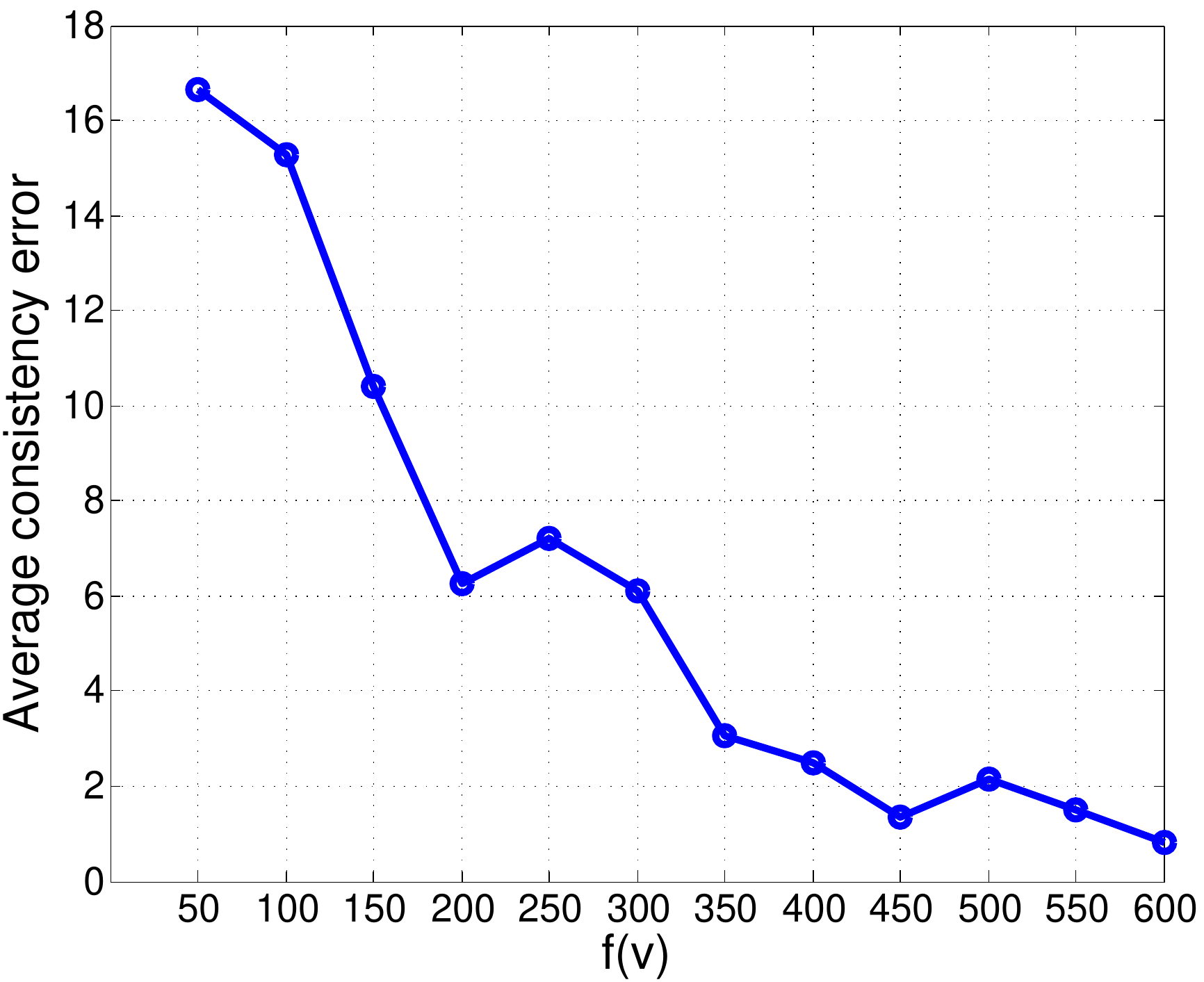} \\
(a) & (b)
\end{tabular}
\caption{The impact of VP strength on the accuracy of dominant VP detection. We show (a) the percentage of images and (b) the average consistency error as a function of our strength measure Eq.~\eqref{eq:strength}.}
\label{fig:acc-f}
\end{figure}

In Figure~\ref{fig:acc-f}(a) and (b), we further plot the percentage of images and the average consistency error as functions of our strength measure, respectively. In particular, Figure~\ref{fig:acc-f}(b) shows that the consistency error decreases substantially when the strength score is higher than 150. This outcome suggests that our strength measure is a good indicator of the reliability of a VP detection result.

%% file: 6application.tex
\section{Photo Composition Application}
\label{sec:retrieval}

In this section, we demonstrate an interesting application of our dominant VP detection method in automatic understanding of photo composition. Nowadays, Cloud-based photo sharing services such as \texttt{flickr.com}, \texttt{photo.net}, and \texttt{dpchallenge.com} have played an increasingly important role in helping the amateurs improve their photography skills. Considering the scenario where a photographer is about to the take a photo of a natural scene, he or she may wonder what photos peers or professional photographers would take in a similar situation. Therefore, given a shot taken by the user, we propose to find exemplar photos about \emph{similar scenes} with \emph{similar points of view} in a large collection of photos. These photos demonstrate the use of various photographic techniques in real world scenarios, hence could potentially be used as feedback to the user on his or her own work.

\subsection{Viewpoint-Specific Image Retrieval}
\label{sec:view-specific}

Given two images $I_i$ and $I_j$, our similarity measure is a sum of two components:
\begin{equation}
D(I_i, I_j) = D_s(I_i, I_j) + D_p(I_i, I_j),
\end{equation}
where $D_s$ and $D_p$ measures the similarity of two images in terms of the scene semantics and the use of linear perspective, respectively. Below we describe each term in detail.

\smallskip
\noindent{\bf Semantic similarity $D_s$:} Recently, it has been shown that generic descriptors extracted from the convolutional neural networks (CNNs) are powerful in capturing the image semantics (\eg, scene types, objects) and have been successfully applied to obtain state-of-the-art image retrieval results~\cite{RazavianASC14}. In our experiment, we adopt the publicly available CNN model trained by~\cite{ChatfieldSVZ14} on the ImageNet ILSVRC challenge dataset\footnote{http://www.vlfeat.org/matconvnet/pretrained/} to compute the semantic similarity. Specifically, we represent each image using the $\ell_2$-normalized output of the second fully connected layer (full7 of~\cite{ChatfieldSVZ14}), and adopt the cosine distance to measure the feature similarity.

\smallskip
\noindent{\bf Perspective similarity $D_p$:} To model the perspective effect in the image, we consider two main factors: (i) the location of the dominant VP and (ii) the position of the associated image elements. For the latter, we focus on the edges consistent with the dominant VP obtained via our contour-based VP detection algorithm. Let $\vv_i$ and $\vv_j$ be the locations of the dominant VPs in images $I_i$ and $I_j$, respectively. We use $\E_i$ (or $\E_j$) to denote the sets of edges consistent with $\vv_i$ (or $\vv_j$). Our perspective similarity measure is defined as:
\begin{equation}
D_p(I_i, I_j) = \gamma_1 \max\left(1- \frac{\|\vv_i - \vv_j\|}{len}, 0\right) + \gamma_2 K(\E_i, \E_j)\;,
\label{eq:perspective-sim}
\end{equation}
where $\|\vv_i - \vv_j\|$ is the Euclidean distance between $\vv_i$ and $\vv_j$, $len$ is the length of the longer side of the image. We resize all the images to $len=500$.

Further, since each edge can be regarded as a set of 2D points on the image, $K(\E_i, \E_j)$ should measure the similarity of two point sets. Here, we use the popular \emph{spatial pyramid matching}~\cite{LazebnikSP06} for its simplicity and efficiency. Generally speaking, this matching scheme is based on a series of increasingly coarser grids on the image. At any fixed resolution, two points are said to match if they fall into the same grid cell. The final matching score is a weighted sum of the number of matches that occur at each level of resolution, where matches found at finer resolutions have higher weights than do matches found at coarser resolutions.

\begin{figure}[t!]
\centering
\includegraphics[height =2in]{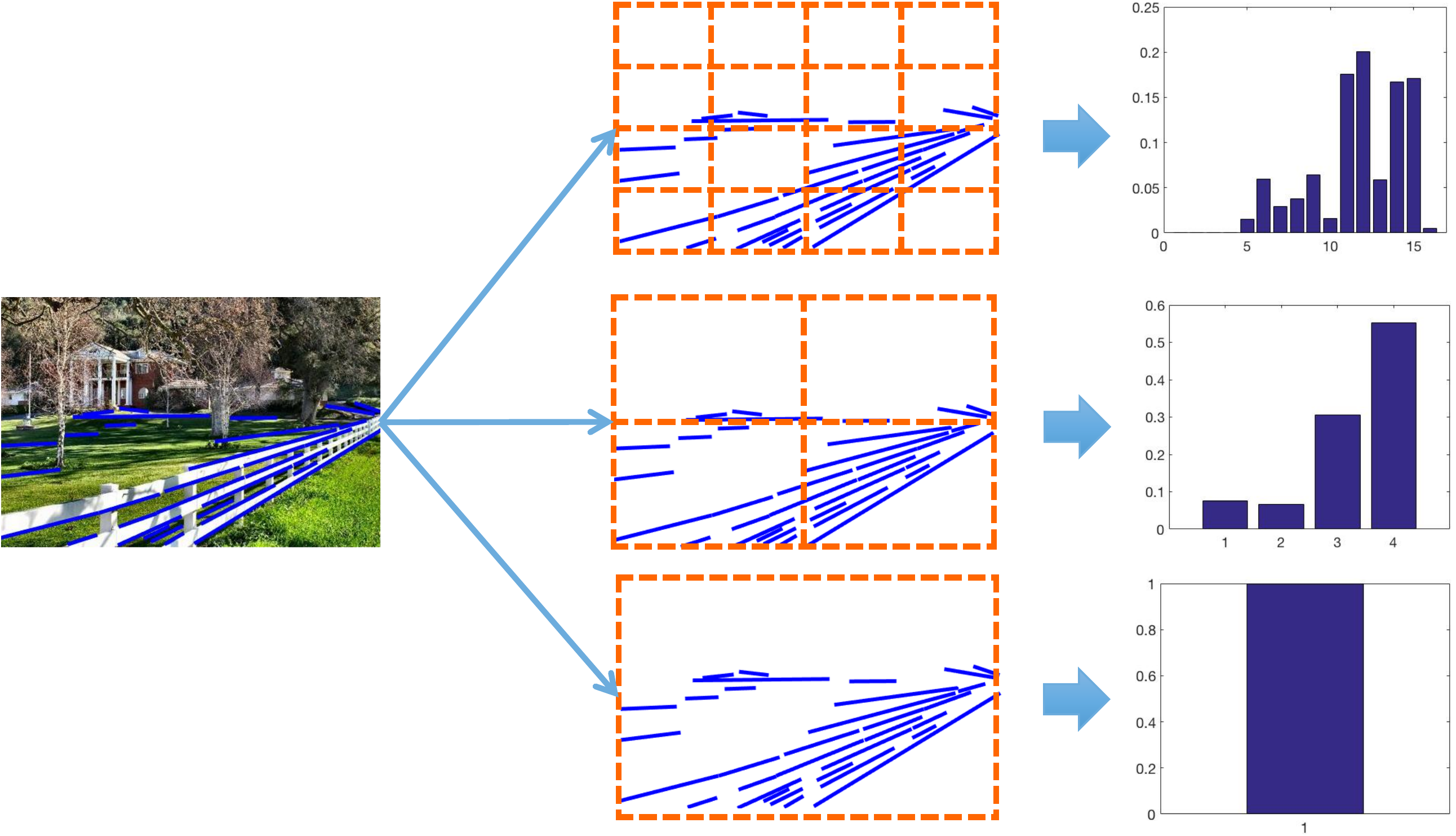}
\caption{Illustration of the construction of a three-level spatial pyramid. Given the set of edges $\E_i$ consistent with the dominant VP, we subdivide the image at three different level of resolutions. For each resolution, we count the number of edge points that fall into each bin to build the histograms $H_i^l, l=0,1,2$.}
\label{fig:pyramid}
\end{figure}

For our problem, we first construct a series of grids at resolutions $0, 1, \ldots, L$, as illustrated in Figure~\ref{fig:pyramid}. Note that the grid at level $l$ has $2^l$ cells along each dimension, so the total number of cells is $2^{2l}$. At level $l$, let $H_i^l(k)$ and $H_j^l(k)$ denote the number of points from $\E_i$ and $\E_j$ that fall into the $k$-th cell, respectively. The number of matches at level $l$ is then given by the histogram intersection function:
\begin{equation}
\mathcal{I}(H_i^l, H_j^l) = \sum_{k} \min \left(H_i^l(k), H_j^l(k)\right)\;.
\end{equation}
Below we write $\mathcal{I}(H_i^l, H_j^l)$ as $\mathcal{I}^l$ for short.

Since the number of matches found at level $l$ also includes all the matches found at the level $l+1$, the number of new matches at level $l$ is given by $\mathcal{I}^l - \mathcal{I}^{l+1}$, $\forall l = 0, \ldots, l-1$. To reward matches found at finer levels, we assign weight $2^{-(L-l)}$ to the matches at level $l$. Note that the weight is inversely proportional to the cell width at that level. Finally, the pyramid matching score is defined as:
\begin{eqnarray}
K^L(\E_i, \E_j) & = & \mathcal{I}^L + \sum_{l=0}^{L-1} \frac{1}{2^{L-l}} (\mathcal{I}^l - \mathcal{I}^{l+1}) \\
& = & \frac{1}{2^L} \mathcal{I}^0 + \sum_{l=1}^L \frac{1}{2^{L-l+1}}\mathcal{I}^l\;.
\end{eqnarray}
Here, we use superscript ``$L$'' to indicate its dependency on the parameter $L$. We empirically set the parameters for viewpoint-specific image retrieval for all experiments to: $\gamma_1 = \gamma_2 = 0.5, L =6$.

\begin{figure*}[p!]
\centering
\begin{tabular}{c|l}
\hspace{-1mm}\cfbox{red}{\includegraphics[height =0.5in]{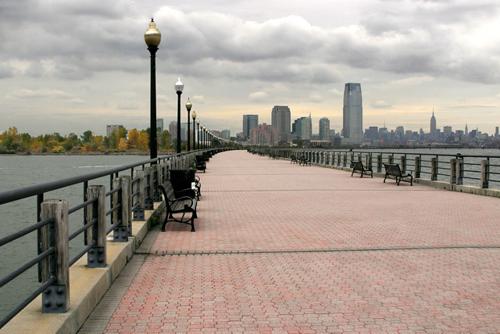}}&
\hspace{-1mm}\cfbox{red}{\includegraphics[height =0.5in]{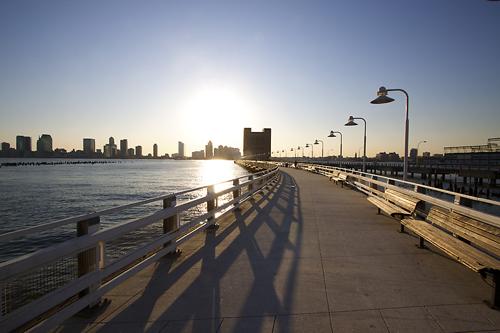}}\hspace{0.1mm}
\includegraphics[height =0.5in]{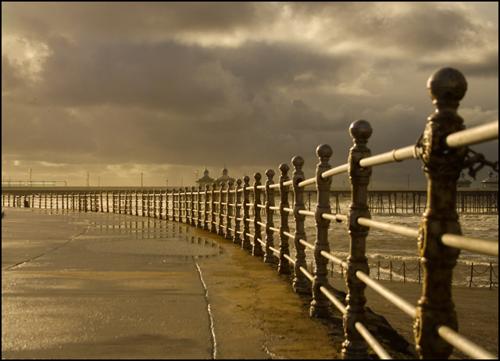}\hspace{0.1mm}
\includegraphics[height =0.5in]{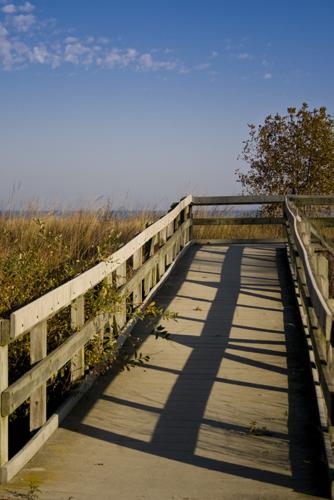}\hspace{0.1mm}
\includegraphics[height =0.5in]{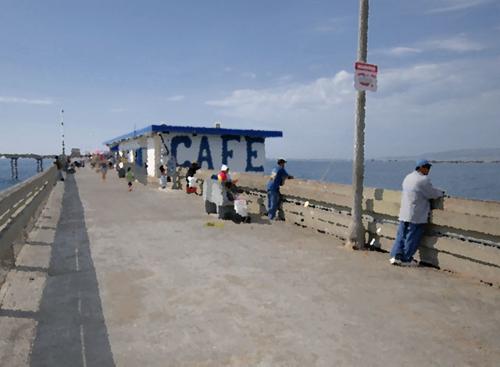}\hspace{0.1mm}
\includegraphics[height =0.5in]{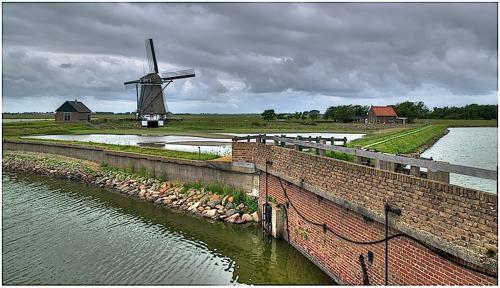}\hspace{0.1mm}
\includegraphics[height =0.5in]{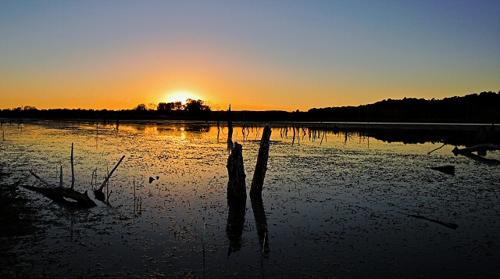}\hspace{0.1mm}
\includegraphics[height =0.5in]{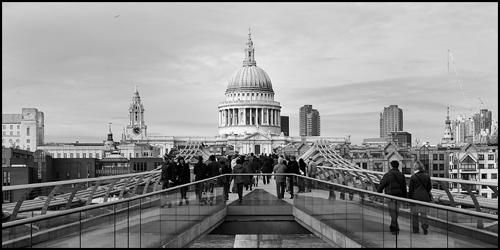}\hspace{0.1mm}
\\
\hspace{-1mm}\includegraphics[height =0.52in]{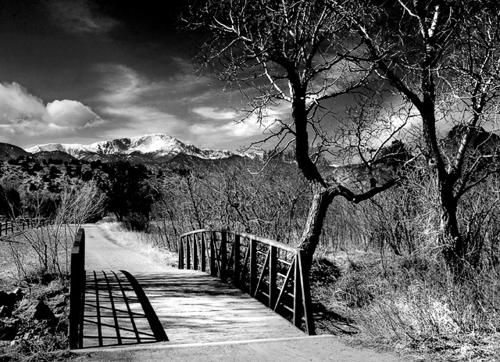}&
\hspace{-1mm}\includegraphics[height =0.52in]{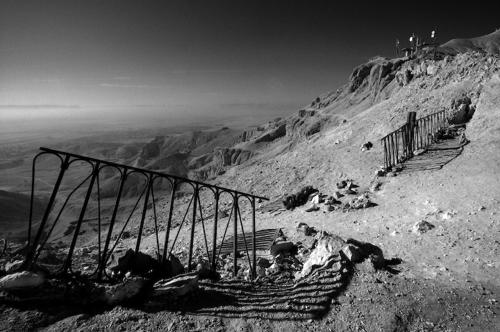}\hspace{0.1mm}
\includegraphics[height =0.52in]{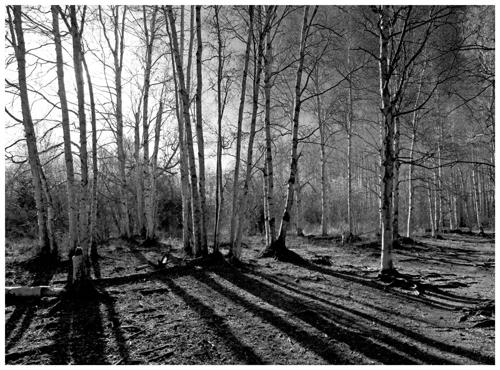}\hspace{0.1mm}
\includegraphics[height =0.52in]{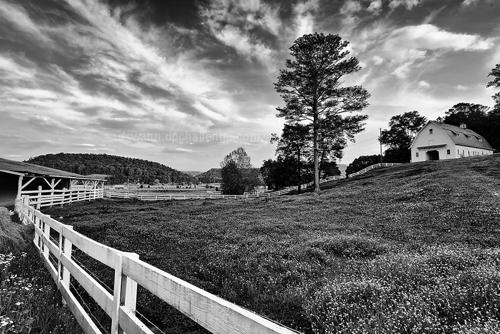}\hspace{0.1mm}
\includegraphics[height =0.52in]{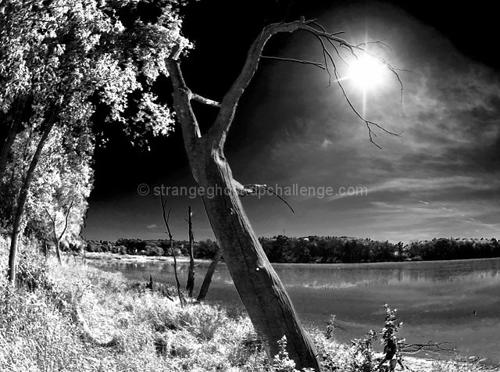}\hspace{0.1mm}
\includegraphics[height =0.52in]{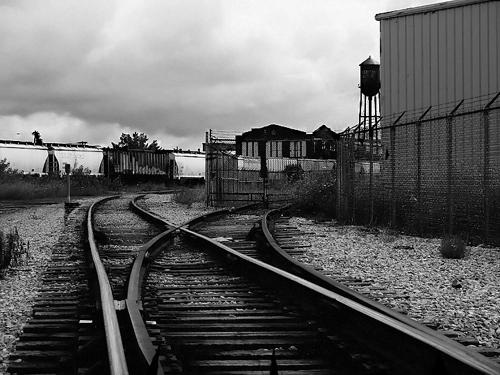}\hspace{0.1mm}
\includegraphics[height =0.52in]{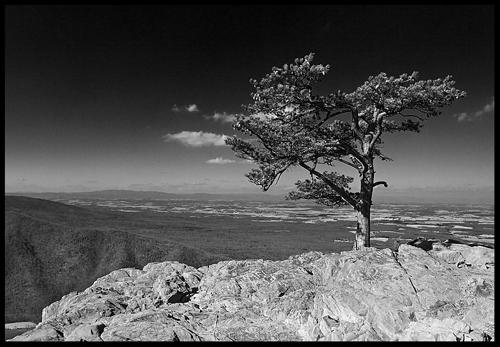}\hspace{0.1mm}
\includegraphics[height =0.52in]{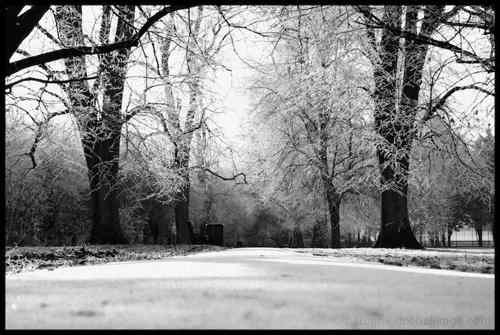}\hspace{0.1mm}
\\
\hspace{-1mm}\includegraphics[height =0.53in]{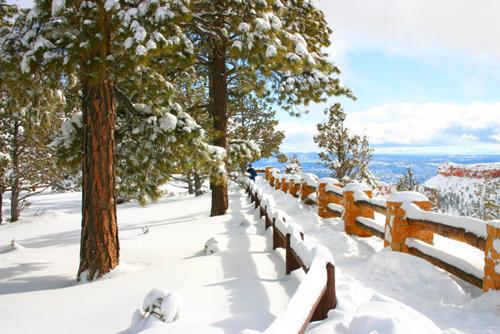}&
\hspace{-1mm}\includegraphics[height =0.53in]{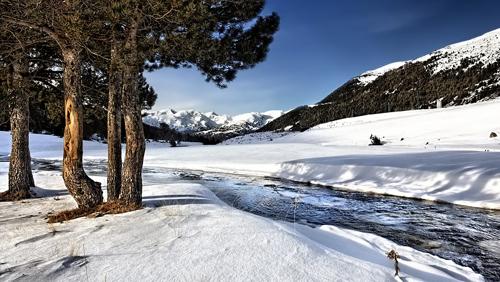}\hspace{0.1mm}
\includegraphics[height =0.53in]{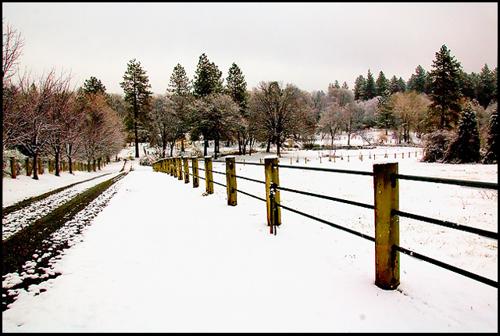}\hspace{0.1mm}
\includegraphics[height =0.53in]{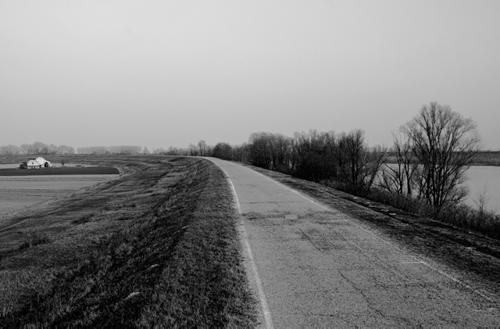}\hspace{0.1mm}
\includegraphics[height =0.53in]{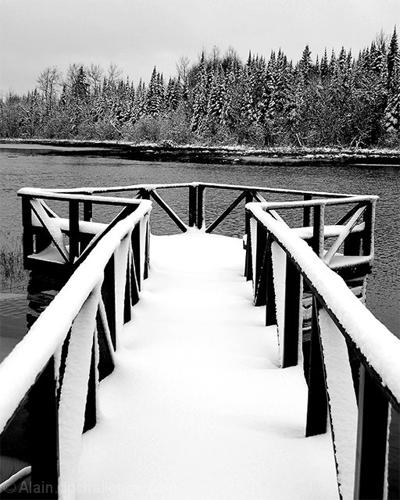}\hspace{0.1mm}
\includegraphics[height =0.53in]{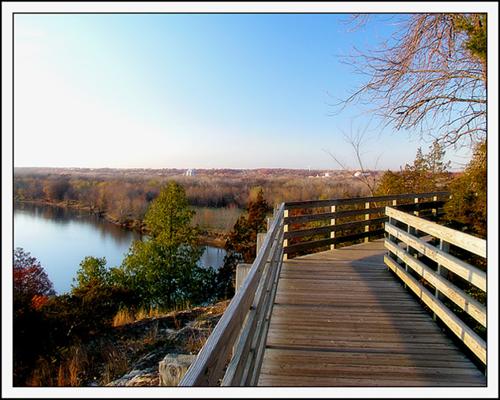}\hspace{0.1mm}
\includegraphics[height =0.53in]{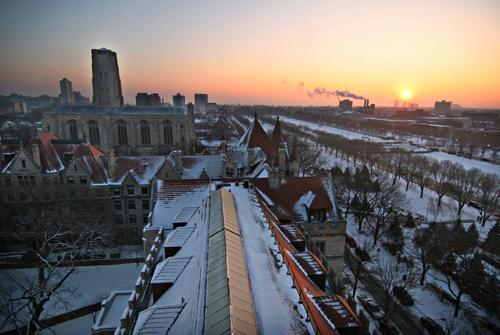}\hspace{0.1mm}
\includegraphics[height =0.53in]{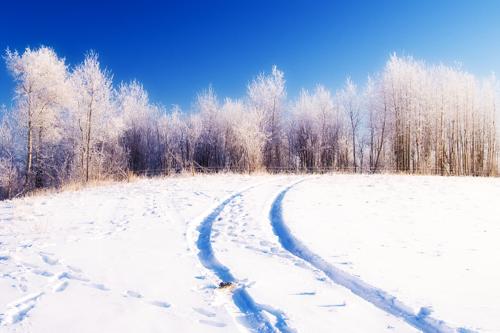}\hspace{0.1mm}
\\
\hspace{-1mm}\includegraphics[height =0.52in]{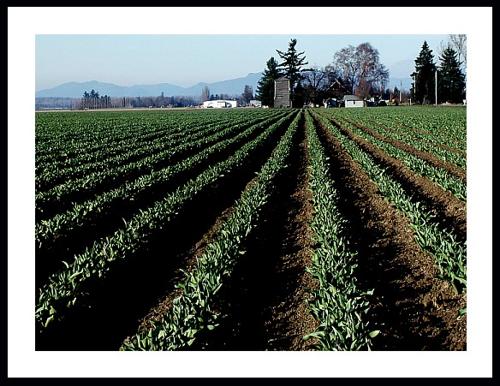}&
\hspace{-1mm}\includegraphics[height =0.52in]{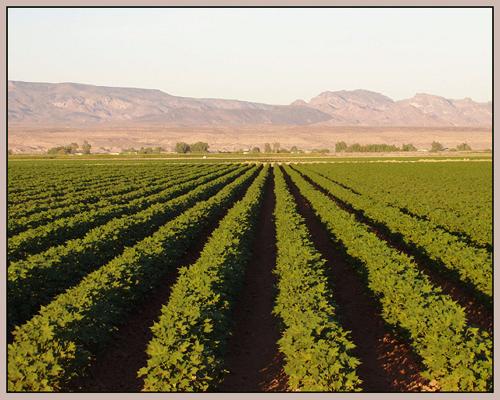}\hspace{0.1mm}
\includegraphics[height =0.52in]{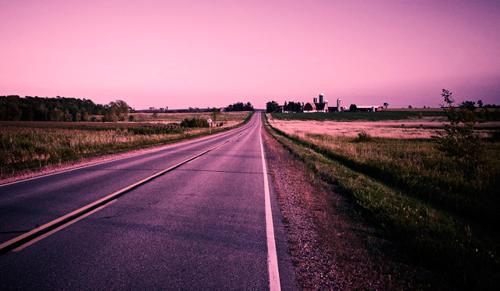}\hspace{0.1mm}
\includegraphics[height =0.52in]{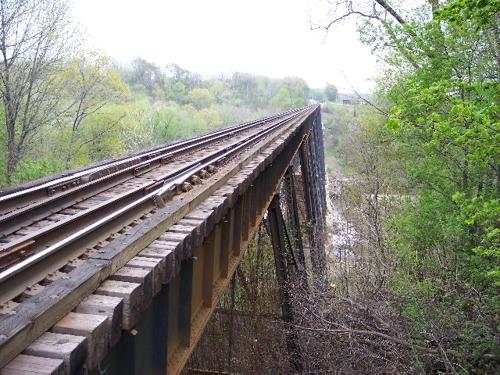}\hspace{0.1mm}
\includegraphics[height =0.52in]{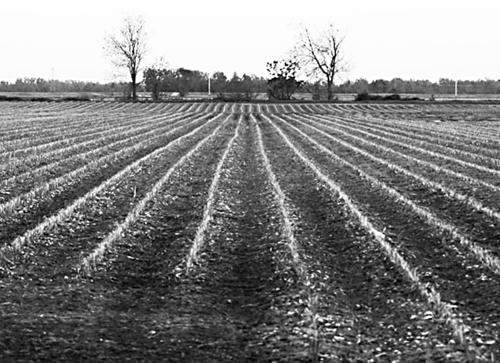}\hspace{0.1mm}
\includegraphics[height =0.52in]{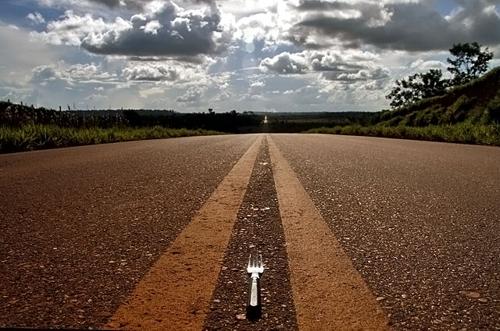}\hspace{0.1mm}
\includegraphics[height =0.52in]{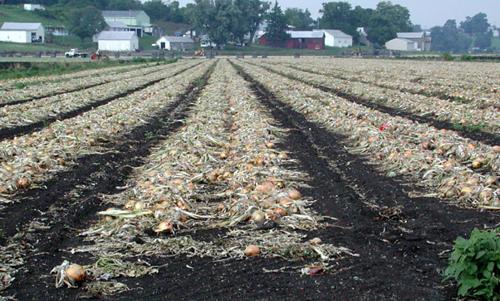}\hspace{0.1mm}
\includegraphics[height =0.52in]{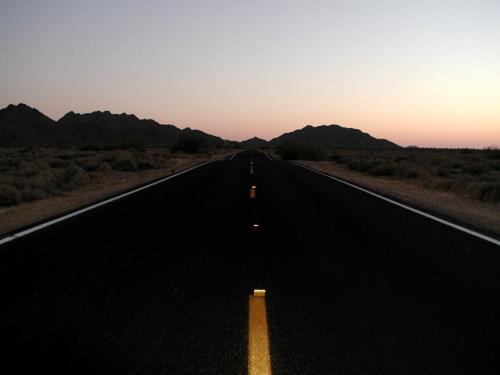}\hspace{0.1mm}
\\
\hspace{-1mm}\includegraphics[height =0.55in]{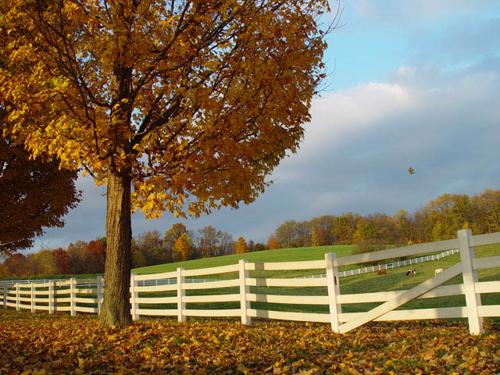}&
\hspace{-1mm}\includegraphics[height =0.55in]{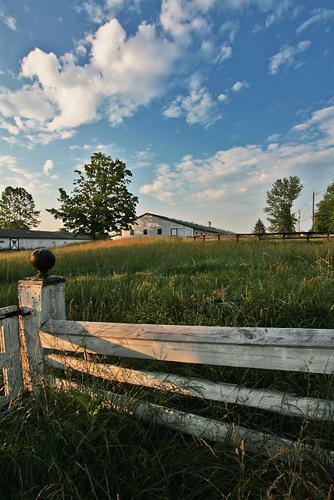}\hspace{0.1mm}
\includegraphics[height =0.55in]{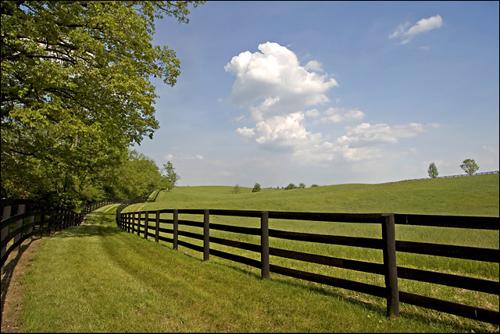}\hspace{0.1mm}
\includegraphics[height =0.55in]{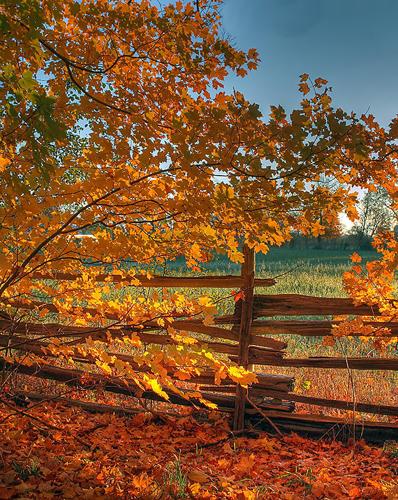}\hspace{0.1mm}
\includegraphics[height =0.55in]{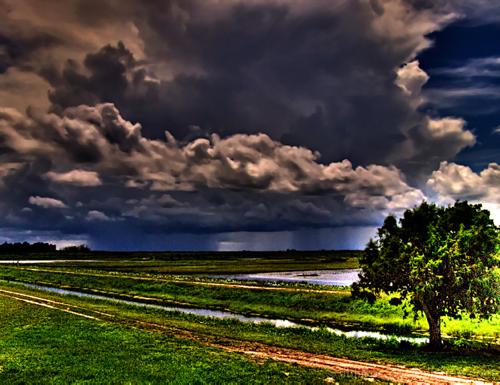}\hspace{0.1mm}
\includegraphics[height =0.55in]{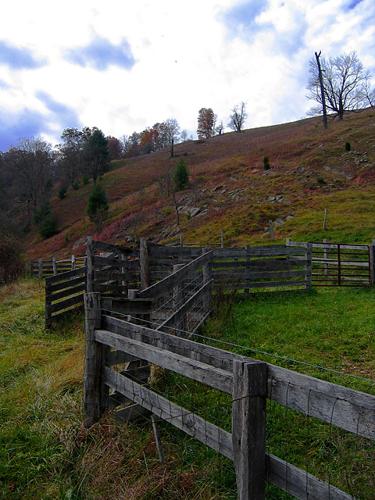}\hspace{0.1mm}
\includegraphics[height =0.55in]{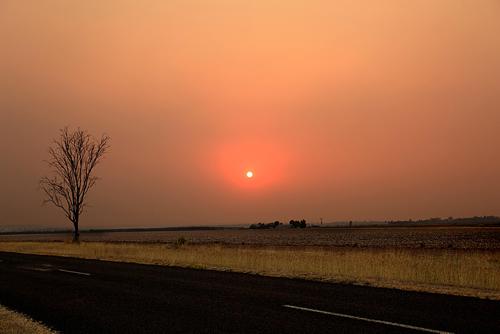}\hspace{0.1mm}
\includegraphics[height =0.55in]{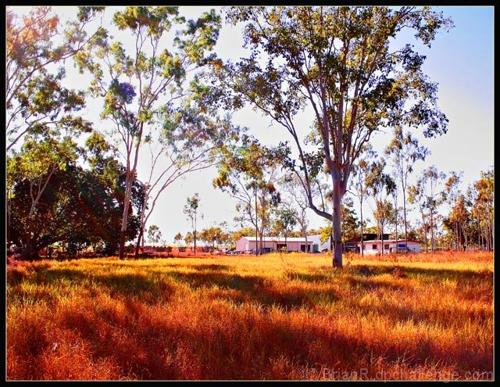}\hspace{0.1mm}
\includegraphics[height =0.55in]{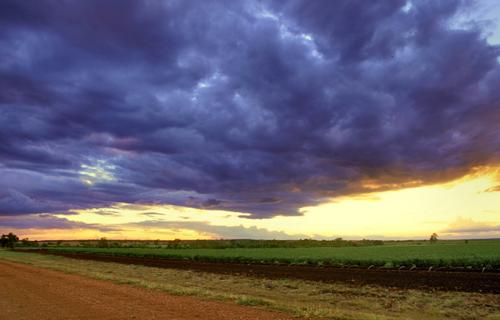}\\
\hspace{-1mm}\cfbox{blue}{\includegraphics[height =0.58in]{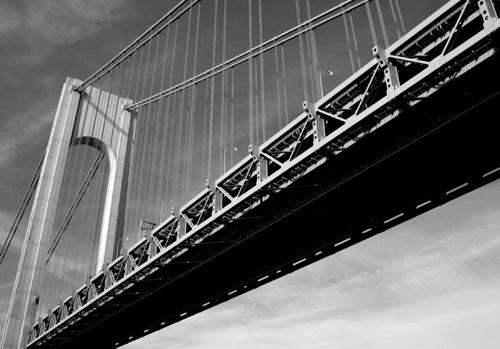}}&
\hspace{-1mm}\includegraphics[height =0.58in]{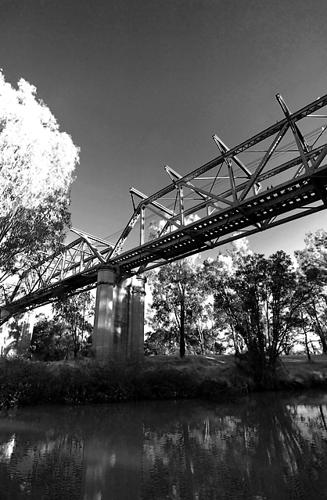}\hspace{0.1mm}
\includegraphics[height =0.58in]{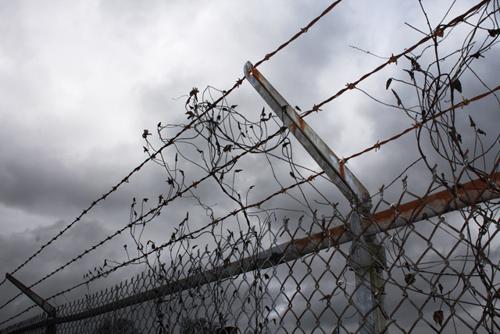}\hspace{0.1mm}
\cfbox{blue}{\includegraphics[height =0.58in]{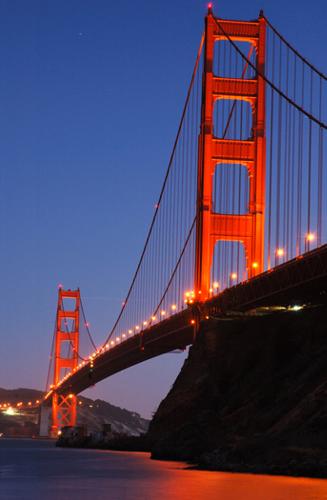}}\hspace{0.1mm}
\includegraphics[height =0.58in]{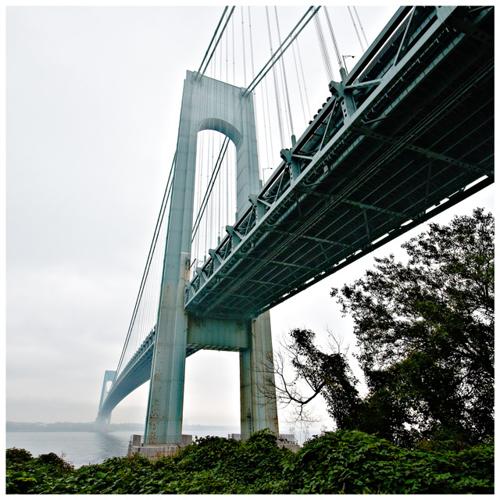}\hspace{0.1mm}
\includegraphics[height =0.58in]{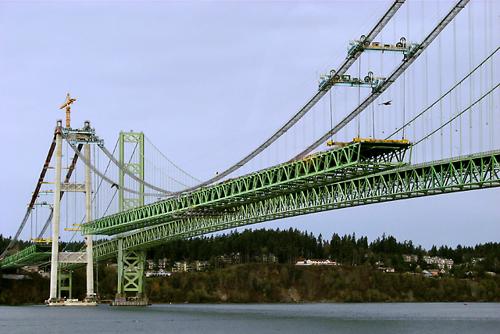}\hspace{0.1mm}
\cfbox{blue}{\includegraphics[height =0.58in]{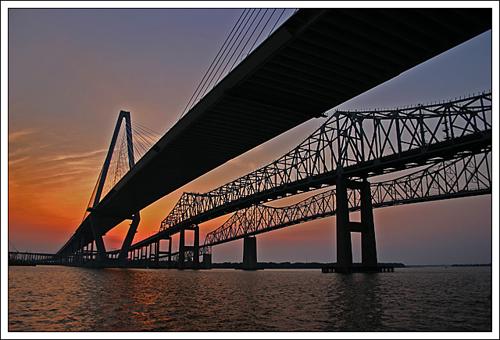}}\hspace{0.1mm}
\includegraphics[height =0.58in]{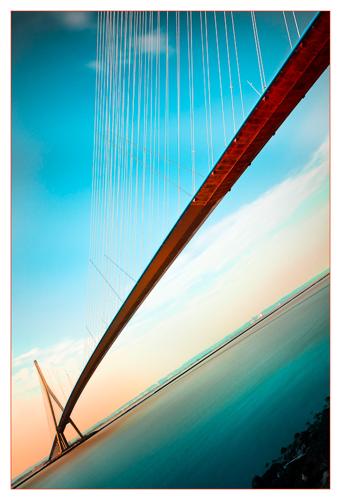}\hspace{0.1mm}
\includegraphics[height =0.58in]{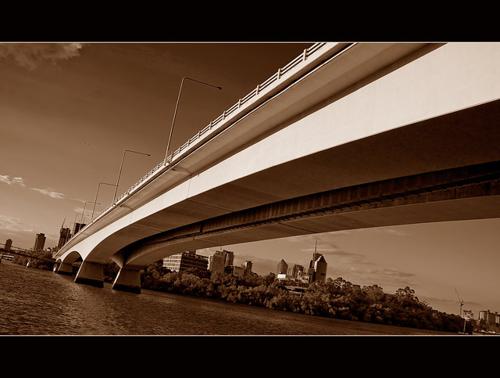}\\
\hspace{-1mm}\includegraphics[height =0.56in]{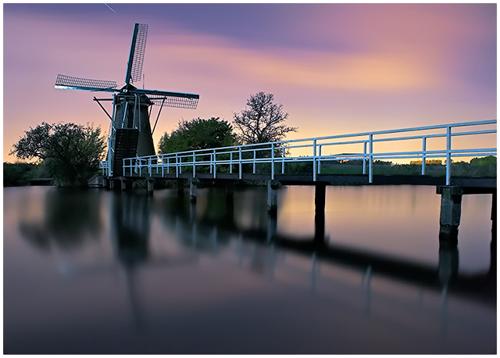}&
\hspace{-1mm}\includegraphics[height =0.56in]{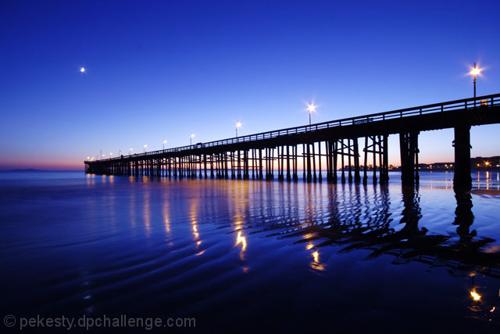}\hspace{0.1mm}
\includegraphics[height =0.56in]{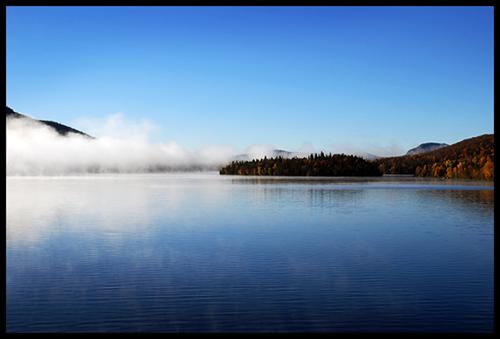}\hspace{0.1mm}
\includegraphics[height =0.56in]{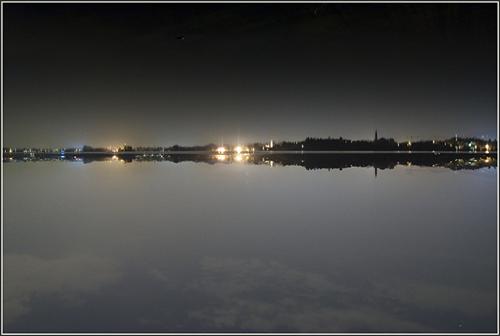}\hspace{0.1mm}
\includegraphics[height =0.56in]{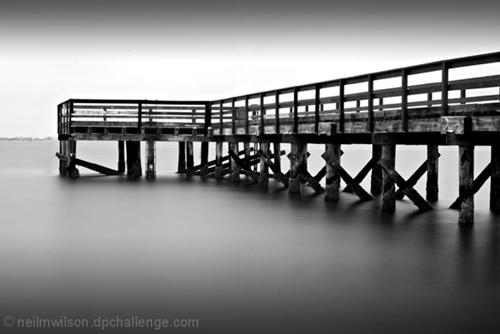}\hspace{0.1mm}
\includegraphics[height =0.56in]{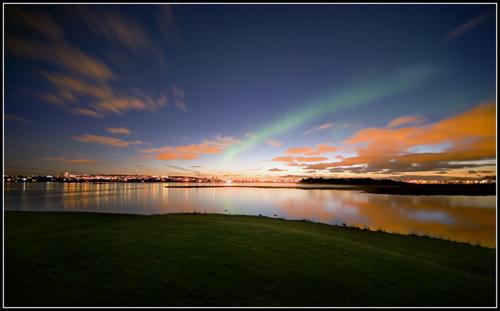}\hspace{0.1mm}
\includegraphics[height =0.56in]{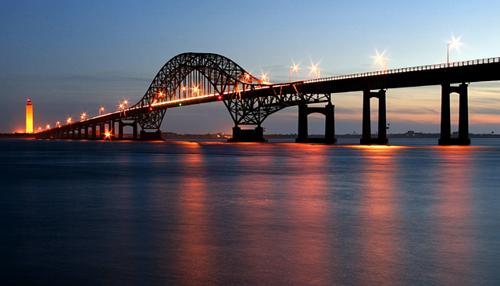}\hspace{0.1mm}
\\
\hspace{-1mm}\includegraphics[height =0.52in]{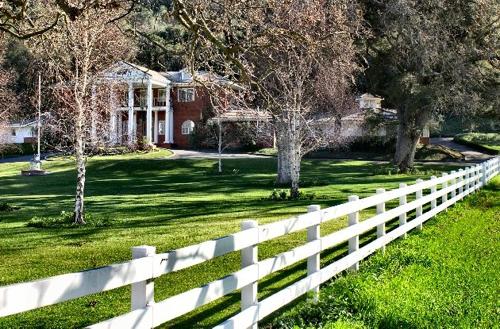}&
\hspace{-1mm}\includegraphics[height =0.52in]{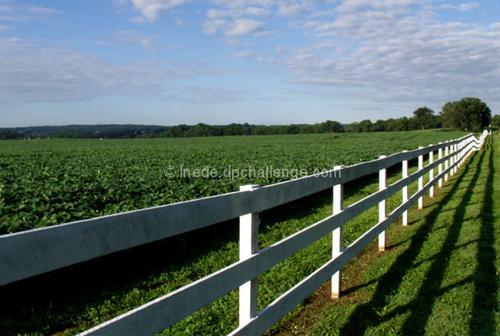}\hspace{0.1mm}
\includegraphics[height =0.52in]{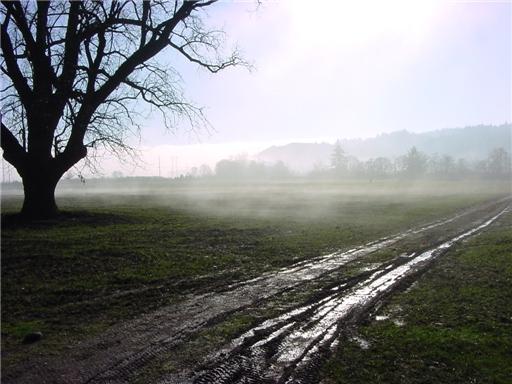}\hspace{0.1mm}
\includegraphics[height =0.52in]{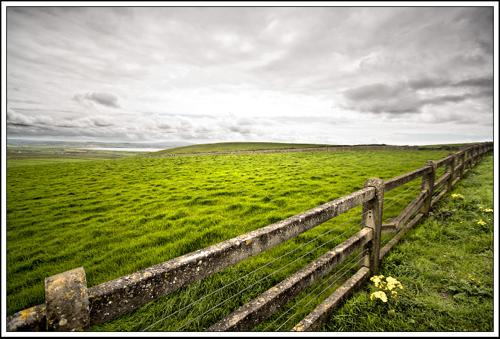}\hspace{0.1mm}
\includegraphics[height =0.52in]{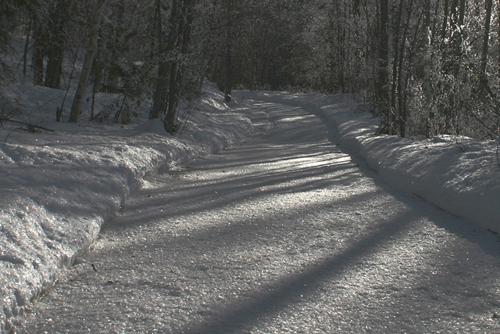}\hspace{0.1mm}
\includegraphics[height =0.52in]{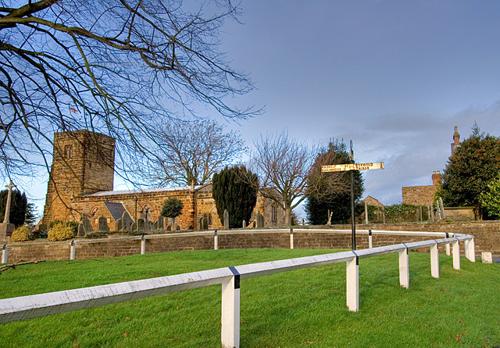}\hspace{0.1mm}
\includegraphics[height =0.52in]{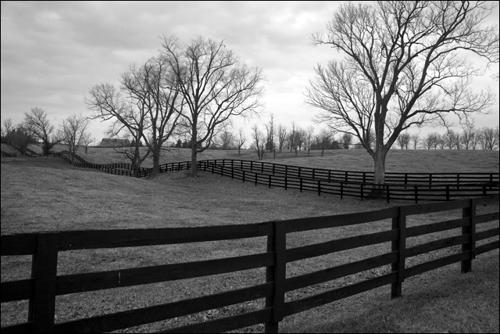}\hspace{0.1mm}
\includegraphics[height =0.52in]{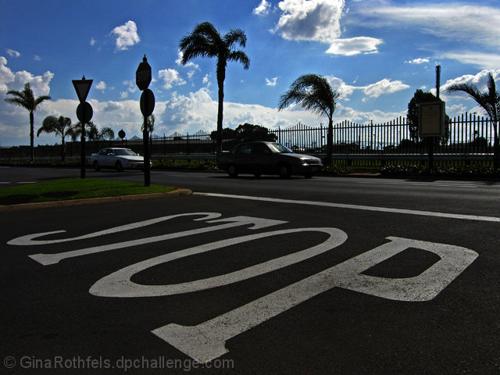}\hspace{0.1mm}
\\
\hspace{-1mm}\cfbox{green}{\includegraphics[height =0.52in]{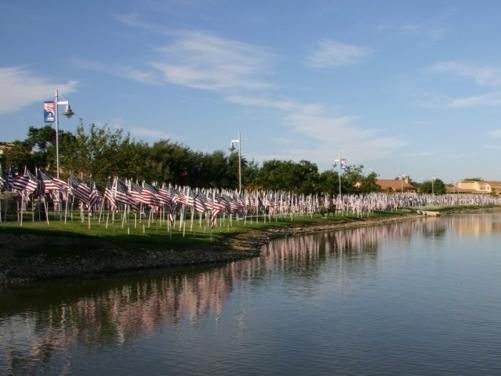}}&
\hspace{-1mm}\includegraphics[height =0.52in]{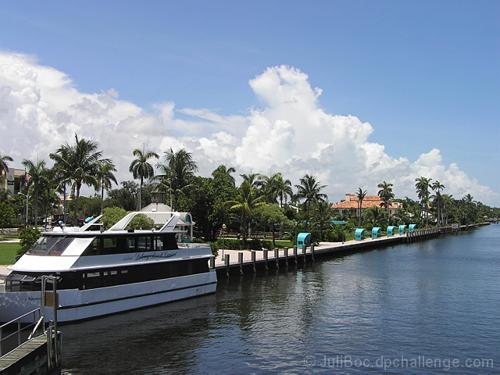}\hspace{0.1mm}
\cfbox{green}{\includegraphics[height =0.52in]{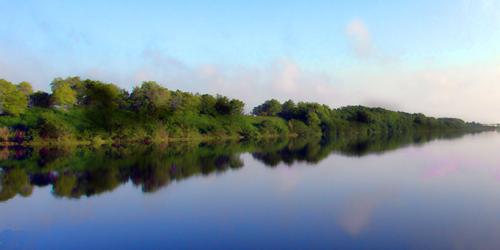}}\hspace{0.1mm}
\includegraphics[height =0.52in]{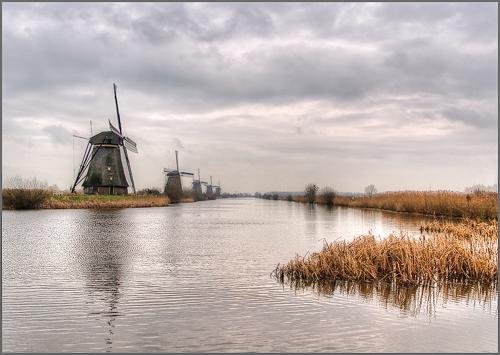}\hspace{0.1mm}
\includegraphics[height =0.52in]{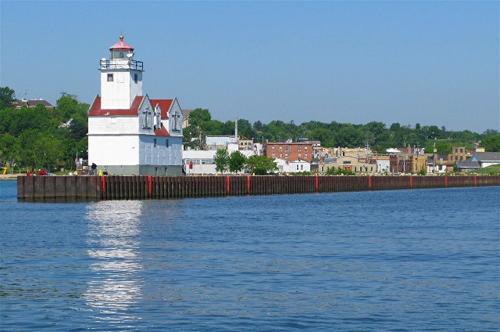}\hspace{0.1mm}
\includegraphics[height =0.52in]{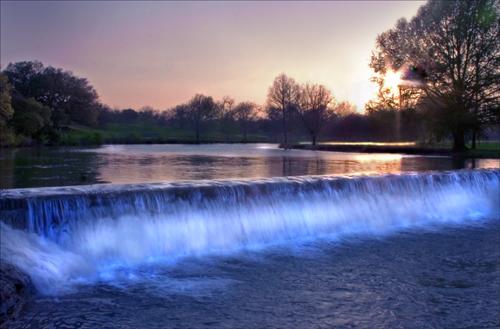}\hspace{0.1mm}
\includegraphics[height =0.52in]{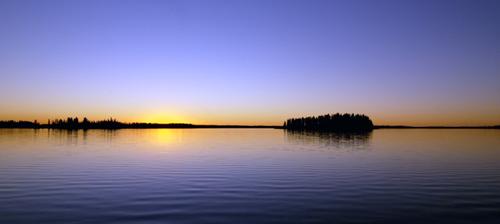}\hspace{0.1mm}
\\
\hspace{-1mm}\includegraphics[height =0.65in]{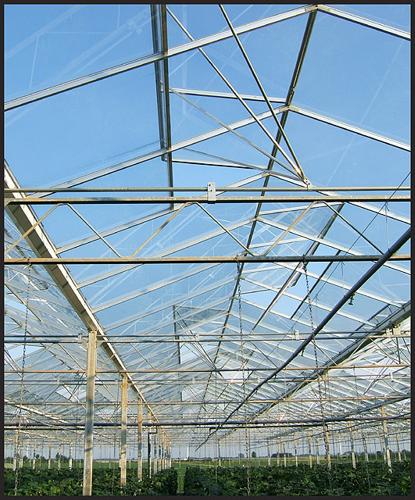}&
\hspace{-1mm}\includegraphics[height =0.65in]{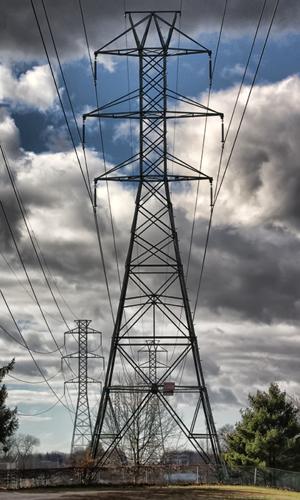}\hspace{0.1mm}
\includegraphics[height =0.65in]{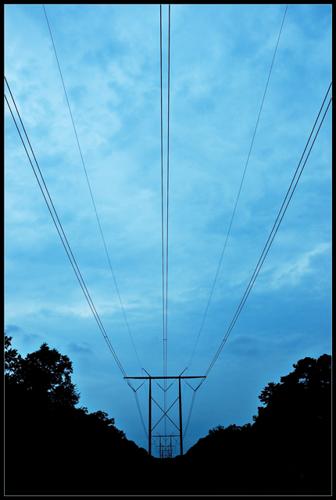}\hspace{0.1mm}
\includegraphics[height =0.65in]{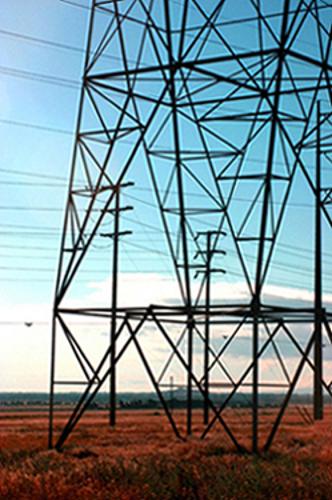}\hspace{0.1mm}
\includegraphics[height =0.65in]{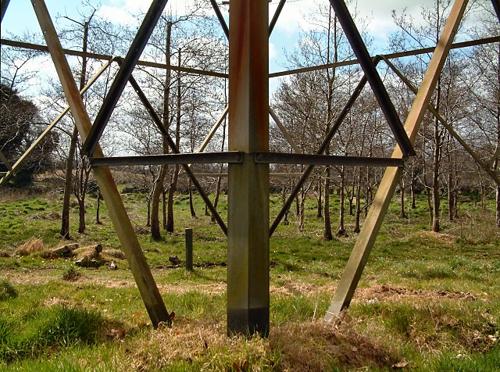}\hspace{0.1mm}
\includegraphics[height =0.65in]{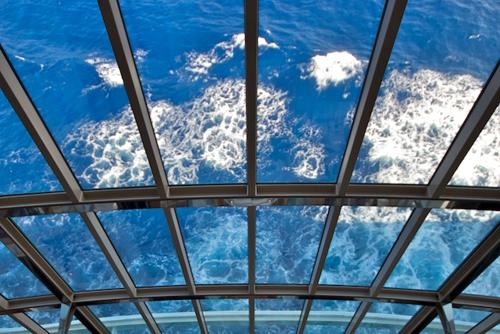}\hspace{0.1mm}
\includegraphics[height =0.65in]{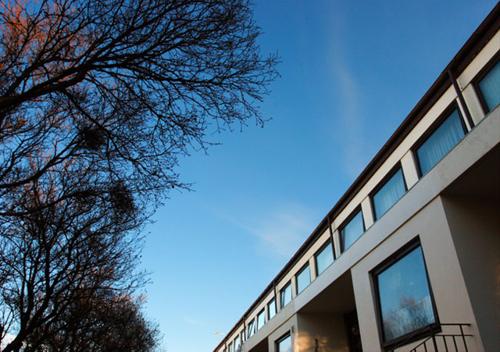}\hspace{0.1mm}
\includegraphics[height =0.65in]{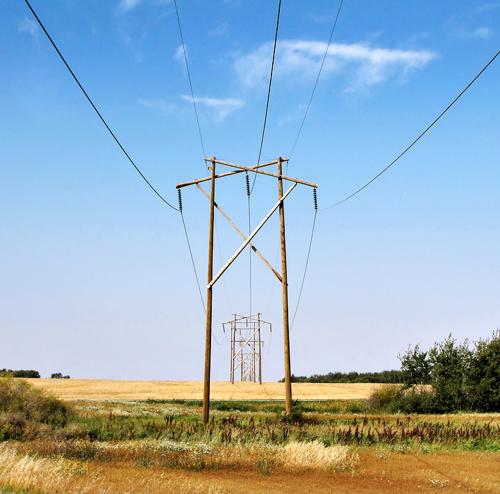}\hspace{0.1mm}
\includegraphics[height =0.65in]{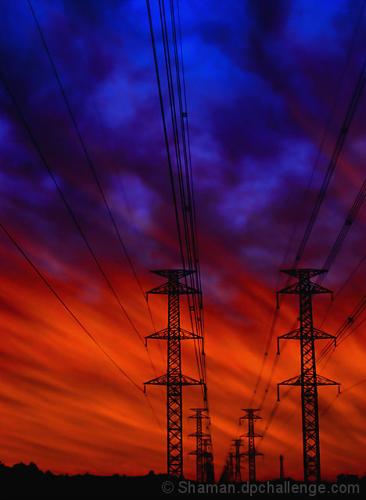}\\
\hspace{-1mm}\cfbox{yellow}{\includegraphics[height =0.62in]{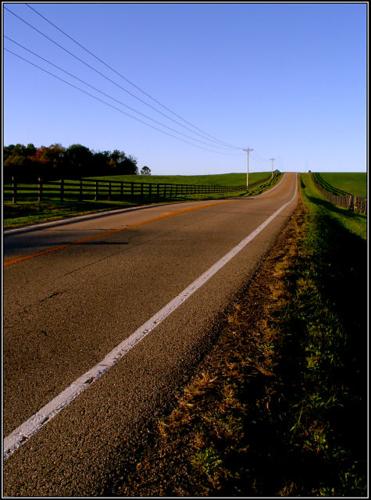}}&
\hspace{-1mm}\cfbox{yellow}{\includegraphics[height =0.62in]{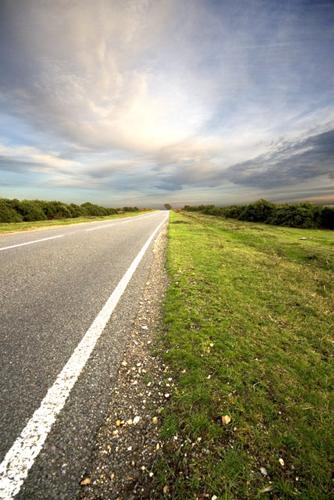}}\hspace{0.1mm}
\includegraphics[height =0.62in]{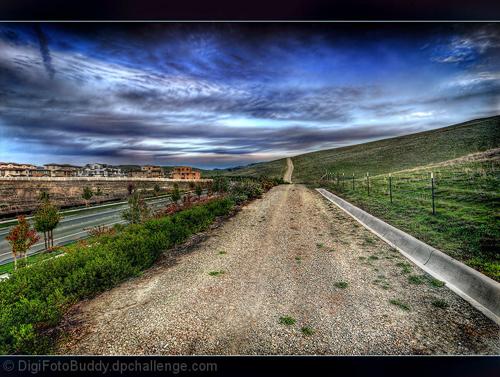}\hspace{0.1mm}
\includegraphics[height =0.62in]{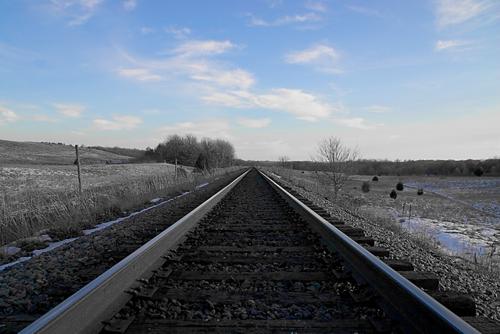}\hspace{0.1mm}
\includegraphics[height =0.62in]{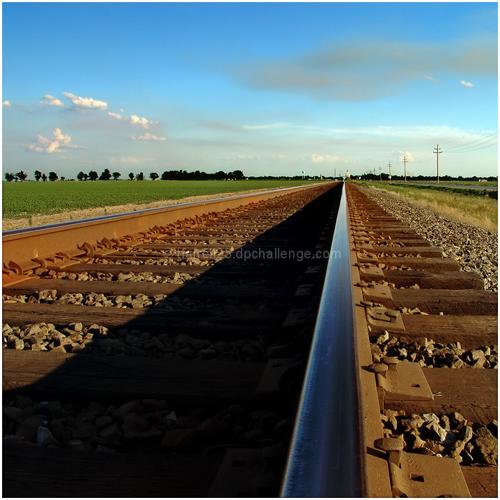}\hspace{0.1mm}
\includegraphics[height =0.62in]{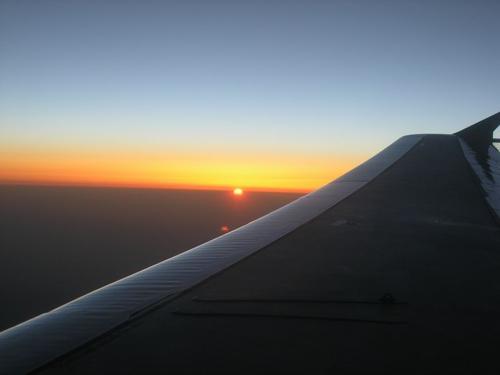}\hspace{0.1mm}
\includegraphics[height =0.62in]{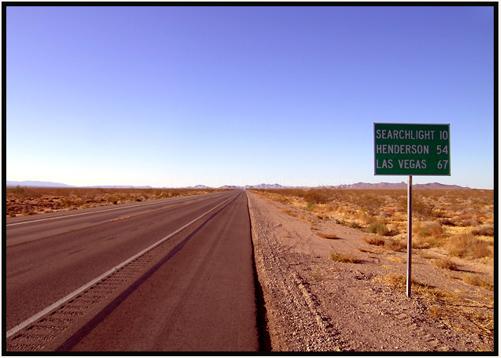}\hspace{0.1mm}
\includegraphics[height =0.62in]{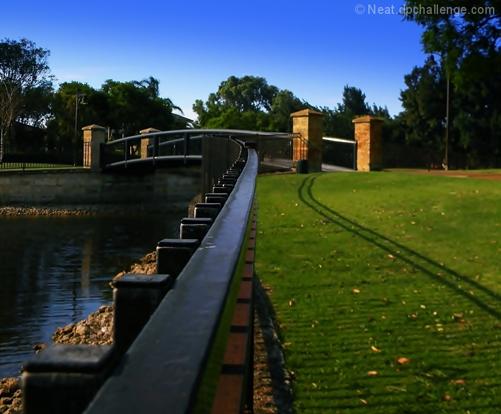}\hspace{0.1mm}
\\
\hspace{-1mm}\includegraphics[height =0.55in]{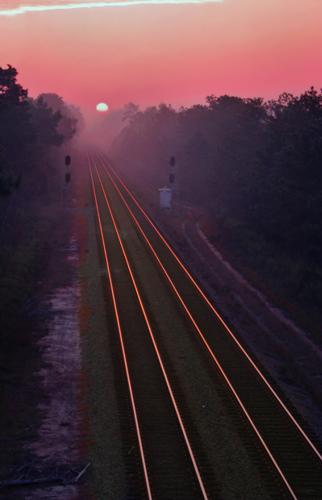}&
\hspace{-1mm}\includegraphics[height =0.55in]{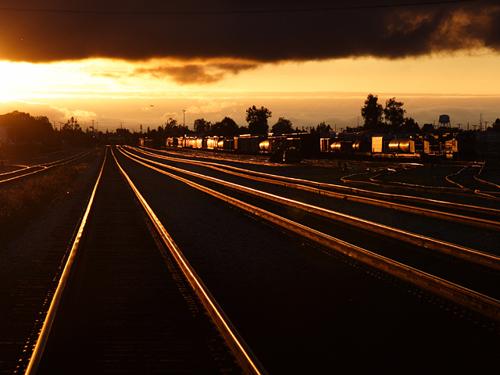}\hspace{0.1mm}
\includegraphics[height =0.55in]{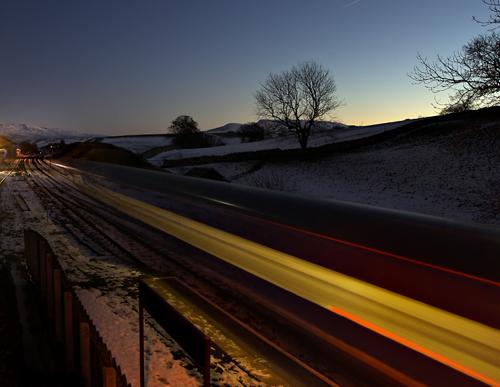}\hspace{0.1mm}
\includegraphics[height =0.55in]{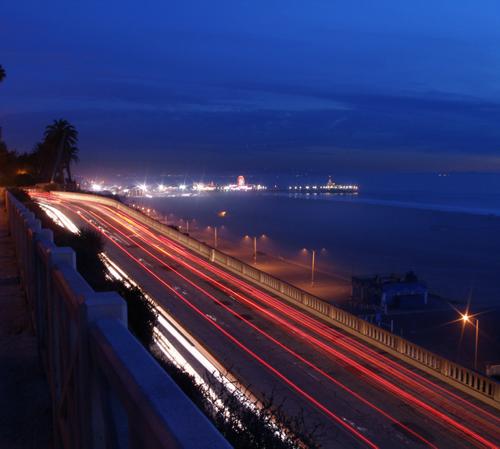}\hspace{0.1mm}
\includegraphics[height =0.55in]{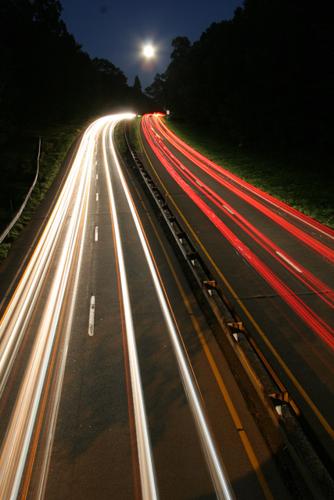}\hspace{0.1mm}
\includegraphics[height =0.55in]{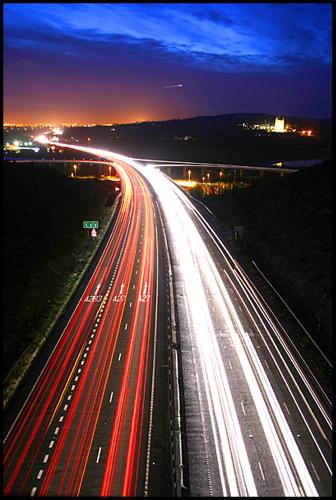}\hspace{0.1mm}
\includegraphics[height =0.55in]{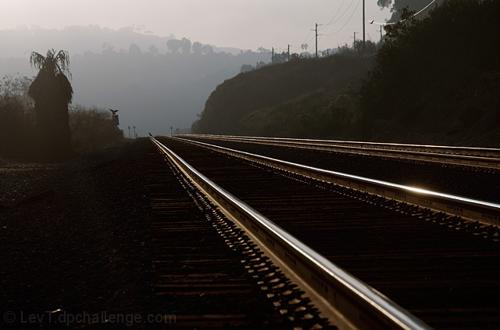}\hspace{0.1mm}
\includegraphics[height =0.55in]{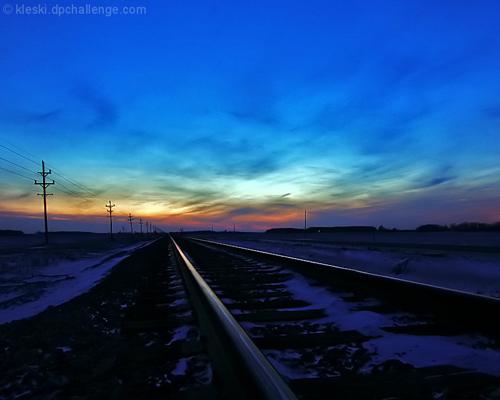}\hspace{0.1mm}
\includegraphics[height =0.55in]{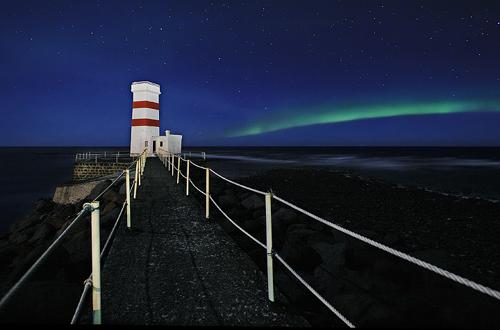}
\\
\hline\vspace{-2mm}\\
\hspace{-1mm}\includegraphics[height =0.59in]{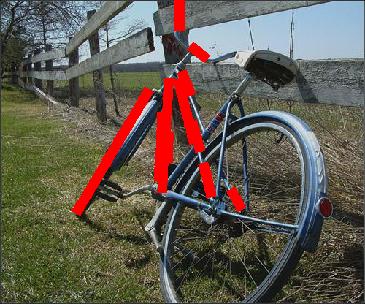}&
\hspace{-1mm}\includegraphics[height =0.59in]{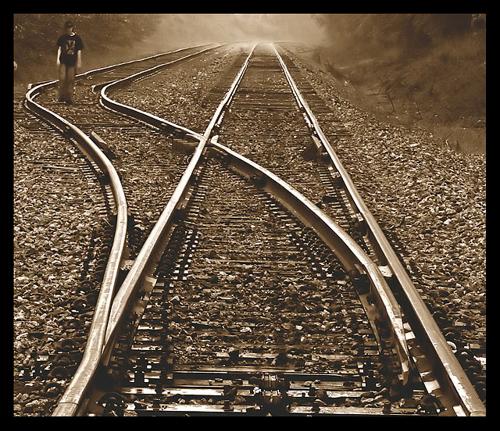}\hspace{0.1mm}
\includegraphics[height =0.59in]{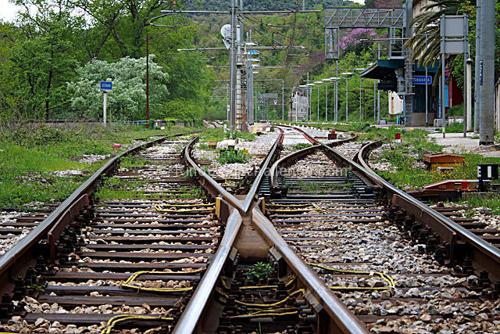}\hspace{0.1mm}
\includegraphics[height =0.59in]{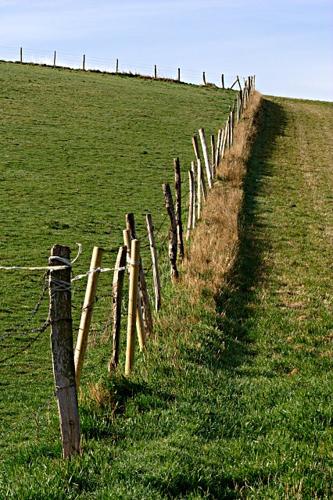}\hspace{0.1mm}
\includegraphics[height =0.59in]{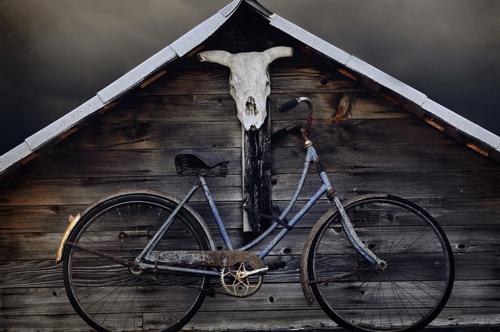}\hspace{0.1mm}
\includegraphics[height =0.59in]{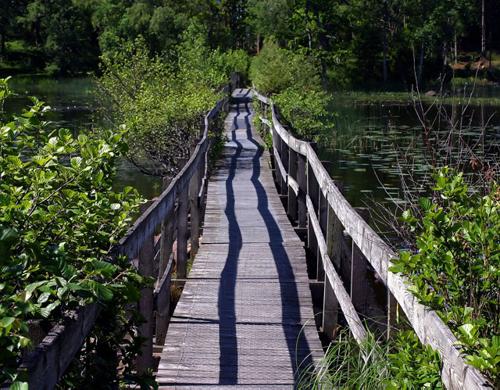}\hspace{0.1mm}
\includegraphics[height =0.59in]{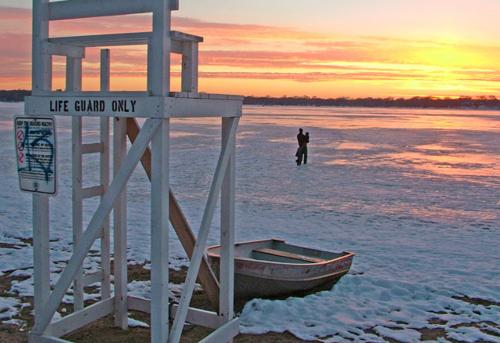}\hspace{0.1mm}
\includegraphics[height =0.59in]{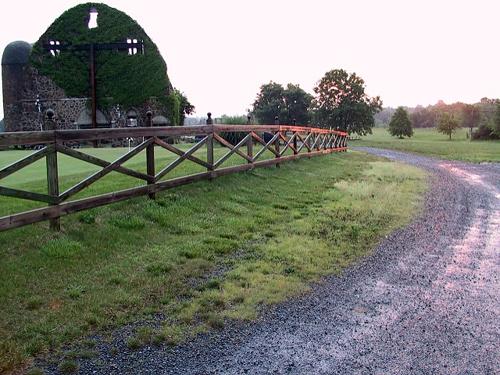}\hspace{0.1mm}
\\
\hspace{-1mm}\includegraphics[height =0.59in]{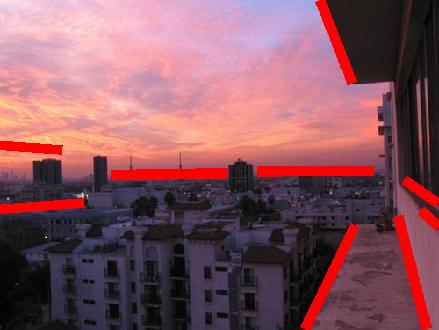}&
\hspace{-1mm}\includegraphics[height =0.59in]{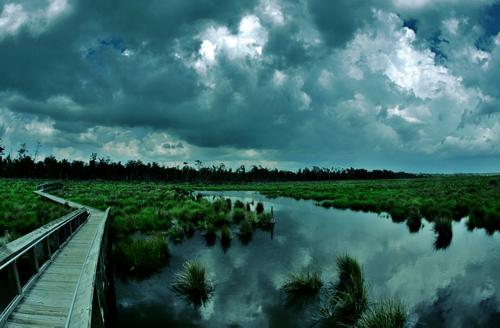}\hspace{0.1mm}
\includegraphics[height =0.59in]{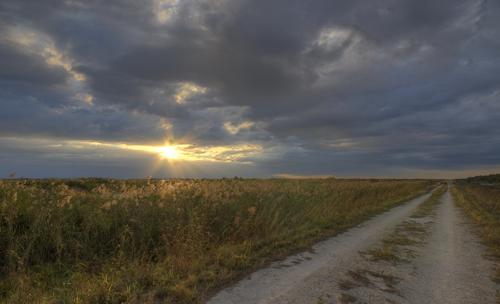}\hspace{0.1mm}
\includegraphics[height =0.59in]{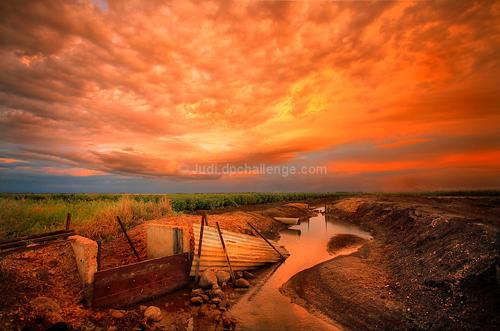}\hspace{0.1mm}
\includegraphics[height =0.59in]{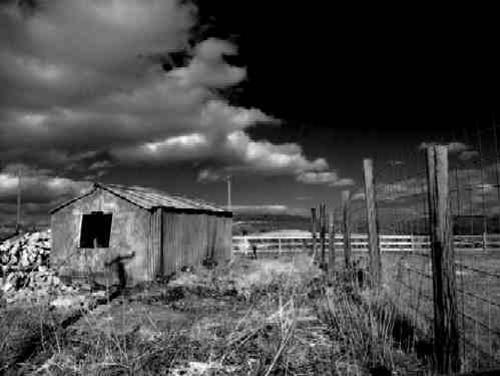}\hspace{0.1mm}
\includegraphics[height =0.59in]{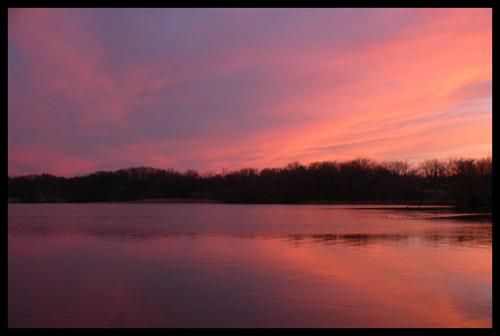}\hspace{0.1mm}
\includegraphics[height =0.59in]{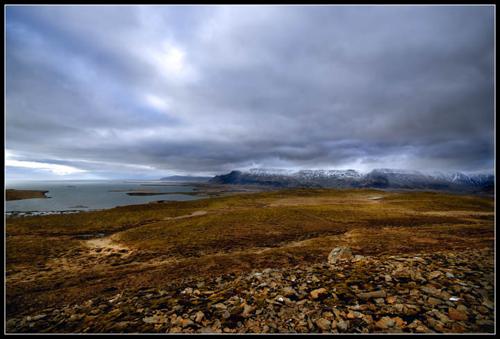}\hspace{0.1mm}
\\
\end{tabular}
\caption{Example viewpoint-specific image retrieval results. Each row shows a query image (first image from the left) and the top-ranked images retrieved by our method. Last two rows show some failure cases.}
\label{fig:rerank}
\end{figure*}

\begin{figure*}[ht!]
\centering
\begin{tabular}{cc}
\begin{tabular}{cl}
\hspace{-2mm}\cfbox{blue}{\includegraphics[height=0.42in]{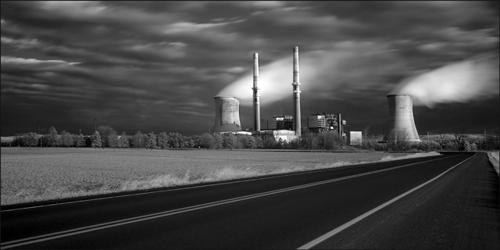}}&\hspace{-2mm}\includegraphics[height=0.41in]{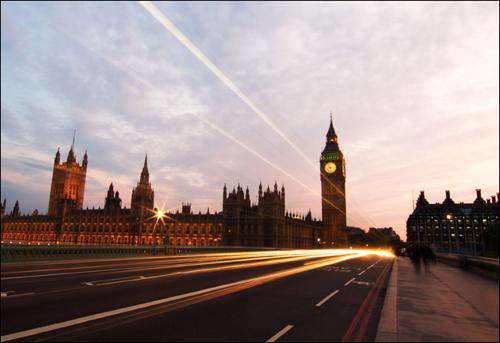}
\hspace{0.1mm}\includegraphics[height=0.41in]{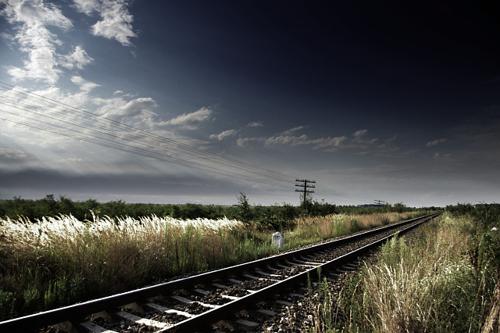}
\hspace{0.1mm}\includegraphics[height=0.41in]{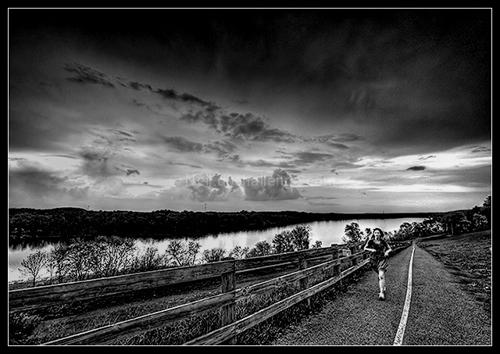}
\hspace{0.1mm}\includegraphics[height=0.41in]{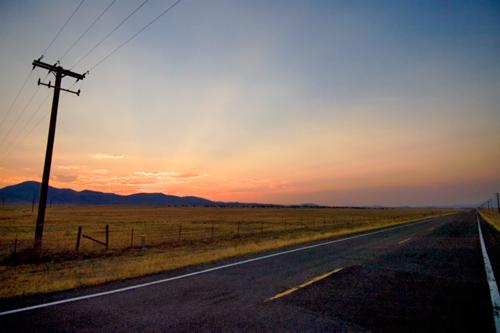}\\
&\hspace{-2mm}\includegraphics[height=0.45in]{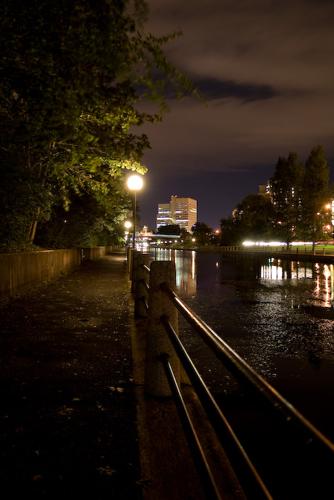}
\hspace{0.1mm}\includegraphics[height=0.45in]{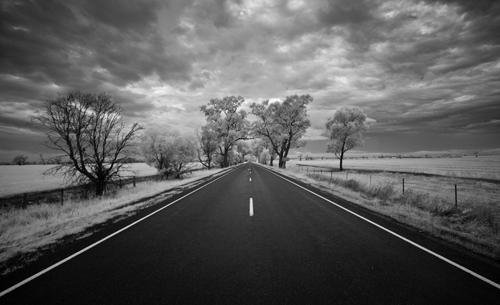}
\hspace{0.1mm}\includegraphics[height=0.45in]{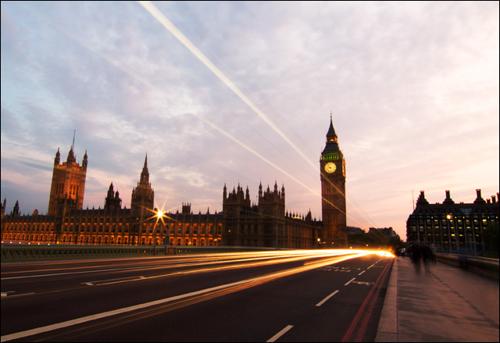}
\hspace{0.1mm}\includegraphics[height=0.45in]{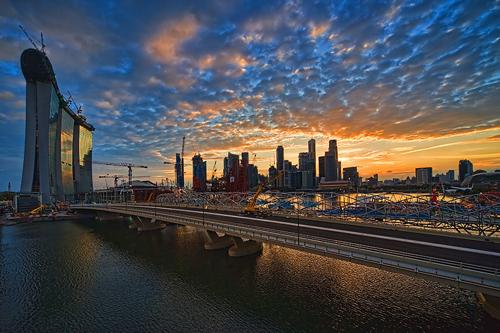}\\
&\hspace{-2mm}\includegraphics[height=0.44in]{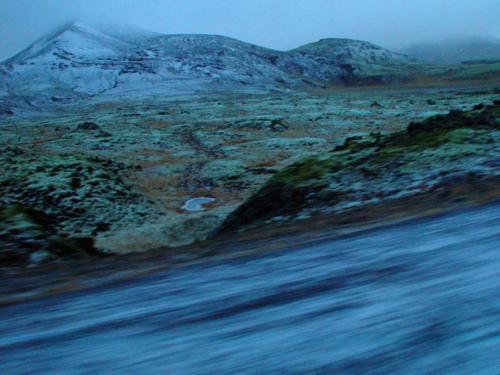}
\hspace{0.1mm}\includegraphics[height=0.44in]{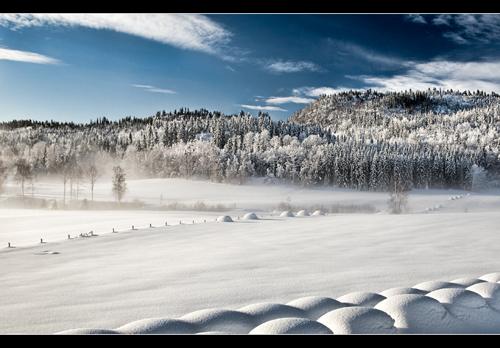}
\hspace{0.1mm}\includegraphics[height=0.44in]{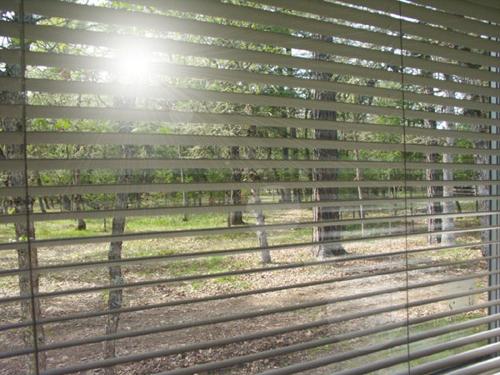}
\hspace{0.1mm}\includegraphics[height=0.44in]{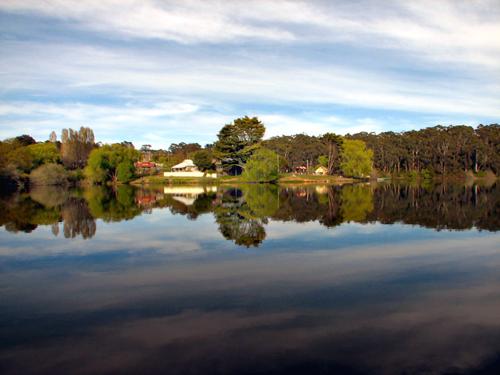}\\
\end{tabular}&\hspace{-2mm}\begin{tabular}{cl}
\hspace{-2mm}\cfbox{blue}{\includegraphics[height=0.467in]{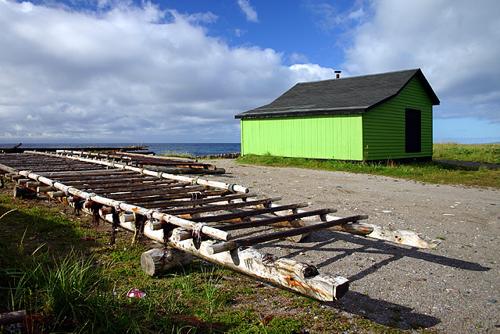}}&\hspace{-2mm}\includegraphics[height=0.47in]{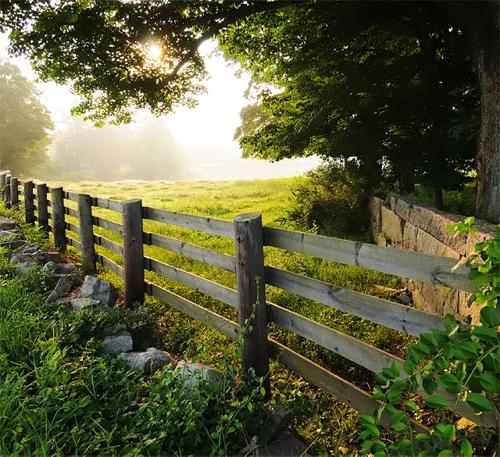}
\hspace{0.1mm}\includegraphics[height=0.47in]{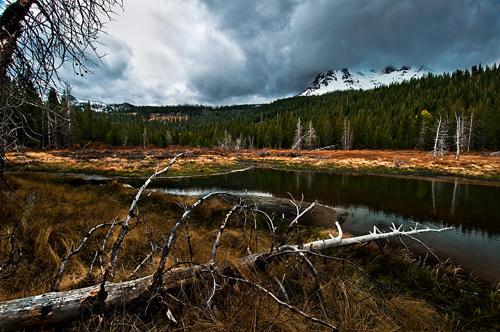}
\hspace{0.1mm}\includegraphics[height=0.47in]{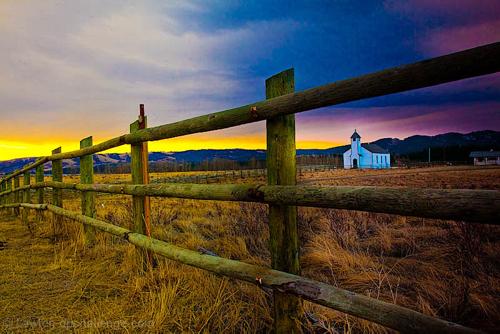}
\hspace{0.1mm}\includegraphics[height=0.47in]{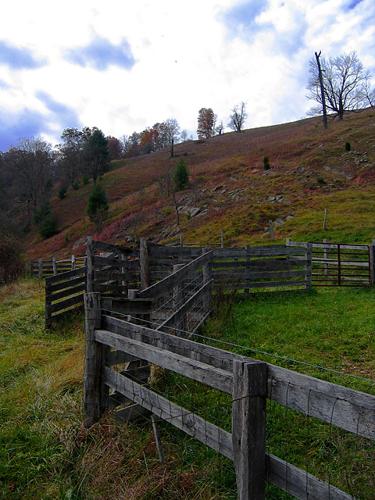}\\
&\hspace{-2mm}\includegraphics[height=0.54in]{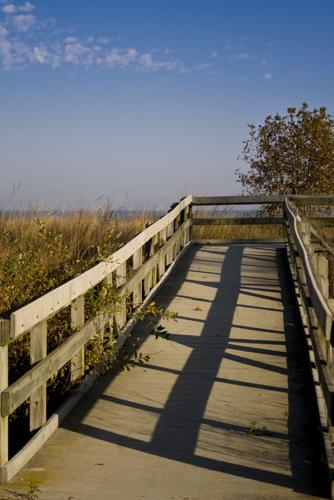}
\hspace{0.1mm}\includegraphics[height=0.54in]{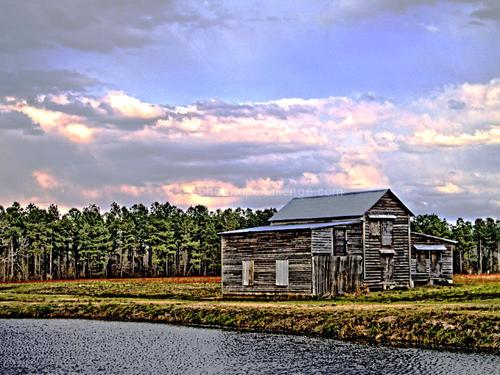}
\hspace{0.1mm}\includegraphics[height=0.54in]{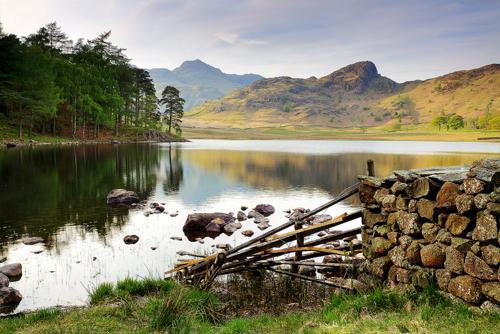}
\hspace{0.1mm}\includegraphics[height=0.54in]{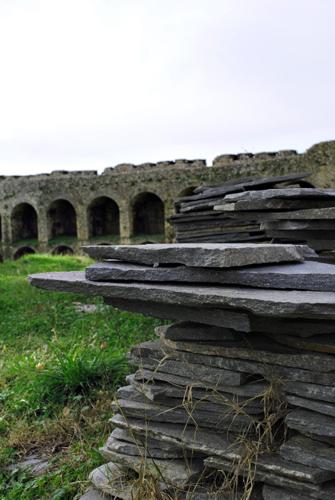}\\
&\hspace{-2mm}\includegraphics[height=0.43in]{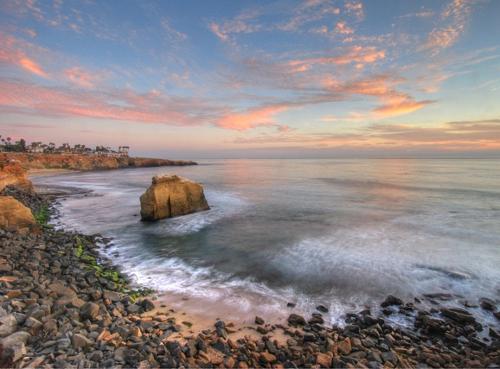}
\hspace{0.1mm}\includegraphics[height=0.43in]{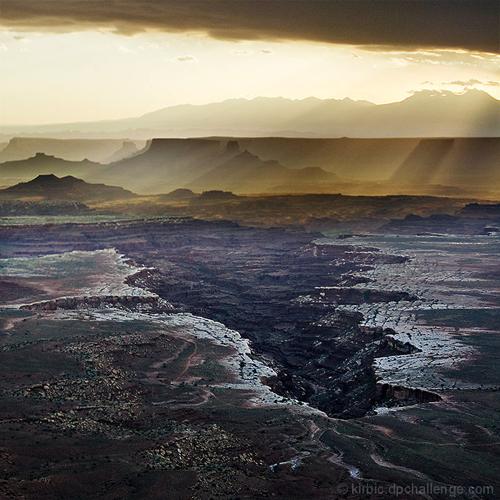}
\hspace{0.1mm}\includegraphics[height=0.43in]{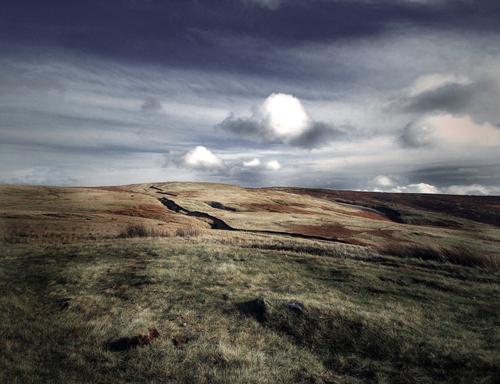}
\hspace{0.1mm}\includegraphics[height=0.43in]{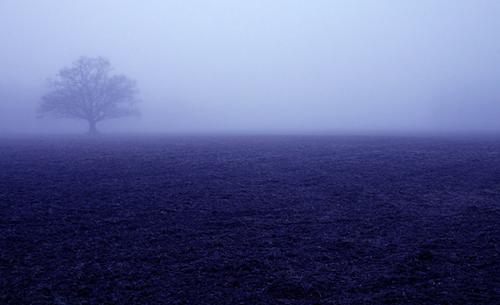}\\
\end{tabular}
\end{tabular}
\vskip 0.1in
\begin{tabular}{cc}
\begin{tabular}{cl}
\hspace{-2mm}\cfbox{blue}{\includegraphics[height=0.52in]{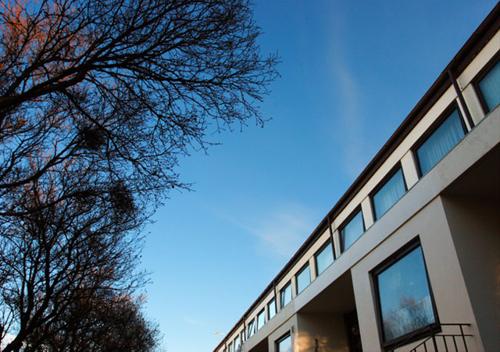}}&\hspace{-2mm}\includegraphics[height=0.58in]{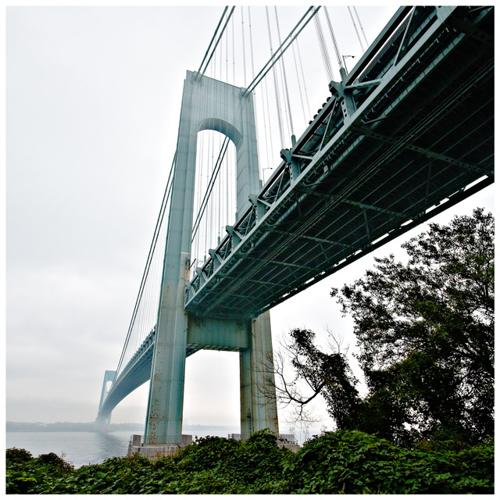}
\hspace{0.1mm}\includegraphics[height=0.58in]{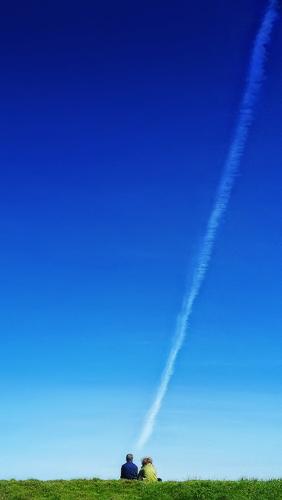}
\hspace{0.1mm}\includegraphics[height=0.58in]{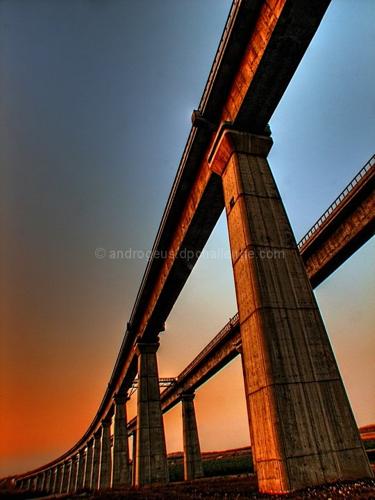}
\hspace{0.1mm}\includegraphics[height=0.58in]{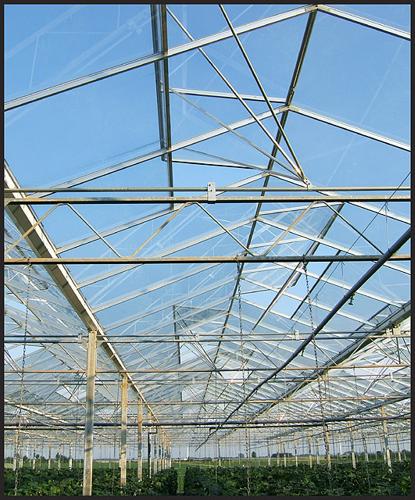}
\hspace{0.1mm}\includegraphics[height=0.58in]{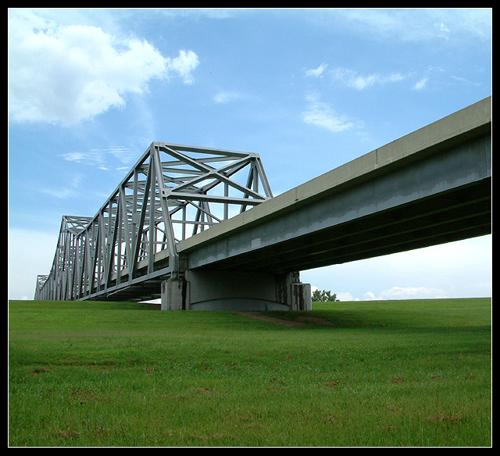}\\
&\hspace{-2mm}\includegraphics[height=0.42in]{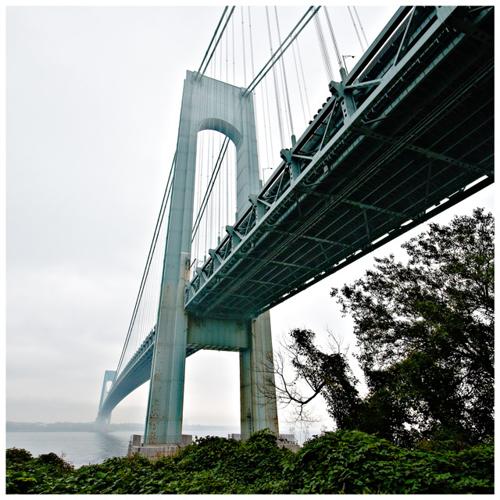}
\hspace{0.1mm}\includegraphics[height=0.42in]{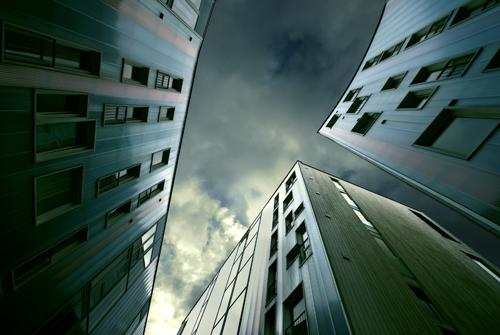}
\hspace{0.1mm}\includegraphics[height=0.42in]{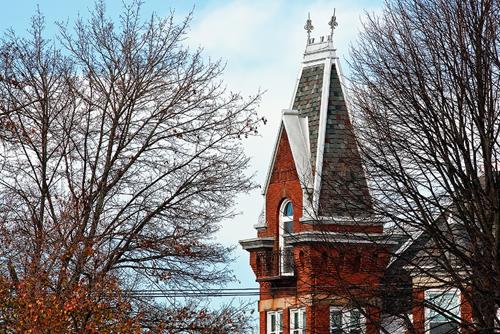}
\hspace{0.1mm}\includegraphics[height=0.42in]{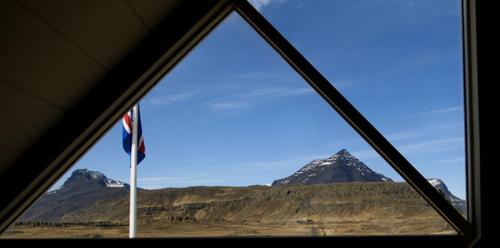}\\
&\hspace{-2mm}\includegraphics[height=0.43in]{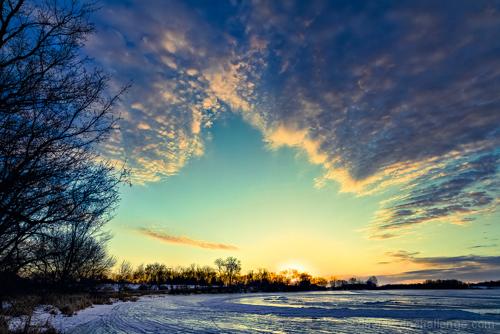}
\hspace{0.1mm}\includegraphics[height=0.43in]{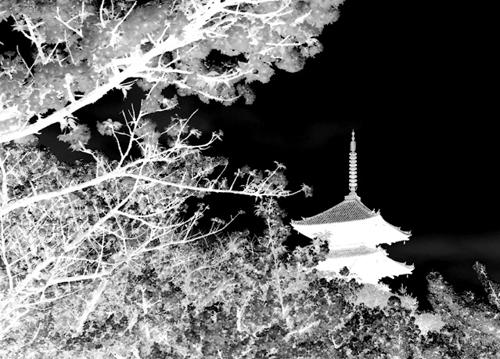}
\hspace{0.1mm}\includegraphics[height=0.43in]{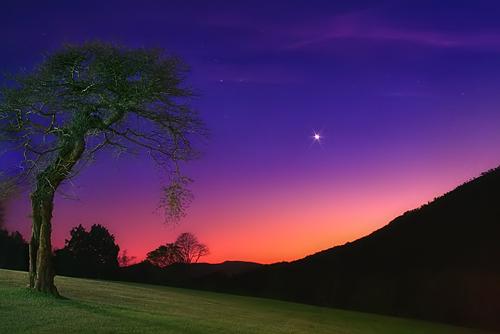}
\hspace{0.1mm}\includegraphics[height=0.43in]{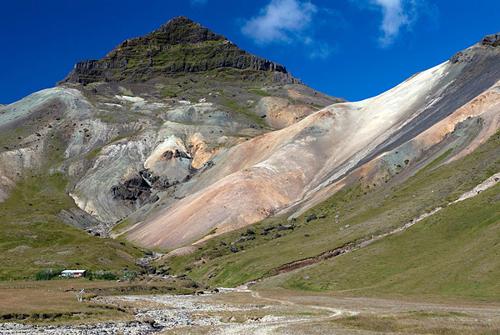}\\
\end{tabular}&\hspace{-2mm}\begin{tabular}{cl}
\hspace{-2mm}\cfbox{blue}{\includegraphics[height=0.51in]{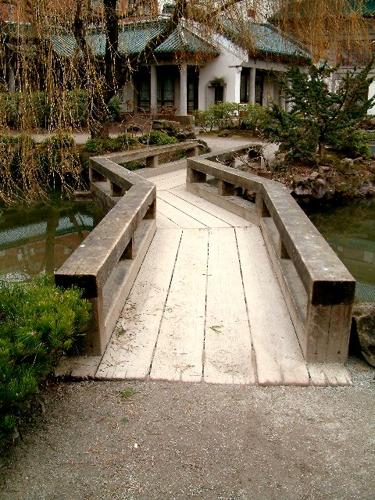}}&\hspace{-2mm}\includegraphics[height=0.51in]{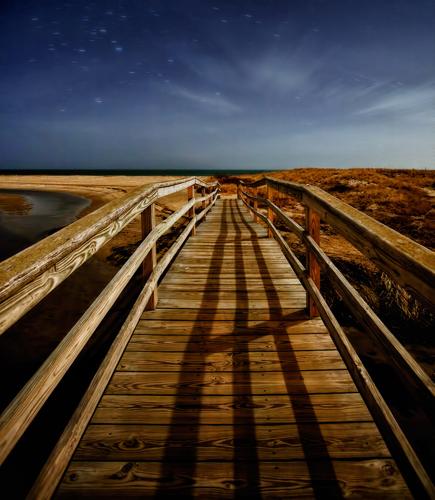}
\hspace{0.1mm}\includegraphics[height=0.51in]{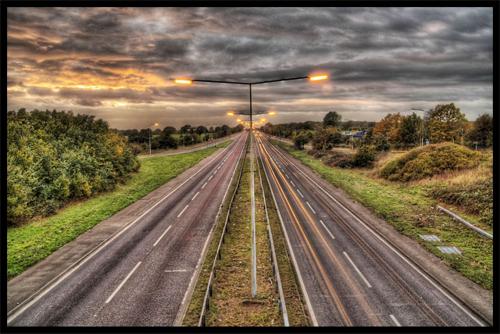}
\hspace{0.1mm}\includegraphics[height=0.51in]{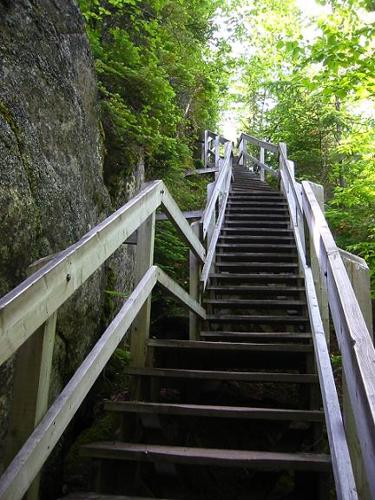}
\hspace{0.1mm}\includegraphics[height=0.51in]{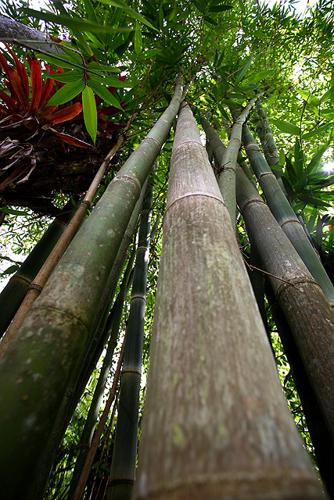}
\hspace{0.1mm}\includegraphics[height=0.51in]{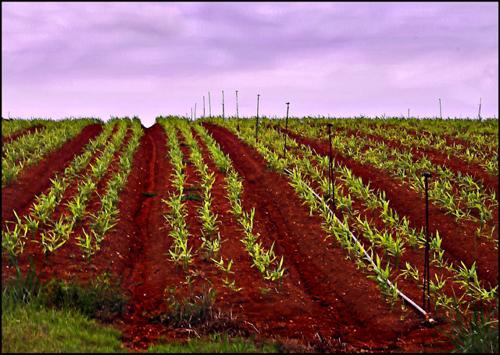}\\
&\hspace{-2mm}\includegraphics[height=0.49in]{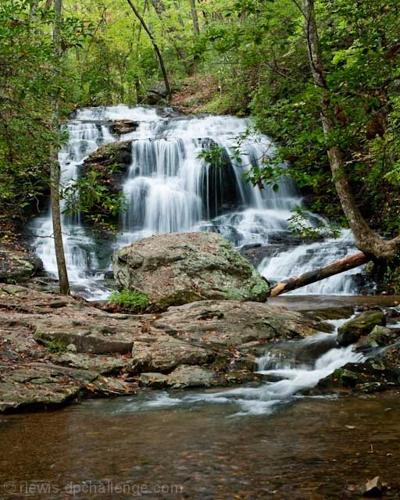}
\hspace{0.1mm}\includegraphics[height=0.49in]{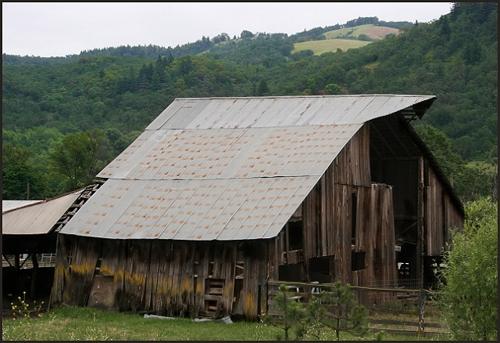}
\hspace{0.1mm}\includegraphics[height=0.49in]{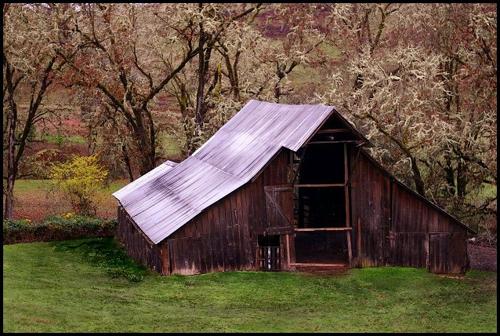}
\hspace{0.1mm}\includegraphics[height=0.49in]{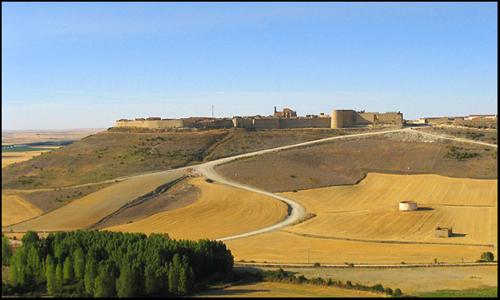}\\
&\hspace{-2mm}\includegraphics[height=0.58in]{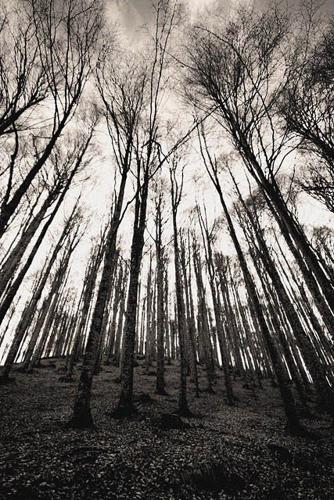}
\hspace{0.1mm}\includegraphics[height=0.58in]{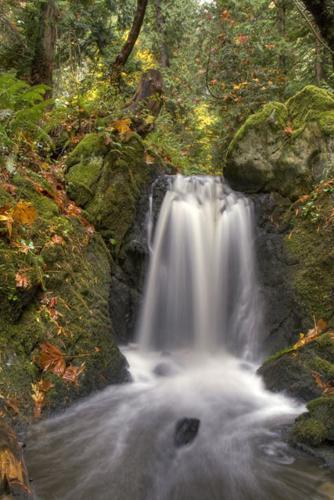}
\hspace{0.1mm}\includegraphics[height=0.58in]{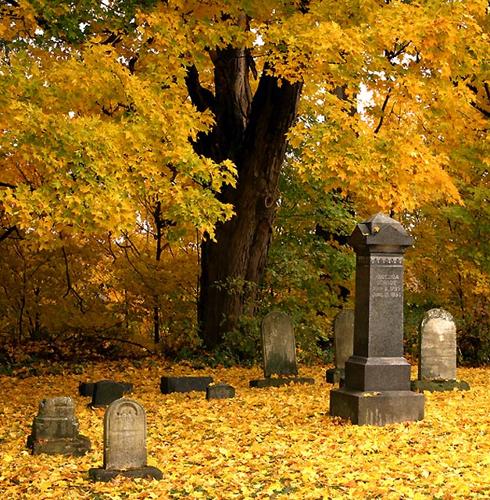}
\hspace{0.1mm}\includegraphics[height=0.58in]{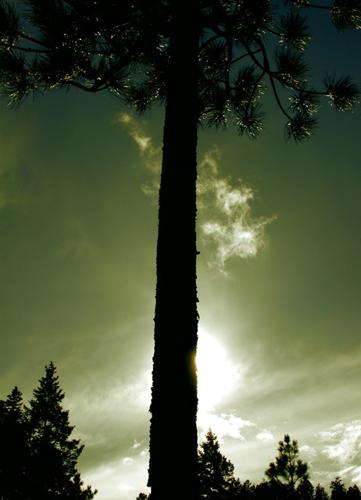}
\hspace{0.1mm}\includegraphics[height=0.58in]{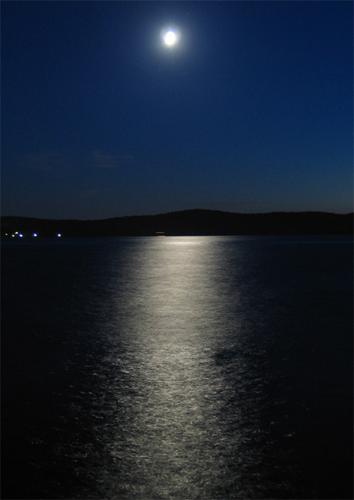}\\
\end{tabular}
\end{tabular}
\caption{Comparison to state-of-the-art retrieval methods. For a query image, we show the top four or five images retrieved by different methods, where each row corresponds to one method. {\bf First row:} Our method. {\bf Second row:} CNN. {\bf Third row:} HOG.}
\label{fig:retrieval1}
\end{figure*}

\subsection{Case Studies}
In our experiments, we use the entire ``landscape'' category of the AVA dataset to study the effectiveness of our new similarity measure. We first run our contour-based VP detection algorithm to detect the dominant VP in the 21,982 images in that category. We only keep those VPs with strength scores higher than 150 for this study because, as mentioned earlier (Section~\ref{sec:5experiments}), detections with low strength scores are often unreliable. If no dominant VP is detected in an image, we simply set the perspective similarity $D_p(I_i, I_j) = 0$.

Figure~\ref{fig:rerank} shows the top-ranked images for various query images in the AVA dataset. It is clear that our method is able to retrieve images with similar content \emph{and} similar viewpoints as the query image. More importantly, the retrieved images exhibit a wide variety in terms of the photographic techniques used, including color, lighting, photographic elements, design principles, etc. Thus, by examining the exemplar images retrieved by our system, amateur photographers may conveniently learn useful techniques to improve the quality of their work. Below we examine a few cases:

\smallskip
\noindent{\it 1$^{st}$ row, red boxes:} This pair of photos highlight the importance of lighting in photography. Instead of taking the picture with an overcast sky, it is better to take the shot close to sunset with a clear sky, as the interplay of sunlight and shadows could make the photo more attractive. Also, it can be desirable to leave more water in the scene to better reflect such interplay.

\smallskip
\noindent{\it 6$^{th}$ row, blue boxes:} These three photos illustrate the different choices of framing and photo composition. While the query image uses a diagonal composition, alternative ways to shoot the bridge scene include using a vertical frame or lowering the camera to include the river.

\smallskip
\noindent{\it 9$^{th}$ row, green boxes:} This case shows an example where photographers sometimes choose unconventional aspect ratios ({\it e.g.}, a wide view) to make the photo more interesting.

\smallskip
\noindent{\it 11$^{th}$ row, yellow boxes:} Compared to the query image, the retrieved image contains more vivid colors (the grass) and texture (the cloud).

\smallskip
The last two rows of Figure~\ref{fig:rerank} show the typically failure cases of our method, in which we also plot the edges corresponding to the detected VP in the query image. In the first example, our VP detection method fails to detect the true dominant VP in the image. For the second example, while the detection is successful, our retrieval system is unable to find images with similar viewpoints. In real-world applications, however, we expect the size of the image database to be much larger than our experimental database (with $\sim$20K images), and the algorithm should be able to retrieve valid results. 

\subsection{Comparison to the State-of-the-Art}
\label{sec:retrieval-exp}

We compare our method to two popular retrieval systems, which are based on the HOG features~\cite{DalalT05, FelzenszwalbGMR10} and the CNN features, respectively. While many image retrieval methods exist in the literature, we choose these two because (i) the CNN features have been recently shown to achieve state-of-the-art performance on semantic image retrieval and (ii) similar to our method, the HOG features are known to be sensitive to image edges, thus providing a good baseline for comparison.

For {\bf HOG}, We represent each image with a rigid grid-like HOG feature $\x_i$~\cite{DalalT05, FelzenszwalbGMR10}. As suggested in~\cite{ShrivastavaMGE11}, we limit its dimensionality to roughly $5K$ by resizing the images to $150\times 100$ or $100\times 150$ and using a cell size of 8 pixels. The feature vectors are normalized by subtracting the mean: $\x_i = \x_i - mean(\x_i)$. We use the cosine distance as the similarity measure. For {\bf CNN}, we directly use $D_s(I_j, I_j)$ discussed in Section~\ref{sec:view-specific} as the final matching score. Obviously, our method reduces to {\bf CNN} if we set $\gamma_1=\gamma_2=0$ in Eq.~\eqref{eq:perspective-sim}.

Figure~\ref{fig:retrieval1} shows the best matches retrieved by all systems for various query images. Both {\bf CNN} and our method are able to retrieve semantically relevant images. However, the images retrieved by {\bf CNN} vary significantly in terms of the viewpoint. In contrast, out method retrieves images with similar viewpoints. While {\bf HOG} is somewhat sensitive to the edge orientations (see the first and fourth examples in Figure~\ref{fig:retrieval1}), our method more effectively captures the viewpoints and perspective effects.

\smallskip
\noindent{\bf Quantitative human subject evaluation.} We further perform a quantitative evaluation on the performance of our retrieval method. Unlike traditional image retrieval benchmarks, currently there is no dataset with ground truth composition (\ie, viewpoint) labels available. In view of this barrier, we instead conducted a user study which asked participants to manually compare the performance of our method with that of {\bf CNN} based on their ability to retrieve images that have \emph{similar semantics and similar viewpoints}. Note that we have excluded {\bf HOG} from this study because (i) it performs substantially worse than our method and {\bf CNN}, and (ii) we are particularly interested in the effectiveness of the new perspective similarity measure $D_p$.

In this study, a collection of 200 query images (with VP strength scores higher than 150) are randomly selected from our new dataset of 1,316 images that each containing a dominant VP (Section~\ref{sec:dataset}). At our user study website, each participant is assigned with a subset of 30 randomly selected query images. For each query, we show the top-8 images retrieved by both systems and ask the participant to rank the performance of the two systems.  To avoid any biases, no information about the two systems was provided during the study. Further, we randomly shuffled the order in which the results of the two systems are shown on each page.

We recruited 10 participants to this study, mostly graduate students with some basic photography knowledge. Overall, participants ranked our system better 76.7\% of the time, whereas {\bf CNN} received higher rankings only 23.3\% of the time. This outcome suggests our system significantly outperforms the state-of-the-art for the viewpoint-specific image retrieval task.

%% file: 7conclusion.tex
\section{Conclusions and Discussions}

In this paper, we study an intriguing problem of detecting the dominant VP in natural landscape images. We develop a new VP detection method, which combines a contour-based edge detector with J-Linkage, and show that it outperforms state-of-the-art methods on a new ground truth dataset. The detected VPs and the associated image elements provides valuable information about the photo composition. As an application of our method, we develop a novel viewpoint-specific image retrieval system which can potentially provide useful on-site feedback to photographers.

\subsection{Performance on Architecture Photos}

While the primary focus of this paper is on natural scenes, we applied the proposed methods to images of structured man-made environments, \eg, architecture photos. For a dominant VP in an architecture photo, a large number of strong edges often converge to it. How does the performance of our dominant VP detection and selection methods compare to the state-of-the-art in such scenarios? Given that there are often more than one VP in architecture photos, would our image retrieval method perform as successfully under such complexity?

\smallskip
\noindent{\bf AVA architecture dataset.} To study the performance of our methods on architecture photos, we again resort to the AVA dataset and use the 12,026 images under the category ``architecture''. Following the same annotation procedure as described in Section~\ref{sec:dataset}, for each image, we determine whether it contains a dominant VP and, if so, label its location. For consistency, we still restrict ourselves to the detection of a single dominant VP and do not consider images with two or more VPs carrying similar visual importance. As a result, we collected a total of 2,401 images with ground-truth annotations.

\begin{figure}[t]
\centering
\begin{tabular}{cc}
\hspace{-2mm}\includegraphics[height =1.42in]{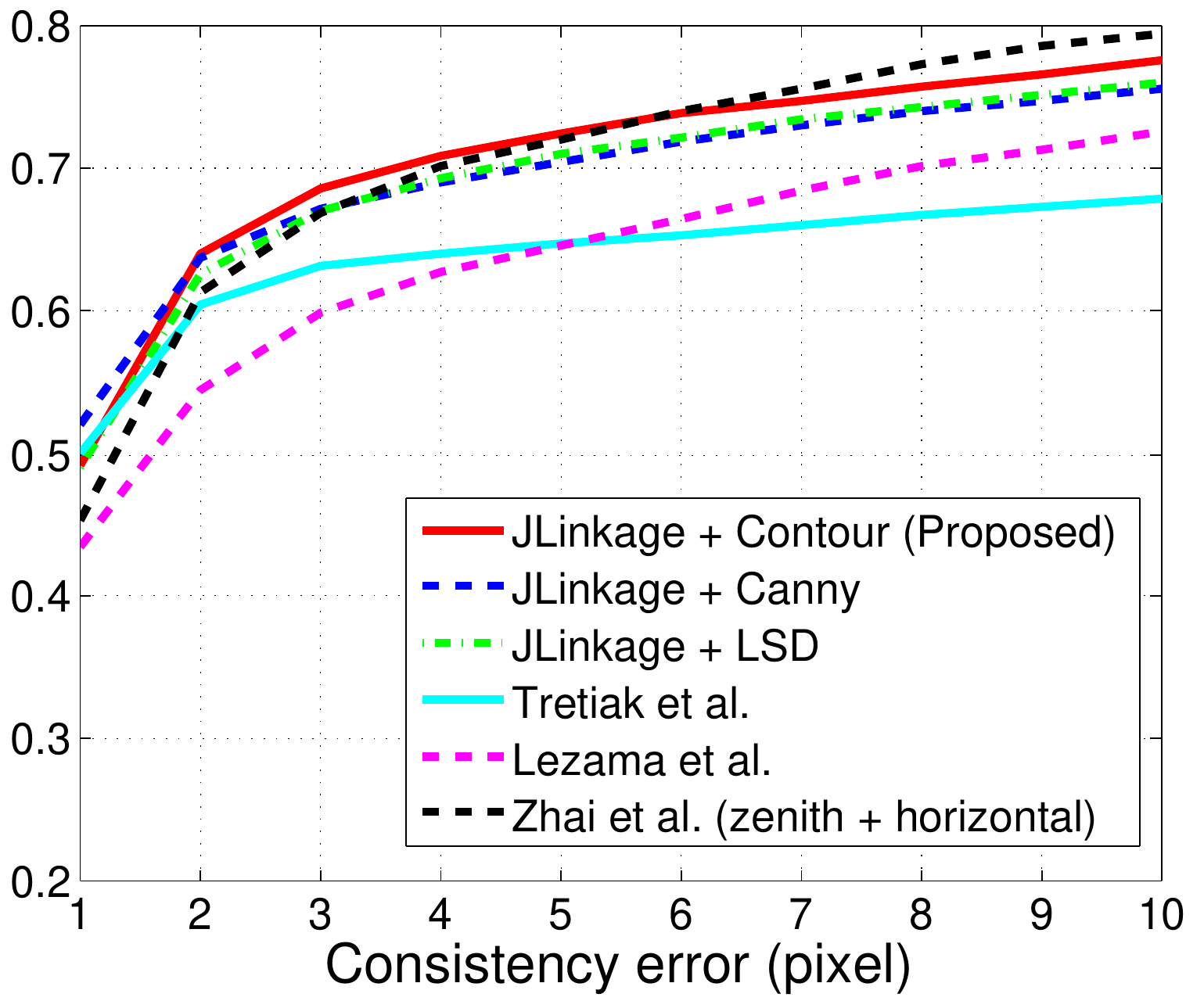} &
\hspace{-2mm}\includegraphics[height =1.42in]{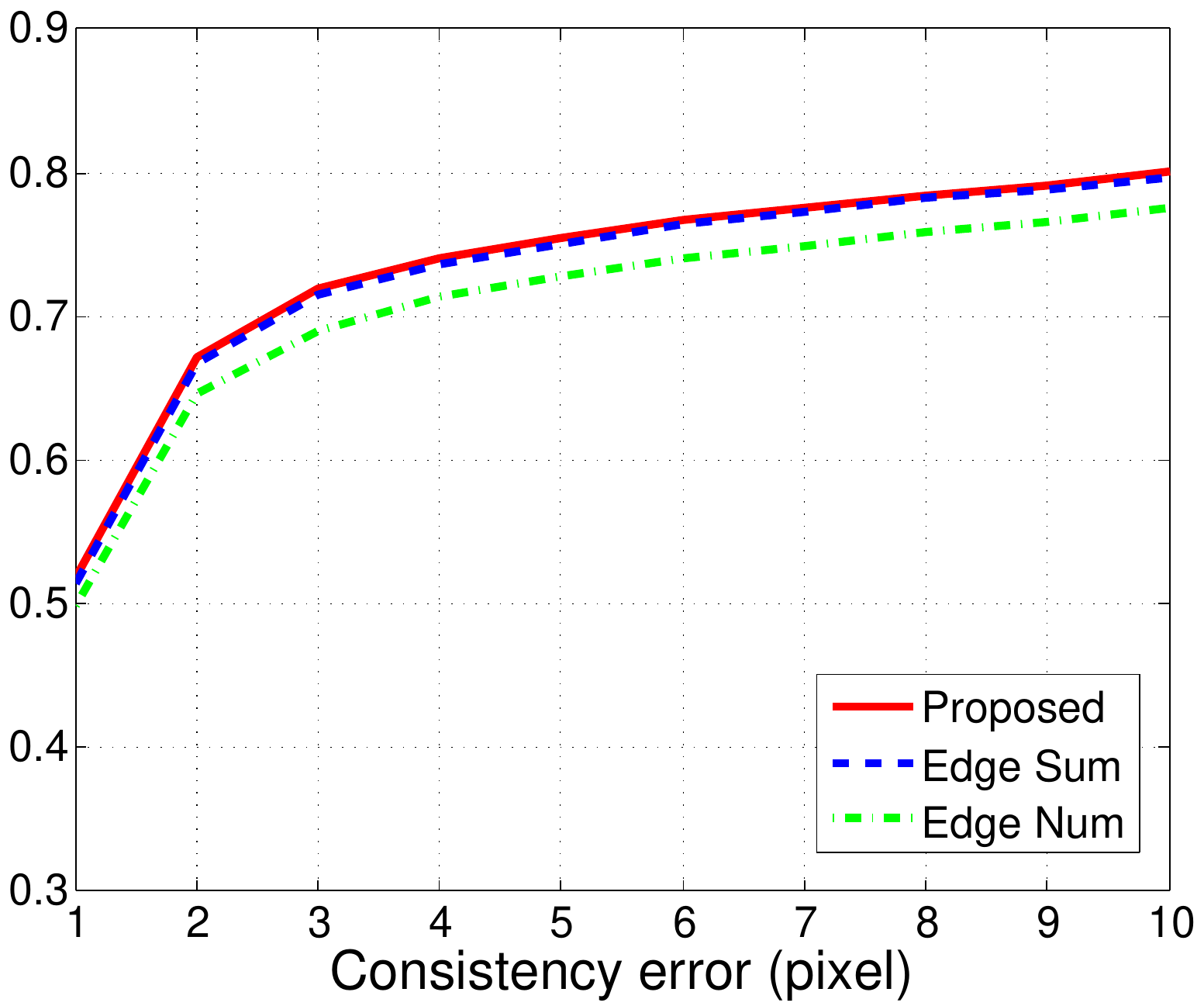} \\
(a)  & (b) 
\end{tabular}
\caption{Experiments on architecture photos. (a) Dominant VP detection. (b) Dominant VP selection.}
\label{fig:result-arch}
\end{figure}

\smallskip
\noindent{\bf Dominant VP detection and selection experiment.} Using the same experiment protocol and parameter settings as described in Section~\ref{sec:detection-exp}, we compare our contour-based VP detection method to the state-of-the-art. As one can see in Figure~\ref{fig:result-arch}(a), our method achieves competitive performance on the architecture photos. Recall that our experiment setting actually favors Zhai~\etal~\cite{ZhaiWJ16} in that we consider both the zenith VP and top-ranked horizontal VP detected by~\cite{ZhaiWJ16} in the evaluation. We also note that, in comparison to Figure~\ref{fig:result-detection}, the gaps in performance among the three edge detection methods appear to be smaller on architecture photos. This finding actually highlights the importance of exploring global structures in natural scenes in which strong edges are often absent. 

\begin{figure*}[ht!]
\centering
\begin{tabular}{cc}
\begin{tabular}{cl}
\hspace{-2mm}\cfbox{blue}{\includegraphics[height=0.44in]{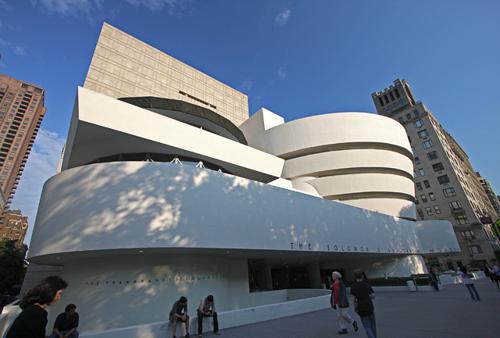}}&\hspace{-2mm}\includegraphics[height=0.44in]{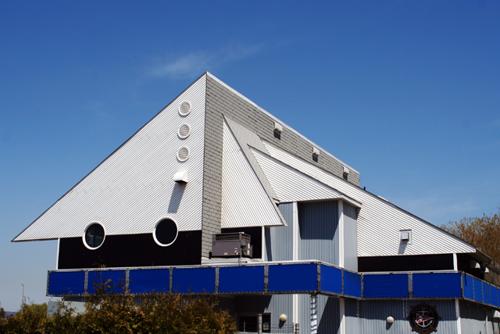}
\hspace{0.1mm}\includegraphics[height=0.44in]{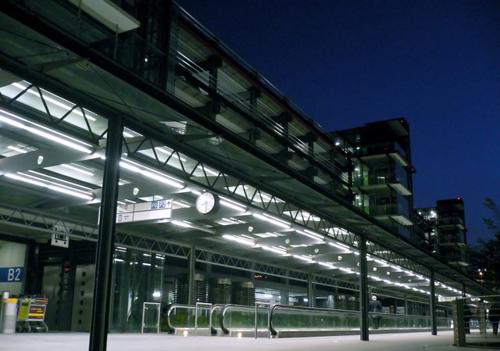}
\hspace{0.1mm}\includegraphics[height=0.44in]{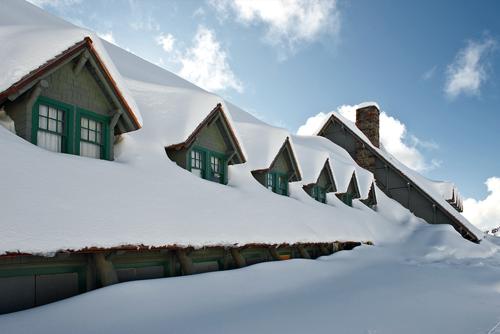}
\hspace{0.1mm}\includegraphics[height=0.44in]{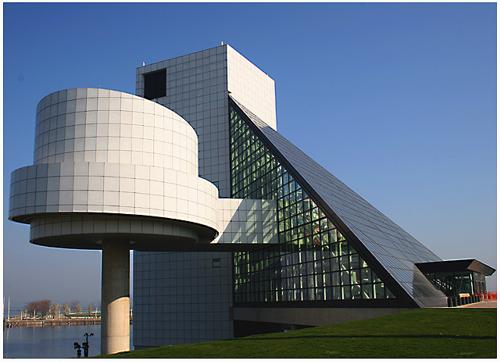}\\
&\hspace{-2mm}\includegraphics[height=0.48in]{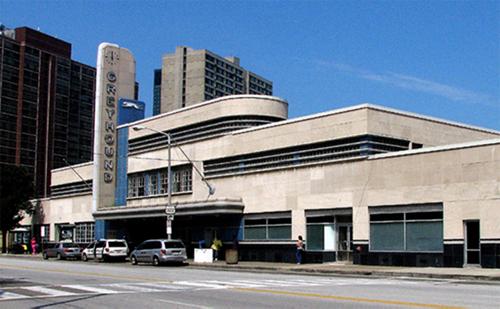}
\hspace{0.1mm}\includegraphics[height=0.48in]{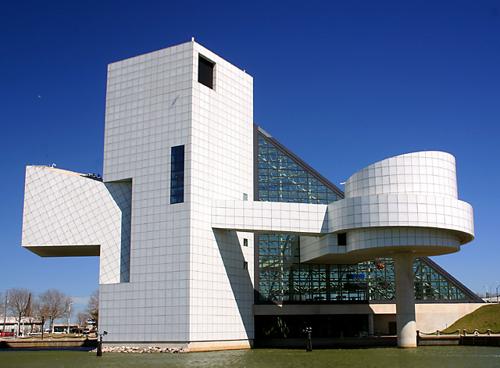}
\hspace{0.1mm}\includegraphics[height=0.48in]{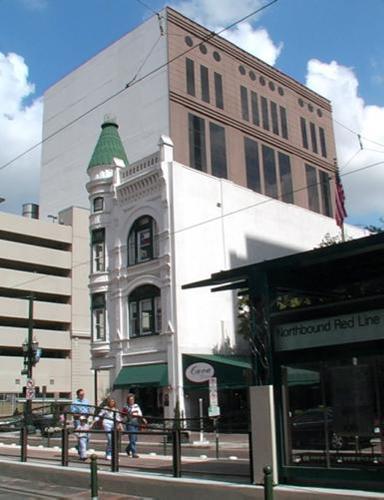}
\hspace{0.1mm}\includegraphics[height=0.48in]{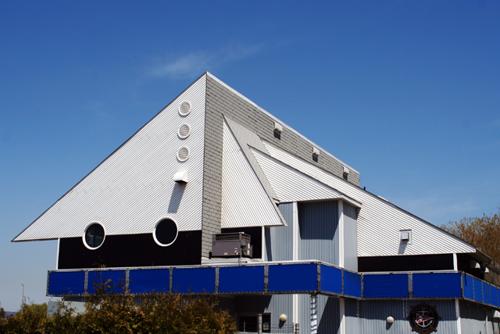}\\
&\hspace{-2mm}\includegraphics[height=0.42in]{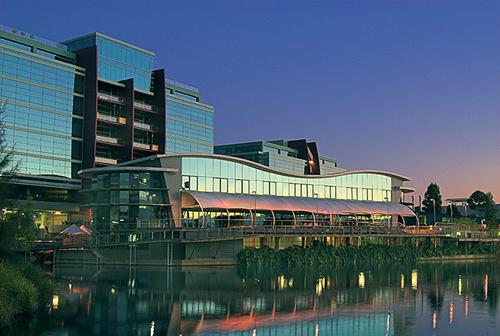}
\hspace{0.1mm}\includegraphics[height=0.42in]{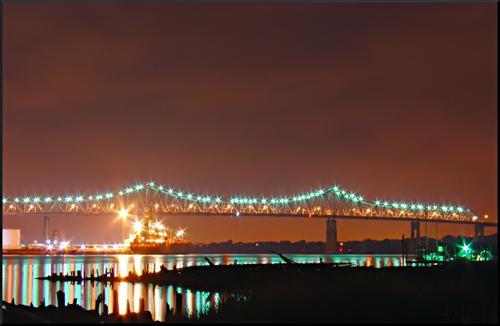}
\hspace{0.1mm}\includegraphics[height=0.42in]{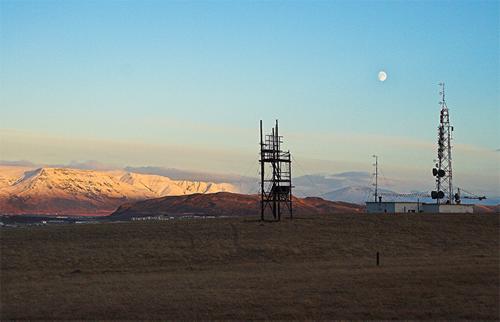}
\hspace{0.1mm}\includegraphics[height=0.42in]{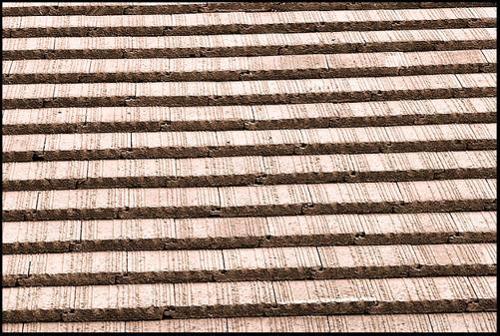}\\
\end{tabular}&\hspace{-2mm}\begin{tabular}{cl}
\hspace{-2mm}\cfbox{blue}{\includegraphics[height=0.43in]{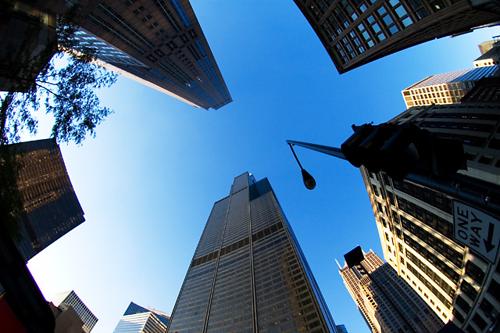}}&\hspace{-2mm}\includegraphics[height=0.43in]{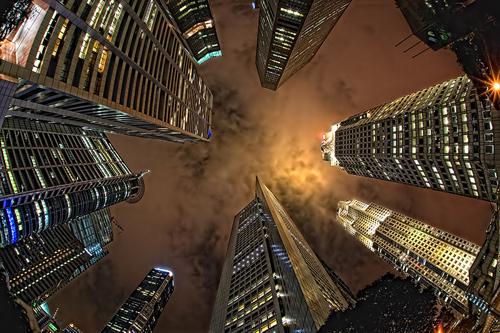}
\hspace{0.1mm}\includegraphics[height=0.43in]{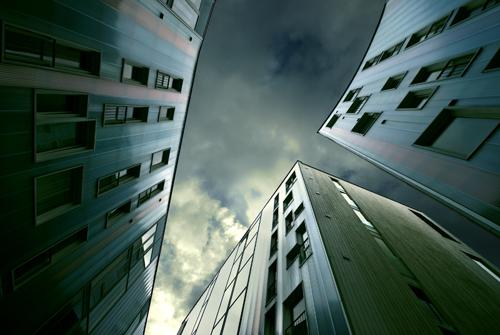}
\hspace{0.1mm}\includegraphics[height=0.43in]{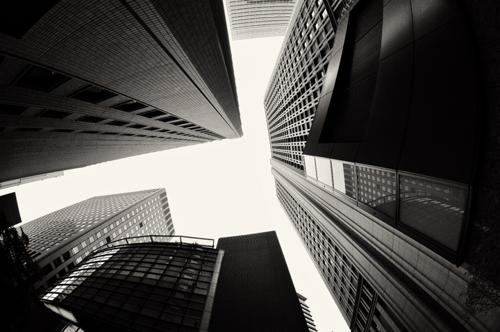}
\hspace{0.1mm}\includegraphics[height=0.43in]{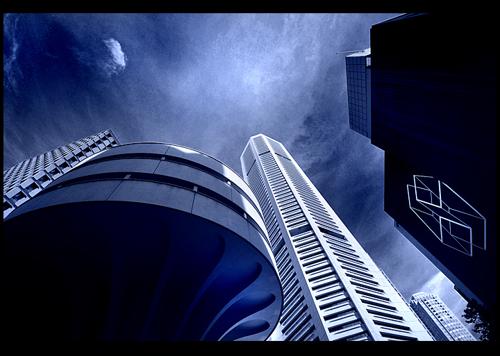}\\
&\hspace{-2mm}\includegraphics[height=0.45in]{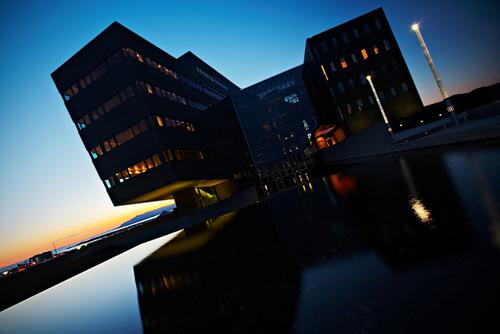}
\hspace{0.1mm}\includegraphics[height=0.45in]{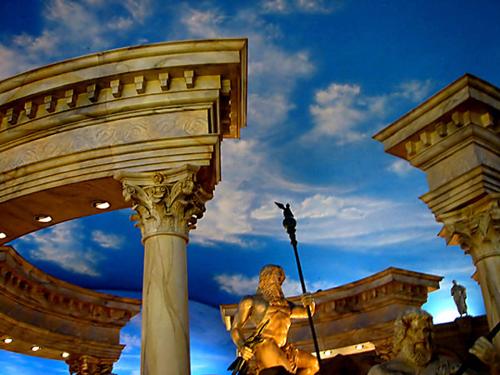}
\hspace{0.1mm}\includegraphics[height=0.45in]{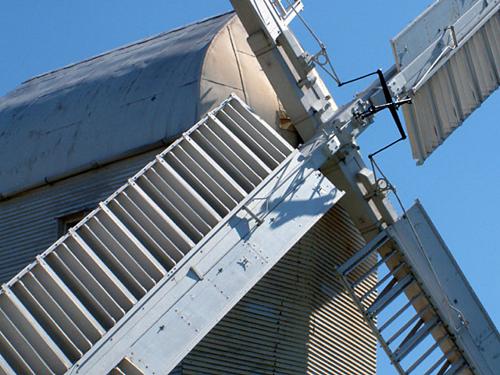}
\hspace{0.1mm}\includegraphics[height=0.45in]{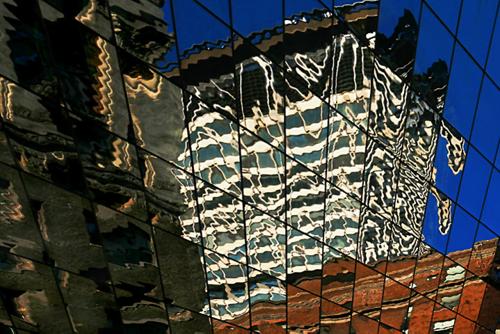}\\
&\hspace{-2mm}\includegraphics[height=0.48in]{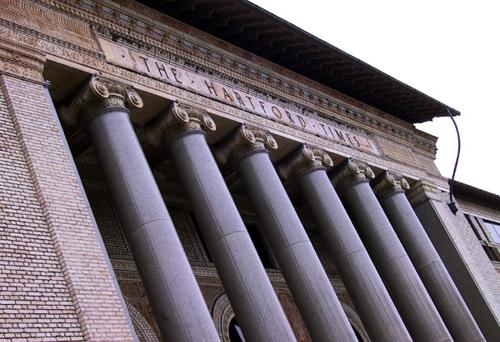}
\hspace{0.1mm}\includegraphics[height=0.48in]{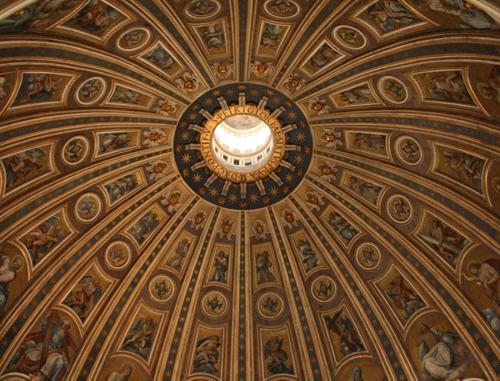}
\hspace{0.1mm}\includegraphics[height=0.48in]{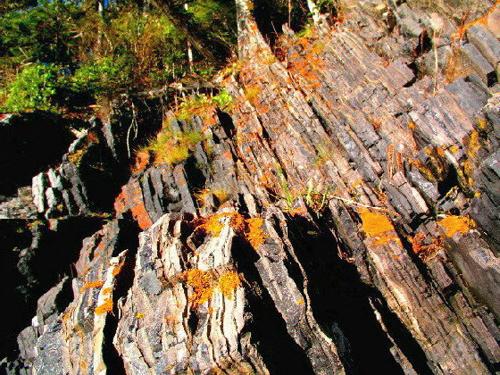}
\hspace{0.1mm}\includegraphics[height=0.48in]{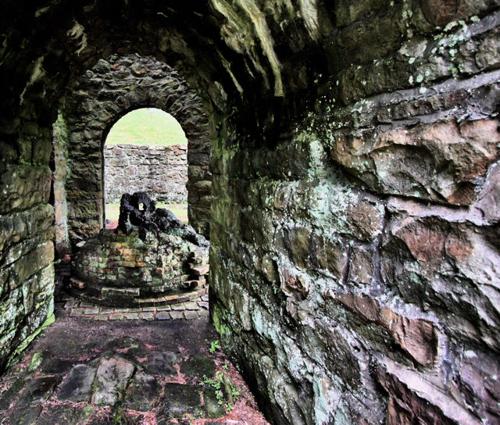}\\
\end{tabular}
\end{tabular}
\vskip 0.1in
\begin{tabular}{cc}
\begin{tabular}{cl}
\hspace{-2mm}\cfbox{blue}{\includegraphics[height=0.54in]{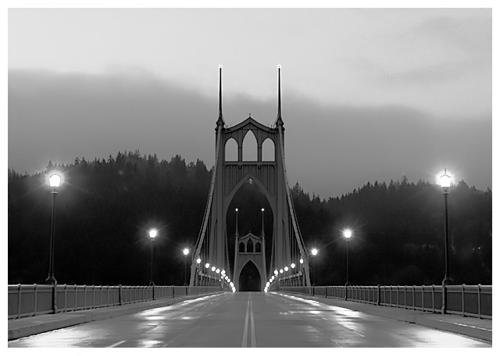}}&\hspace{-2mm}\includegraphics[height=0.54in]{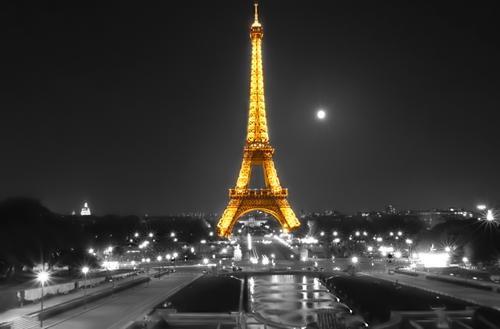}
\hspace{0.1mm}\includegraphics[height=0.54in]{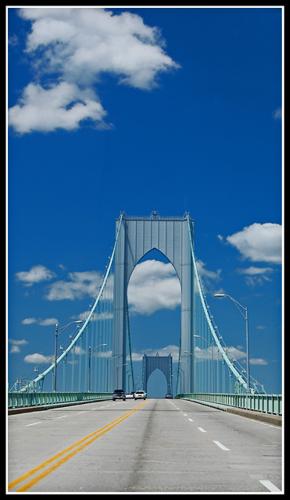}
\hspace{0.1mm}\includegraphics[height=0.54in]{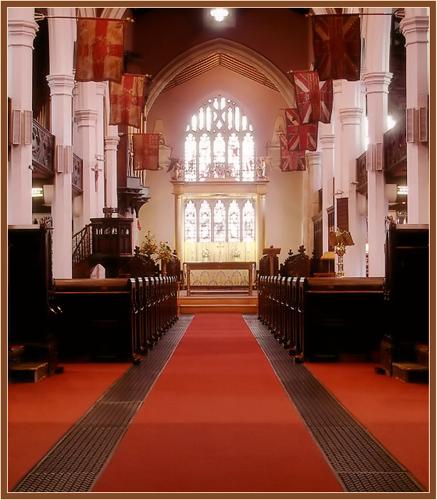}
\hspace{0.1mm}\includegraphics[height=0.54in]{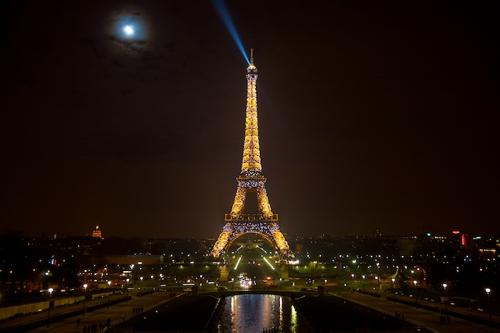}\\
&\hspace{-2mm}\includegraphics[height=0.43in]{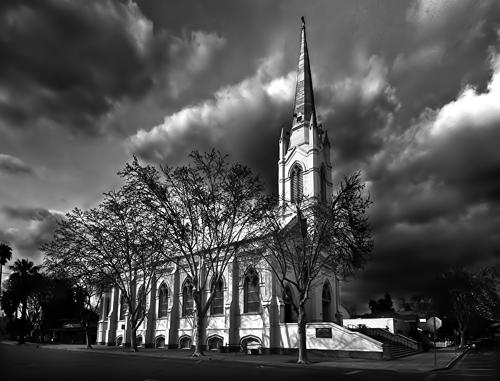}
\hspace{0.1mm}\includegraphics[height=0.43in]{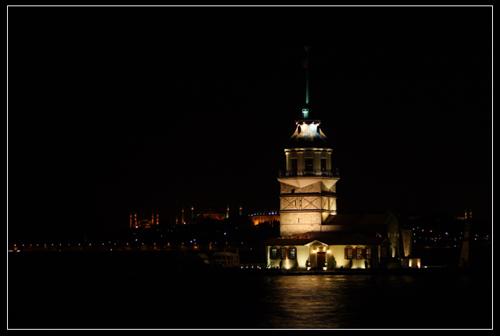}
\hspace{0.1mm}\includegraphics[height=0.43in]{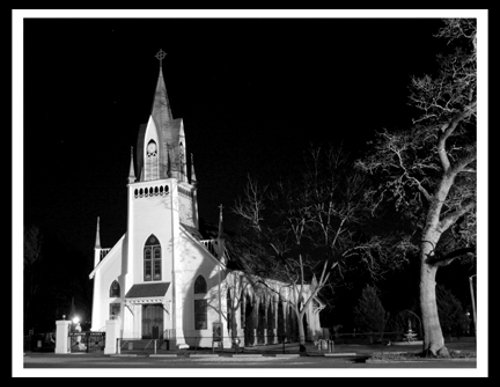}
\hspace{0.1mm}\includegraphics[height=0.43in]{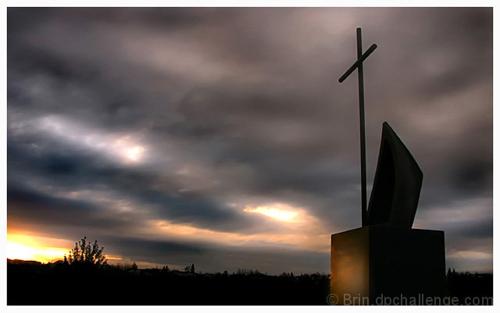}\\
&\hspace{-2mm}\includegraphics[height=0.41in]{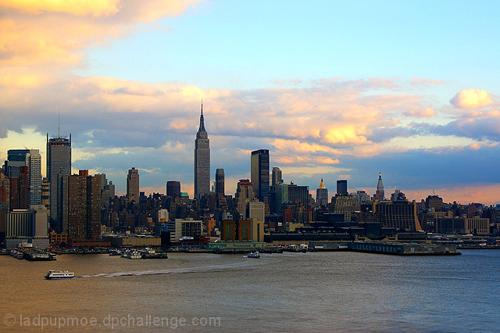}
\hspace{0.1mm}\includegraphics[height=0.41in]{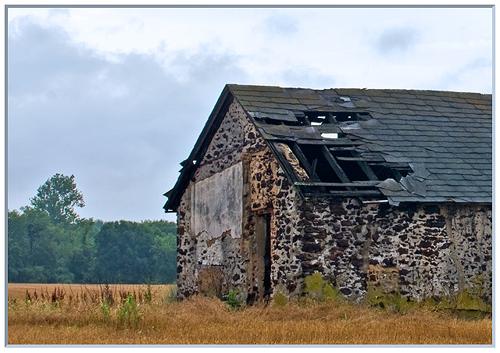}
\hspace{0.1mm}\includegraphics[height=0.41in]{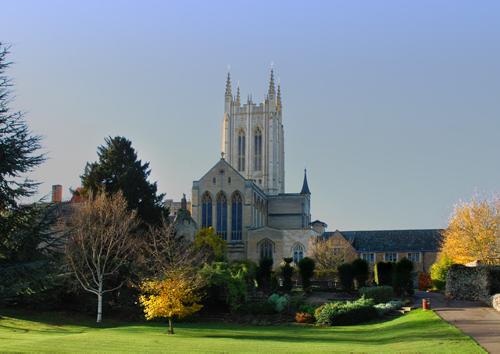}
\hspace{0.1mm}\includegraphics[height=0.41in]{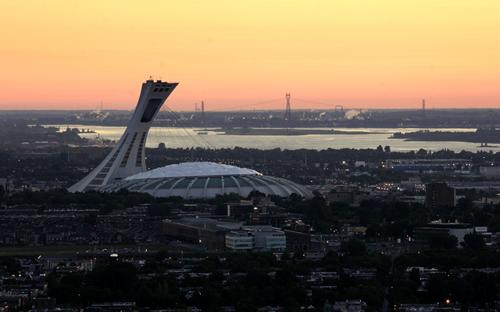}\\
\end{tabular}&\hspace{-2mm}\begin{tabular}{cl}
\hspace{-2mm}\cfbox{blue}{\includegraphics[height=0.58in]{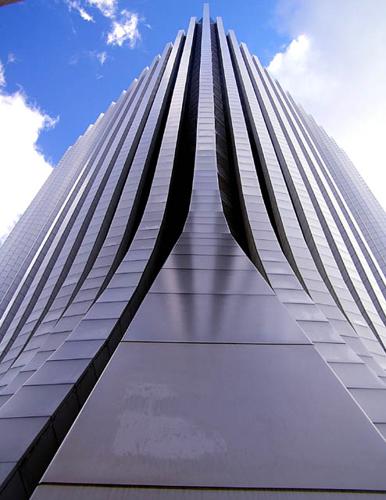}}&\hspace{-2mm}\includegraphics[height=0.58in]{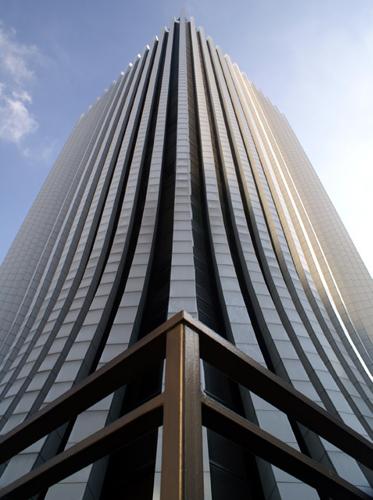}
\hspace{0.1mm}\includegraphics[height=0.58in]{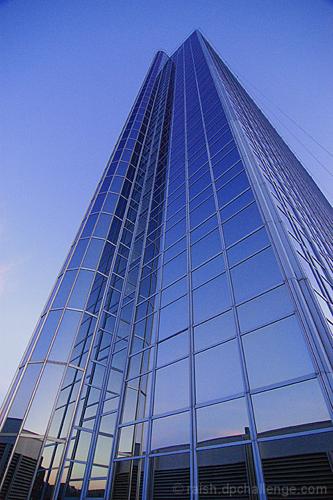}
\hspace{0.1mm}\includegraphics[height=0.58in]{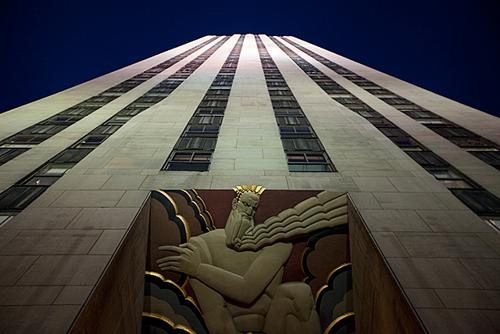}
\hspace{0.1mm}\includegraphics[height=0.58in]{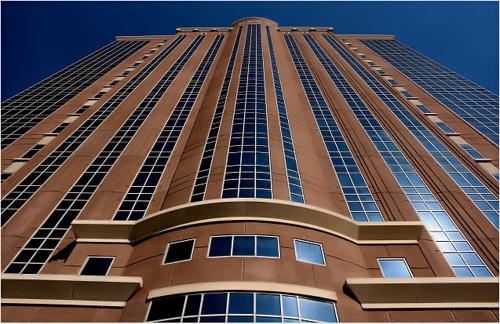}\\
&\hspace{-2mm}\includegraphics[height=0.54in]{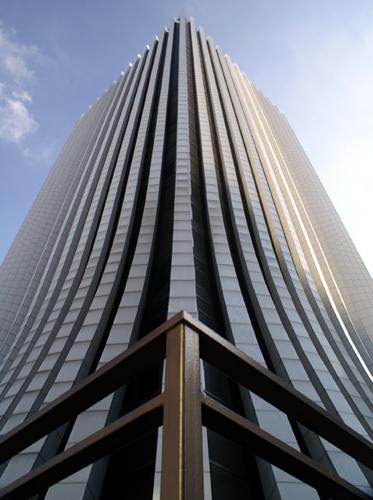}
\hspace{0.1mm}\includegraphics[height=0.54in]{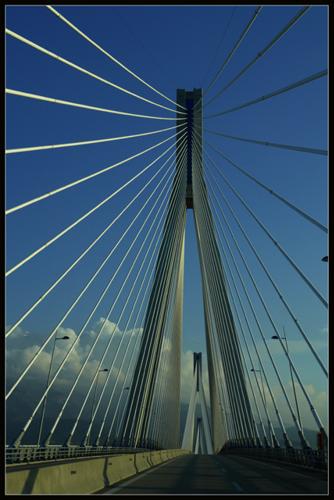}
\hspace{0.1mm}\includegraphics[height=0.54in]{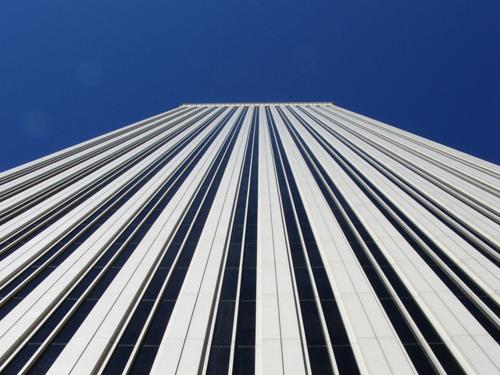}
\hspace{0.1mm}\includegraphics[height=0.54in]{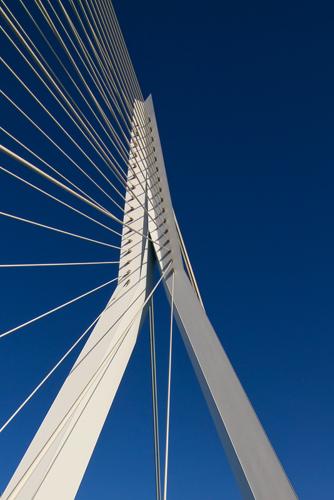}
\hspace{0.1mm}\includegraphics[height=0.54in]{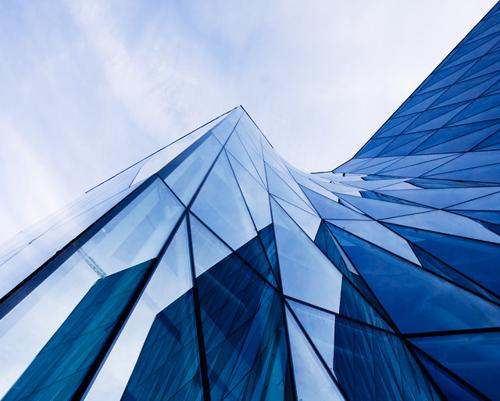}\\
&\hspace{-2mm}\includegraphics[height=0.58in]{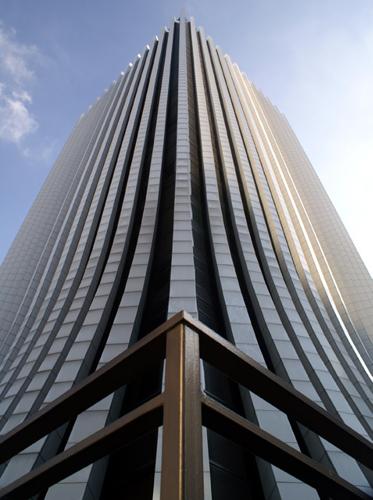}
\hspace{0.1mm}\includegraphics[height=0.58in]{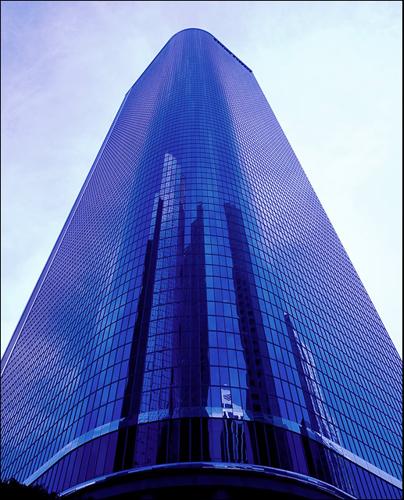}
\hspace{0.1mm}\includegraphics[height=0.58in]{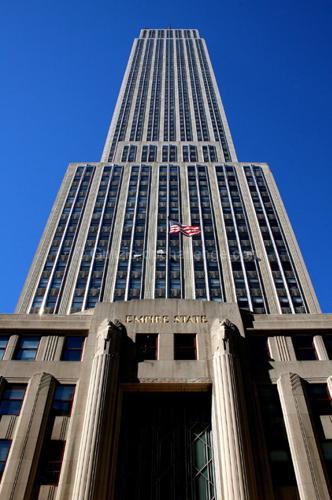}
\hspace{0.1mm}\includegraphics[height=0.58in]{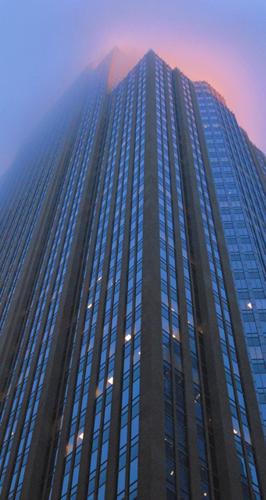}
\hspace{0.1mm}\includegraphics[height=0.58in]{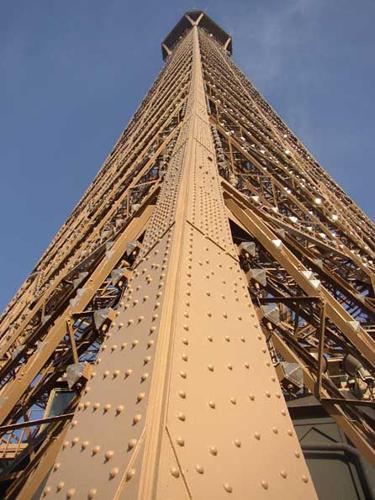}\\
\end{tabular}
\end{tabular}
\caption{Comparison to state-of-the-art retrieval methods on architecture photos. For a query image, we show the top four or five images retrieved by different methods, where each row corresponds to one method. {\bf First row:} Our method. {\bf Second row:} CNN. {\bf Third row:} HOG.}
\label{fig:retrieval-arch}
\end{figure*}

We have also explored the effectiveness of the strength measure proposed in Section~\ref{sec:selection} on architecture photos. As shown in Figure~\ref{fig:result-arch}(b), our proposed measure and {\tt Edge Sum} achieve very similar performance in selecting the dominant VP, and both outperform {\tt Edge Num}. The result suggests that while considering edge length increases the accuracy, the impact of the distance from edge to the VP appears to be insignificant. This observation may be because, when there is a large number of edges, the length of the converging edges dominates the VP selection.

\smallskip
\noindent{\bf Viewpoint-specific image retrieval experiment.} For this experiment, we use the entire ``architecture'' category of the AVA dataset. Using the same experiment protocol and parameter settings as in Section~\ref{sec:retrieval}, we compare our method to {\bf HOG} and {\bf CNN}. 
Figure~\ref{fig:retrieval-arch} shows the retrieval results for various query images in the AVA dataset.

\begin{figure}[t]
\centering
\begin{tabular}{c|l}
\hspace{-1mm}\includegraphics[height =0.74in]{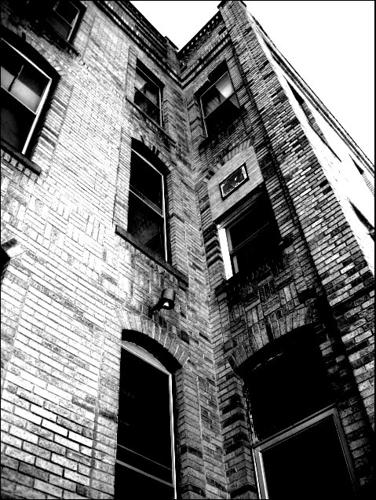}&
\hspace{-1mm}\includegraphics[height =0.74in]{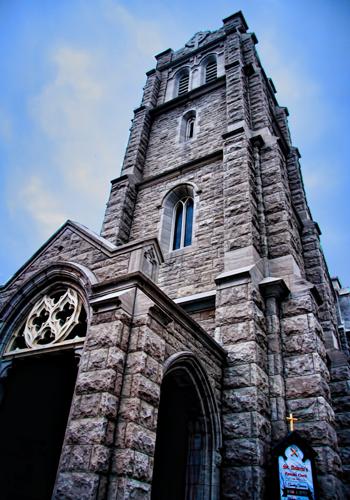}\hspace{0.1mm}
\includegraphics[height =0.74in]{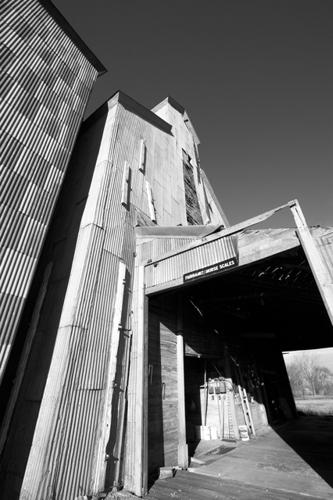}\hspace{0.1mm}
\includegraphics[height =0.74in]{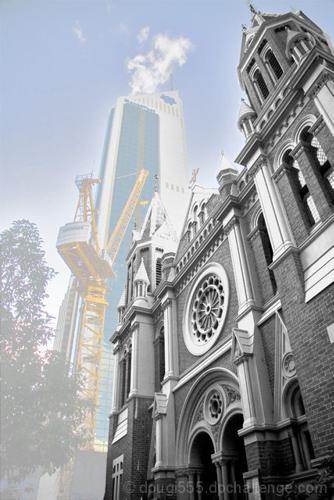}\hspace{0.1mm}
\includegraphics[height =0.74in]{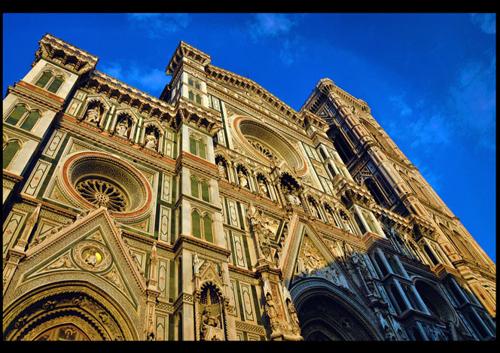}\hspace{0.1mm}
\\
\hspace{-1mm}\includegraphics[height =0.74in]{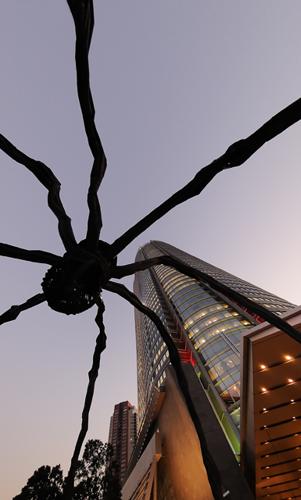}&
\hspace{-1mm}\includegraphics[height =0.74in]{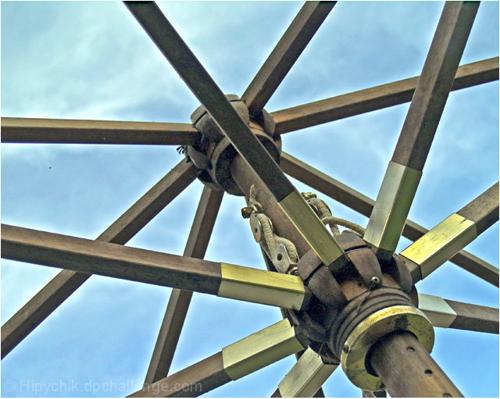}\hspace{0.1mm}
\includegraphics[height =0.74in]{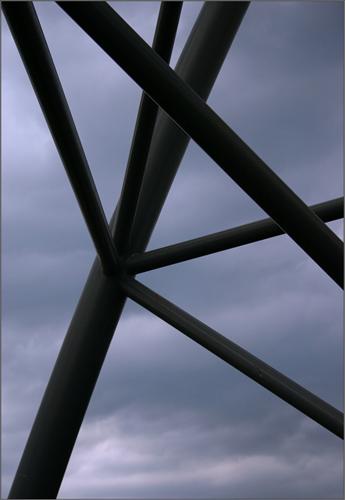}\hspace{0.1mm}
\includegraphics[height =0.74in]{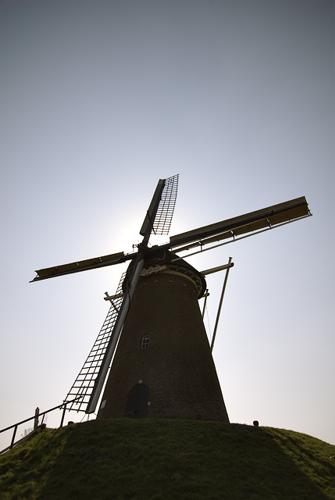}\hspace{0.1mm}
\includegraphics[height =0.74in]{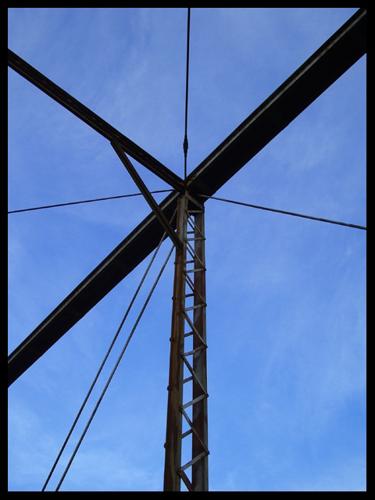}\hspace{0.1mm}
\\
\end{tabular}
\caption{Some failure cases of our method on architecture photos. Each row shows a query image (first image from the left) and the top-ranked images retrieved by our method.}
\label{fig:rerank-arch}
\end{figure}

To quantitatively evaluate the performance, we use the website developed in Section~\ref{sec:retrieval-exp} to conduct a user study on the retrieval results for architecture photos. We have again recruited 10 participants, mostly graduate students with some basic photography knowledge. Overall, our system is ranked better 70.0\% of the time (compared to 76.7\% for landscape images). We attribute the decrease in performance to the presence of multiple vanishing points in some architecture photos. As these VPs may carry similar visual weights, our method is limited to choosing one from all the candidates, which results in some ambiguity in the retrieval results. We show such a case in Figure~\ref{fig:rerank-arch} (first row), where our algorithm is able to detect the vertical VP in the images, but the horizontal orientation of the buildings remains ambiguous. Another type of failure cases results from issues with scene semantics, as shown in Figure~\ref{fig:rerank-arch} (second row). In this case, the CNN descriptor fails to capture the building in the query image, making the retrieved images less informative.

\smallskip
In summary, the experiment results on architecture photos demonstrate that the techniques we introduce in this paper are applicable to other scene types, and importantly, they are particularly suitable and effective for natural scenes.

\subsection{Detecting Multiple Vanishing Points in Natural Scenes}

So far we have restricted our attention to the dominant VP in the image. As another interesting extension, we examine how our method performs on natural landscape images when the number of VPs in each image is not fixed. We use the entire AVA landscape dataset and label all the VPs in each image. Note that a VP is labeled as long as there are two lines in the scene converging to it, and we no longer require the VPs to lie within or near the image frame. In this way, we have obtained 1,928 images with one or more VPs.

\begin{figure}[t]
\centering
\includegraphics[height =1.8in]{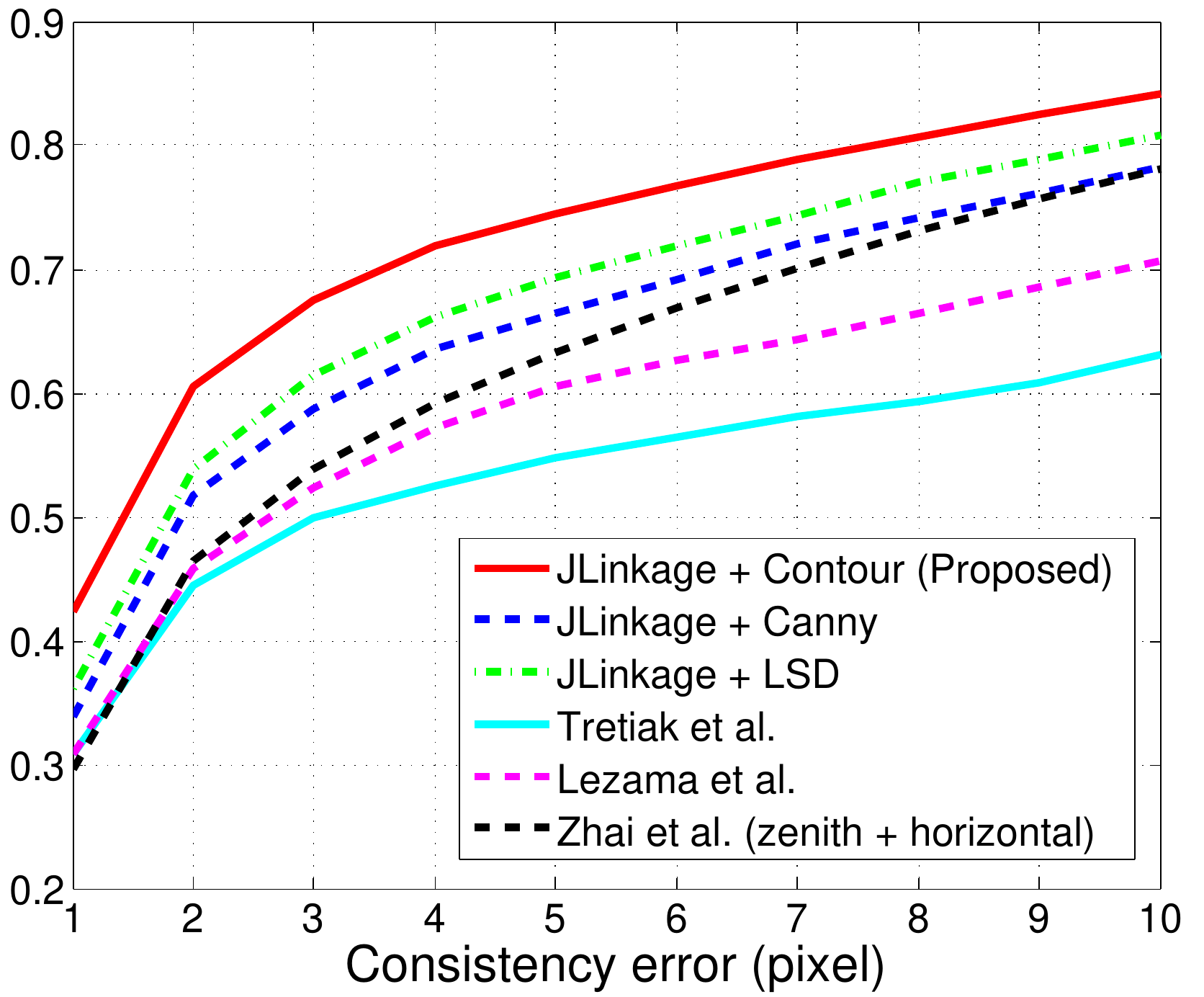}
\caption{Experiment results on detecting multiple VPs in natural landscape images.}
\label{fig:result-multi}
\end{figure}

For fair comparison, in this experiment we keep the top-3 detections of each method, since three is the maximum number of VPs we have found in any image in the AVA landscape dataset. Note that for~\cite{ZhaiWJ16} we consider both the zenith VP and the top-3 horizontal VPs in the evaluation. For each ground truth VP in an image, we find the closest detection and compute the consistency error as given in Eq.~\eqref{eq:evaluation}. As one can see in Figure~\ref{fig:result-multi}, the proposed method again outperforms all other methods. This result suggests that our method is not limited to the dominant VP, but that it can be employed as a general tool for VP detection in natural scenes.

\subsection{Limitations}

One limitation of our current system is that it is not designed to handle images in which the linear perspective is absent. For example, to convey a sense of depth, other techniques such as diminishing objects and atmospheric perspective have also been used. Instead of relying solely on the linear perspective, experienced photographers often employ multiple design principles such as balance, contrast, unity, and illumination. In the future, we plan to explore these factors for extensive understanding of photo composition.